\documentclass[preprint,12pt,authoryear]{elsarticle}

\usepackage{graphicx}%
\usepackage{multirow}%
\usepackage{amsmath,amssymb,amsfonts}%
\usepackage{amsthm}%
\usepackage{mathrsfs}%
\usepackage[title]{appendix}%
\usepackage{xcolor}%
\usepackage{bbm}
\usepackage{textcomp}%
\usepackage{manyfoot}%
\usepackage{blindtext}
\usepackage{booktabs}%
\usepackage{algorithm}%
\DeclareMathOperator*{\argmax}{argmax}
\DeclareMathOperator*{\argmin}{argmin}
\usepackage{algorithmicx}%
\usepackage{algpseudocode}%
\usepackage{multibib}
\usepackage{listings}%
\usepackage{anyfontsize}
\usepackage{colortbl}
\usepackage{gensymb}%
\newcommand{\tabincell}[2]{\begin{tabular}{@{}#1@{}}#2\end{tabular}}

\usepackage[
    framemethod=TikZ,
    roundcorner=5pt,
    leftmargin =0cm,
    middlelinewidth=0pt,
    backgroundcolor=gray!10,
]{mdframed}

\usepackage{fancyvrb}

\makeatletter
\def\PYG@reset{\let\PYG@it=\relax \let\PYG@bf=\relax%
    \let\PYG@ul=\relax \let\PYG@tc=\relax%
    \let\PYG@bc=\relax \let\PYG@ff=\relax}
\def\PYG@tok#1{\csname PYG@tok@#1\endcsname}
\def\PYG@toks#1+{\ifx\relax#1\empty\else%
    \PYG@tok{#1}\expandafter\PYG@toks\fi}
\def\PYG@do#1{\PYG@bc{\PYG@tc{\PYG@ul{%
    \PYG@it{\PYG@bf{\PYG@ff{#1}}}}}}}
\def\PYG#1#2{\PYG@reset\PYG@toks#1+\relax+\PYG@do{#2}}

\@namedef{PYG@tok@w}{\def\PYG@tc##1{\textcolor[rgb]{0.73,0.73,0.73}{##1}}}
\@namedef{PYG@tok@c}{\let\PYG@it=\textit\def\PYG@tc##1{\textcolor[rgb]{0.24,0.48,0.48}{##1}}}
\@namedef{PYG@tok@cp}{\def\PYG@tc##1{\textcolor[rgb]{0.61,0.40,0.00}{##1}}}
\@namedef{PYG@tok@k}{\let\PYG@bf=\textbf\def\PYG@tc##1{\textcolor[rgb]{0.00,0.50,0.00}{##1}}}
\@namedef{PYG@tok@kp}{\def\PYG@tc##1{\textcolor[rgb]{0.00,0.50,0.00}{##1}}}
\@namedef{PYG@tok@kt}{\def\PYG@tc##1{\textcolor[rgb]{0.69,0.00,0.25}{##1}}}
\@namedef{PYG@tok@o}{\def\PYG@tc##1{\textcolor[rgb]{0.40,0.40,0.40}{##1}}}
\@namedef{PYG@tok@ow}{\let\PYG@bf=\textbf\def\PYG@tc##1{\textcolor[rgb]{0.67,0.13,1.00}{##1}}}
\@namedef{PYG@tok@nb}{\def\PYG@tc##1{\textcolor[rgb]{0.00,0.50,0.00}{##1}}}
\@namedef{PYG@tok@nf}{\def\PYG@tc##1{\textcolor[rgb]{0.00,0.00,1.00}{##1}}}
\@namedef{PYG@tok@nc}{\let\PYG@bf=\textbf\def\PYG@tc##1{\textcolor[rgb]{0.00,0.00,1.00}{##1}}}
\@namedef{PYG@tok@nn}{\let\PYG@bf=\textbf\def\PYG@tc##1{\textcolor[rgb]{0.00,0.00,1.00}{##1}}}
\@namedef{PYG@tok@ne}{\let\PYG@bf=\textbf\def\PYG@tc##1{\textcolor[rgb]{0.80,0.25,0.22}{##1}}}
\@namedef{PYG@tok@nv}{\def\PYG@tc##1{\textcolor[rgb]{0.10,0.09,0.49}{##1}}}
\@namedef{PYG@tok@no}{\def\PYG@tc##1{\textcolor[rgb]{0.53,0.00,0.00}{##1}}}
\@namedef{PYG@tok@nl}{\def\PYG@tc##1{\textcolor[rgb]{0.46,0.46,0.00}{##1}}}
\@namedef{PYG@tok@ni}{\let\PYG@bf=\textbf\def\PYG@tc##1{\textcolor[rgb]{0.44,0.44,0.44}{##1}}}
\@namedef{PYG@tok@na}{\def\PYG@tc##1{\textcolor[rgb]{0.41,0.47,0.13}{##1}}}
\@namedef{PYG@tok@nt}{\let\PYG@bf=\textbf\def\PYG@tc##1{\textcolor[rgb]{0.00,0.50,0.00}{##1}}}
\@namedef{PYG@tok@nd}{\def\PYG@tc##1{\textcolor[rgb]{0.67,0.13,1.00}{##1}}}
\@namedef{PYG@tok@s}{\def\PYG@tc##1{\textcolor[rgb]{0.73,0.13,0.13}{##1}}}
\@namedef{PYG@tok@sd}{\let\PYG@it=\textit\def\PYG@tc##1{\textcolor[rgb]{0.73,0.13,0.13}{##1}}}
\@namedef{PYG@tok@si}{\let\PYG@bf=\textbf\def\PYG@tc##1{\textcolor[rgb]{0.64,0.35,0.47}{##1}}}
\@namedef{PYG@tok@se}{\let\PYG@bf=\textbf\def\PYG@tc##1{\textcolor[rgb]{0.67,0.36,0.12}{##1}}}
\@namedef{PYG@tok@sr}{\def\PYG@tc##1{\textcolor[rgb]{0.64,0.35,0.47}{##1}}}
\@namedef{PYG@tok@ss}{\def\PYG@tc##1{\textcolor[rgb]{0.10,0.09,0.49}{##1}}}
\@namedef{PYG@tok@sx}{\def\PYG@tc##1{\textcolor[rgb]{0.00,0.50,0.00}{##1}}}
\@namedef{PYG@tok@m}{\def\PYG@tc##1{\textcolor[rgb]{0.40,0.40,0.40}{##1}}}
\@namedef{PYG@tok@gh}{\let\PYG@bf=\textbf\def\PYG@tc##1{\textcolor[rgb]{0.00,0.00,0.50}{##1}}}
\@namedef{PYG@tok@gu}{\let\PYG@bf=\textbf\def\PYG@tc##1{\textcolor[rgb]{0.50,0.00,0.50}{##1}}}
\@namedef{PYG@tok@gd}{\def\PYG@tc##1{\textcolor[rgb]{0.63,0.00,0.00}{##1}}}
\@namedef{PYG@tok@gi}{\def\PYG@tc##1{\textcolor[rgb]{0.00,0.52,0.00}{##1}}}
\@namedef{PYG@tok@gr}{\def\PYG@tc##1{\textcolor[rgb]{0.89,0.00,0.00}{##1}}}
\@namedef{PYG@tok@ge}{\let\PYG@it=\textit}
\@namedef{PYG@tok@gs}{\let\PYG@bf=\textbf}
\@namedef{PYG@tok@gp}{\let\PYG@bf=\textbf\def\PYG@tc##1{\textcolor[rgb]{0.00,0.00,0.50}{##1}}}
\@namedef{PYG@tok@go}{\def\PYG@tc##1{\textcolor[rgb]{0.44,0.44,0.44}{##1}}}
\@namedef{PYG@tok@gt}{\def\PYG@tc##1{\textcolor[rgb]{0.00,0.27,0.87}{##1}}}
\@namedef{PYG@tok@err}{\def\PYG@bc##1{{\setlength{\fboxsep}{\string -\fboxrule}\fcolorbox[rgb]{1.00,0.00,0.00}{1,1,1}{\strut ##1}}}}
\@namedef{PYG@tok@kc}{\let\PYG@bf=\textbf\def\PYG@tc##1{\textcolor[rgb]{0.00,0.50,0.00}{##1}}}
\@namedef{PYG@tok@kd}{\let\PYG@bf=\textbf\def\PYG@tc##1{\textcolor[rgb]{0.00,0.50,0.00}{##1}}}
\@namedef{PYG@tok@kn}{\let\PYG@bf=\textbf\def\PYG@tc##1{\textcolor[rgb]{0.00,0.50,0.00}{##1}}}
\@namedef{PYG@tok@kr}{\let\PYG@bf=\textbf\def\PYG@tc##1{\textcolor[rgb]{0.00,0.50,0.00}{##1}}}
\@namedef{PYG@tok@bp}{\def\PYG@tc##1{\textcolor[rgb]{0.00,0.50,0.00}{##1}}}
\@namedef{PYG@tok@fm}{\def\PYG@tc##1{\textcolor[rgb]{0.00,0.00,1.00}{##1}}}
\@namedef{PYG@tok@vc}{\def\PYG@tc##1{\textcolor[rgb]{0.10,0.09,0.49}{##1}}}
\@namedef{PYG@tok@vg}{\def\PYG@tc##1{\textcolor[rgb]{0.10,0.09,0.49}{##1}}}
\@namedef{PYG@tok@vi}{\def\PYG@tc##1{\textcolor[rgb]{0.10,0.09,0.49}{##1}}}
\@namedef{PYG@tok@vm}{\def\PYG@tc##1{\textcolor[rgb]{0.10,0.09,0.49}{##1}}}
\@namedef{PYG@tok@sa}{\def\PYG@tc##1{\textcolor[rgb]{0.73,0.13,0.13}{##1}}}
\@namedef{PYG@tok@sb}{\def\PYG@tc##1{\textcolor[rgb]{0.73,0.13,0.13}{##1}}}
\@namedef{PYG@tok@sc}{\def\PYG@tc##1{\textcolor[rgb]{0.73,0.13,0.13}{##1}}}
\@namedef{PYG@tok@dl}{\def\PYG@tc##1{\textcolor[rgb]{0.73,0.13,0.13}{##1}}}
\@namedef{PYG@tok@s2}{\def\PYG@tc##1{\textcolor[rgb]{0.73,0.13,0.13}{##1}}}
\@namedef{PYG@tok@sh}{\def\PYG@tc##1{\textcolor[rgb]{0.73,0.13,0.13}{##1}}}
\@namedef{PYG@tok@s1}{\def\PYG@tc##1{\textcolor[rgb]{0.73,0.13,0.13}{##1}}}
\@namedef{PYG@tok@mb}{\def\PYG@tc##1{\textcolor[rgb]{0.40,0.40,0.40}{##1}}}
\@namedef{PYG@tok@mf}{\def\PYG@tc##1{\textcolor[rgb]{0.40,0.40,0.40}{##1}}}
\@namedef{PYG@tok@mh}{\def\PYG@tc##1{\textcolor[rgb]{0.40,0.40,0.40}{##1}}}
\@namedef{PYG@tok@mi}{\def\PYG@tc##1{\textcolor[rgb]{0.40,0.40,0.40}{##1}}}
\@namedef{PYG@tok@il}{\def\PYG@tc##1{\textcolor[rgb]{0.40,0.40,0.40}{##1}}}
\@namedef{PYG@tok@mo}{\def\PYG@tc##1{\textcolor[rgb]{0.40,0.40,0.40}{##1}}}
\@namedef{PYG@tok@ch}{\let\PYG@it=\textit\def\PYG@tc##1{\textcolor[rgb]{0.24,0.48,0.48}{##1}}}
\@namedef{PYG@tok@cm}{\let\PYG@it=\textit\def\PYG@tc##1{\textcolor[rgb]{0.24,0.48,0.48}{##1}}}
\@namedef{PYG@tok@cpf}{\let\PYG@it=\textit\def\PYG@tc##1{\textcolor[rgb]{0.24,0.48,0.48}{##1}}}
\@namedef{PYG@tok@c1}{\let\PYG@it=\textit\def\PYG@tc##1{\textcolor[rgb]{0.24,0.48,0.48}{##1}}}
\@namedef{PYG@tok@cs}{\let\PYG@it=\textit\def\PYG@tc##1{\textcolor[rgb]{0.24,0.48,0.48}{##1}}}

\def\PYGZbs{\char`\\}
\def\PYGZus{\char`\_}

\def\PYGZsh{\char`\#}

\def\PYGZhy{\char`\-}
\def\PYGZsq{\char`\'}
\def\PYGZdq{\char`\"}

\makeatother

\newcommand{\ms}[2]{{#1}{\footnotesize $\,\pm${#2}}}
\newcommand{\msb}[2]{\textbf{{#1}}$\,\pm${\footnotesize {#2}}}      
\newcommand{\msbb}[2]{\underline{{#1}}$\,\pm${\footnotesize {#2}}}  

\usepackage{nameref,zref-xr}
\usepackage[colorlinks=true,citecolor=blue,urlcolor=black]{hyperref}
\zxrsetup{toltxlabel}

\journal{Medical Image Analysis}

\begin{document}

\begin{frontmatter}

\title{Estimating Performance for Safe Deployment of Machine Learning Models} 

\author[1,2]{Zeju Li}
\author[3]{Konstantinos Kamnitsas}
\author[4]{Mobarak I. Hoque}
\author[5]{Mélanie Roschewitz}
\author[3,5,7]{Cheng Ouyang}
\author[6,5]{Chen Chen}
\author[5]{Ben Glocker}

\affiliation[1]{organization={College of Biomedical Engineering, Fudan University},
            state={Shanghai},
            country={China}}
\affiliation[2]{organization={Oxford Centre for Integrative Neuroimaging, University of Oxford},
            state={Oxford},
            country={UK}}
\affiliation[3]{organization={Department of Engineering Science, University of Oxford},
            state={Oxford},
            country={UK}}
\affiliation[4]{organization={Division of Informatics, Imaging and Data Science, University of Manchester},
            state={Manchester},
            country={UK}}
\affiliation[5]{organization={Department of Computing, Imperial College London},
            state={London},
            country={UK}}
\affiliation[6]{organization={School of Computer Science, University of Sheffield},
            state={Sheffield},
            country={UK}}
\affiliation[7]{organization={Institute of Clinical Sciences, Imperial College London},
            state={London},
            country={UK}}


\begin{abstract}
The main challenge in applying deep neural networks to real-world scenarios lies in distributional discrepancies between training and test datasets, often caused by variations in data acquisition conditions. Assessing the performance of pre-trained models without labels is crucial for safety, but this is complicated by deep models' overly optimistic probabilistic outputs, often exacerbated by class imbalances, leading to unreliable predictions.
To address these issues, we introduce MOVAL, a novel MOdel eVALuation tool that estimates performance using unlabeled test data. MOVAL incorporates class-wise confidence calibration to adjust model outputs, allowing for the estimation of metrics beyond accuracy and mitigating the detrimental effects of class imbalance on confidence calibration. Additionally, we propose an ensemble approach that combines multiple calibrated confidence scores for a more robust solution.
We validate MOVAL on natural and medical image datasets with various acquisition shifts, covering tasks from classification to segmentation. It achieves state-of-the-art results across evaluation metrics such as accuracy, F1-Score, AUC, and more. Furthermore, we demonstrate that MOVAL provides accurate performance estimations at the class, instance, and pixel levels, ensuring the transparent and safe deployment of machine learning models. \textcolor{black}{We provide MOVAL as a deployment tool for monitoring and estimating performance under a range of domain shifts.}
\end{abstract}




\end{frontmatter}


\section{Introduction}
\label{sec:intro}

The witnessed success of machine learning is built upon the independently and identically distributed (IID) assumption. However, this assumption does not always hold in real-world applications due to domain shifts, where the training and test data come from different distributions. There are multiple kinds of domain shifts, but we primarily consider acquisition shifts, as they have the most direct and significant impact on model performance, causing models to fail silently. Therefore, trustworthy safeguards are needed to ensure model performance with minimal assumptions and to provide prompt feedback without access to ground truth~\citep{eche2021toward}. Existing algorithms for model evaluation are either task-specific or rely on globally calibrated probabilities, which do not provide a comprehensive evaluation beyond accuracy or Dice similarity coefficient (DSC)~\citep{ZENK2025103392}. In this work, we combine recent advances in model calibration and confidence score estimation to develop a new model evaluation method that can be applied to various types of tasks, informing model failure with multiple estimated metrics and meaningful indicators at different levels.

\textcolor{black}{Developing reliable, real-time performance assessment systems is crucial for the safe deployment of machine learning in clinical practice. Such systems help identify the limits of model applicability and support practitioners in iteratively refining their training strategies. For instance, when a skin-lesion classification model is trained on data from a source domain, deployment decisions in a new target domain can be guided by the model’s predicted accuracy on all unlabeled test images. Additionally, because magnetic resonance imaging (MRI) is typically non-quantitative, it is often unclear whether a model remains valid when applied in a local hospital where acquisition protocols differ and expert annotations may be unavailable. This approach helps prevent overfitting when optimizing results using a small number of labeled test images by using predicted accuracy as a surrogate to validation loss.}

\subsection{Related Works}
\label{subsec:related}

Many methods have been proposed for this scenario, but they often lack robustness or have limited applicability. For example, distance metric between the training and test datasets can be a good indicator for model performance, but requires the access to validation data every time of estimating~\citep{horvitz1952generalization, chen2021mandoline} and/or need another regression model~\citep{deng2021labels, robinson2018real}. Reference-based methods~\citep{fan2006reverse, wang2020deep, valindria2017reverse, li2022towards} estimate model performance by comparing the prediction results to a predefined reference (i.e. shape template for segmentation), but lack the ability to leverage the probability of the model output, and are mostly limited to segmentation tasks with strong shape priors. \textcolor{black}{As a separate but related line of work, conformal prediction methods~\citep{barber2023conformal} provide prediction sets with formal coverage guarantees, which can increase confidence at deployment but do not directly reflect the underlying model’s performance.} \textcolor{black}{These are related but distinct goals: conformal prediction provides set-valued uncertainty guarantees for individual predictions under specific assumptions (e.g., exchangeability or its relaxations), whereas MOVAL estimates aggregate, class-, instance-, and pixel-wise performance metrics (accuracy, F1-Score, sensitivity, precision, AUC) without providing per-prediction coverage guarantees.}

\textcolor{black}{Confidence-score-based methods, which estimate the accuracy of a trained model using the mean value of the maximum class probability (MCP), are more efficient and are believed to be accurate for model evaluation~\citep{elsahar2019annotate, mehrtash2020confidence, garg2022leveraging, islam2021class, baek2022agreement, ouyang2022improved, roschewitz2023distance, lu2023characterizing, press2024entropy}.} However, this approach is largely limited to accuracy estimation, which can be a misleading metric for imbalanced datasets, a common occurrence in real-world classification tasks, particularly in segmentation. For example, accuracy can approach 100$\%$ in imbalanced datasets, even if all samples are classified as belonging to the majority class. In medical applications, it is often preferable to track clinically relevant metrics such as sensitivity or specificity, or metrics that take into account class imbalance, such as area under the curve (AUC) or F1-Score.

Model calibration, also known as confidence calibration, is the process of adjusting the probability outputs of a trained model to better reflect the true likelihood of its predictions. This adjustment is typically performed using validation data, aiming to ensure that the calibrated model's confidence scores align with the observed accuracy on unseen test data. A model is said to be well-calibrated when the predicted confidence aligns with the true probability that the associated model prediction is correct. However, most methods assume that every class performs equally well~\citep{guo2017calibration, guillory2021predicting, garg2022leveraging, kivimaki2024confidence, mehrtash2020confidence}, thus calibrating the probabilities of all classes in the same way. In practice, this is not true, especially in the presence of class imbalance, where predictions for majority classes tend to be overconfident, while predictions for minority classes are often underconfident~\citep{li2020analyzing}. Consequently, after calibration, although global metrics may be aligned, the discrepancy between different classes becomes even larger. This not only increases the estimation error on test data but also leads to misleading results when examining class-wise or instance-wise performance~\citep{li2022estimating}.

\textcolor{black}{Several methods target multi-class calibration with class-dependent transformations, including Dirichlet calibration~\citep{kull2019beyond}, matrix scaling and its regularized variants, and related approaches~\citep{zong2024dirichlet, lecoz2024confidence, zhao2021calibrating, valk2023assuming}. These methods can, in principle, apply different corrections across classes via a full $c \times c$ transformation. However, they are fit by minimizing an aggregate calibration loss (e.g., negative log-likelihood or expected calibration error) over the whole validation set rather than a class-specific objective, so they do not explicitly enforce that each class's calibrated confidence matches that class's true performance; correcting class imbalance is not their stated goal.}

\subsection{Contributions}
\label{subsec:contribution}

We address these open challenges with MOVAL, a new MOdel eVALuation \textcolor{black}{framework, benchmark, and deployable tool} for assessing robust and comprehensive model performance under domain shifts without the need for manual labels. \textcolor{black}{Specifically, we propose to enhance confidence score-based approaches through class-wise calibration, which aligns confidence estimates with performance metrics using class-specific parameters. This not only improves accuracy estimation by reducing bias between classes but also naturally extends confidence-based methods to the estimation of other metrics, including F1-Score, sensitivity, precision, and AUC. To our knowledge, MOVAL is the first benchmark to evaluate how well confidence scores estimate metrics beyond accuracy.} \textcolor{black}{Rather than introducing a fundamentally new calibration algorithm, our contribution lies in systematically unifying and extending existing calibration, confidence-score, and ensemble ideas across a broader range of metrics and tasks than prior work.}

\textcolor{black}{MOVAL is the first unified confidence score-based estimation framework applicable to tasks beyond classification—including 2D and 3D segmentation; existing works are primarily limited to classification. We systemically extend various confidence scores—beyond MCP—to the problem of performance estimation. This includes scores like Entropy~\citep{steinhardt2016unsupervised}, Doctor~\citep{granese2021doctor}, and Energy~\citep{liu2020energy}, which were designed for identifying model failure but have not been extensively discussed in the context of performance estimation. Importantly, we find that an ensemble of multiple confidence scores provides the most robust and accurate performance estimation.} We provide this as a ready-to-use tool and anticipate that practitioners will integrate it into their existing deployment pipelines as a standard procedure~\footnote{\url{https://pypi.org/project/moval/}}. MOVAL does not require validation data at the inference stage, impose assumptions on the tasks, or rely on additional regression models, making it lightweight, flexible, and reproducible. Through experiments on six tasks across 6,572 test settings, MOVAL outperforms state-of-the-art methods in several areas, reducing estimation error by 43\% in real-world classification tasks and by over 50\% in segmentation tasks. Additionally, it provides detailed indicator results at the class-wise, instance-wise, and pixel-wise levels.

We presented our preliminary results at MICCAI 2022~\citep{li2022estimating}, which mainly focused on the discussion of class-wise calibration for performance estimation, \textcolor{black}{where it was shown to outperform class-wise calibration baselines such as vector scaling}. Building on that, this paper further discusses: 1) the estimation of metrics other than accuracy, 2) estimation for 2D and 3D segmentation tasks, 3) estimation with different confidence scores and their ensemble with additional evidence, and most importantly, 4) the development of an open-source tool for the community to enable the safe deployment of machine learning models.

\section{Preliminaries: Confidence-Based Model Evaluation}
\label{sec:preliminaries}

\subsection{Problem Setup}

In the context of \(c\)-class image classification, our goal is to establish a mapping between the image space \(\mathcal{X} \subset \mathbbm{R}^d\) and the label space \(\mathcal{Y} = \{1, 2, \ldots, c\}\). This framework can be extended to segmentation tasks by assuming that an image patch $\boldsymbol{x}_i$ is sampled from \(\mathcal{X}\), with \(y_i \in \mathcal{Y}\) representing the label of the central pixels.

We assume that two labeled image sets, $\mathcal{D}^{Tr} = \{ (\boldsymbol{x}_i, y_i) \}_{i = 1}^I$ and $\mathcal{D}^V = \{ (\boldsymbol{x}_i, y_i) \}_{i = 1}^N$, drawn from the same source domain, are available for model training and validation. Once trained, the model is applied to a target domain $\mathcal{D}^{Te}$, where the distribution differs from the source domain, a common scenario in real-world applications. Our goal is to estimate its performance, including accuracy, F1-Score, sensitivity, precision, and AUC, on the target domain to support decision-making when only a set of test images $\mathcal{D}^{Te} = \{ \boldsymbol{x}^{Te}_i \}_{i = 1}^M$ is available without any ground truth.

We propose MOVAL, a framework for estimating model performance under domain shifts without requiring ground truth labels, specifically for modern machine learning models like convolutional neural networks (CNNs). MOVAL leverages logits (pre-softmax outputs) to estimate metrics across the dataset. To ensure these probabilities are useful for performance estimation under domain shifts, they are adjusted to match real metrics on a labeled validation set from the source domain.

After training a neural network \( f \) on the training dataset \(\mathcal{D}^{Tr}\), we aim to develop a calibration model based on its performance on the validation dataset \(\mathcal{D}^V = \{ (\boldsymbol{x}_i, y_i) \}_{i=1}^N\). Specifically, given an input sample \(\boldsymbol{x}_i\) from the validation set, the network output \( f(\boldsymbol{x}_i) \) (i.e., logits) is obtained. Our objective is to minimize the following equation based on the validation dataset with respect to the parameters in the MOVAL framework \(\mathcal{M}\):

\begin{equation}
    \bigg| 
    \mathcal{M}\Big(\{f(\boldsymbol{x}_i)\}_{i=1}^N\Big)
    -
    \mathcal{C}\Big(\{(f(\boldsymbol{x}_i), y_i)\}_{i=1}^N\Big)
    \bigg|,
\end{equation}

\noindent \textcolor{black}{where} \(\mathcal{C}\) refers to the calculation of a metric that evaluates the model's performance across the entire labeled validation dataset. In contrast, \(\mathcal{M}\) estimates these metrics based solely on the network outputs and $\big| \cdot \big|$ refers to a desired measurement of the gap, such as L1 distance. After optimization, \(\mathcal{M}\) can be applied directly to any test dataset, and we expect the estimated metrics to offer more accurate insights into performance, particularly in the presence of domain shifts.

MOVAL maps a set of logits to the final metrics through three key procedures: confidence score computation, model calibration, and metric calculation (Fig.~\ref{fig:method}b). To enable the calibration process, a tunable parameter is typically added to the confidence score computation. Different calibration techniques, such as Temperature Scaling (TS)~\citep{guo2017calibration}, Differences of Confidences (DoC)~\citep{guillory2021predicting}, and Average Thresholded Confidence (ATC)~\citep{garg2022leveraging}, define this parameter in distinct ways. MOVAL extends these methods by altering the optimization objectives on a per-class basis—\(\mathcal{C}\) and \(\mathcal{M}\) calculate the metric for each class individually, applying class-specific parameters. This not only enhances accuracy estimation under class imbalance but is also a key step in expanding the MOVAL framework to metrics beyond accuracy and tasks beyond classification—areas that have been largely unexplored. 

\textcolor{black}{MOVAL's flexibility makes it more effective for supporting deployment decisions for machine learning models, by making the trade-offs among different performance metrics explicit.} For example, in brain lesion segmentation from MRI data, accuracy alone is insufficient. To properly evaluate segmentation performance, tracking the F1-Score and focusing on pixel-level insights is critical. If minimizing missed lesions is a priority, sensitivity estimation is essential. Alternatively, when assessing post-processing techniques aimed at reducing false positives without significantly affecting overall performance, precision becomes the most appropriate metric to track.

\begin{figure*}[ht]
  \centering
   \includegraphics[width=\linewidth]{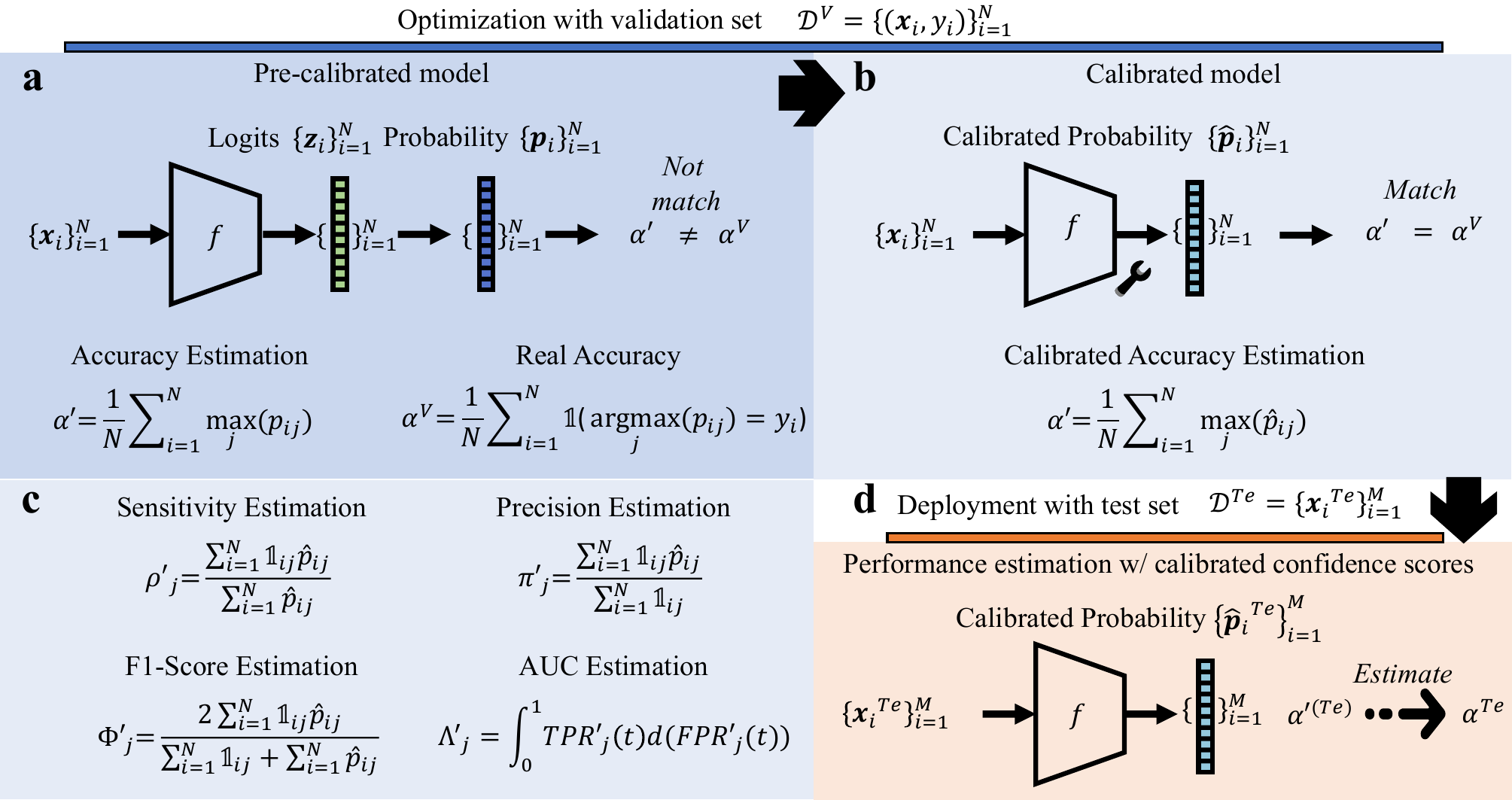}
   \caption{\textcolor{black}{Detailed descriptions of the confidence calibration process.} \textcolor{black}{Here $p_{ij}$ denotes the raw (uncalibrated) softmax probability that unit $i$ belongs to class $j$, where $i$ indexes a sample for classification and a pixel/voxel for segmentation, assuming a single volume is under consideration; after calibration, $p_{ij}$ is replaced by the calibrated probability $\hat{p}_{ij}$ used to compute all reported metric estimates.}
   \textbf{(a)}: Based on the model probabilities $\boldsymbol{p}_i$, multiple performance metrics can be estimated. As we illustrate with the example of accuracy, vanilla model probabilities are well known to be overconfident. Therefore, the estimation is prone to being too optimistic and not accurate. 
   \textbf{(b)}: The calibration process adjusts the probabilities such that the estimation aligns with the real metrics on the validation dataset. 
   \textbf{(c)}: The same principle can be applied to estimating other group-wise metrics.
   \textbf{(d)}: The calibrated probabilities are expected to produce more accurate estimations when exposed to domain shifts.
   }
   \label{fig:preliminaries}
\end{figure*}

\subsection{Accuracy Estimation}

The confidence score indicates the likelihood that the predicted label is correct and can estimate accuracy for a specific dataset. We first describe a general accuracy estimation framework based on MCP, which aligns well with existing confidence-based methods.

Given an input $\boldsymbol{x}_i$, the trained model $f$ generates the output logit $\boldsymbol{z}_i = f(\boldsymbol{x}_i) \in \mathbbm{R}^c$. The predicted probability vector $\boldsymbol{p}_i$ is then computed by applying the softmax function to $\boldsymbol{z}_i$, where $p_{ij} = \sigma(\boldsymbol{z}_i)_j = \frac{\mathrm{e}^{z_{ij}}}{\sum_{j=1}^{c}\mathrm{e}^{z_{ij}}}$ for the $j$'th class. The predicted class label $y^{\prime}_i$ corresponds to the class with the highest probability, given by $y^{\prime}_i = \argmax_j (p_{ij})$. \textcolor{black}{The confidence score $q_i$ for this sample, modeled by MCP, can be calculated as $q_i = \max_j (p_{ij})$.}

The confidence score is a number ranging from 0 to 1; the higher the number, the more likely the sample is correctly classified. Ideally, we want the confidence to have quantitative meaning that informs the decision-making process. Specifically, for the entire validation dataset $\mathcal{D}^V$, \textcolor{black}{we calculate the average confidence with $\alpha' = \frac{1}{N}\sum_{i=1}^{N}q_{i}$, and expect it to reflect the real accuracy.} In practice, we can apply a similar procedure to the test dataset under domain shifts $\mathcal{D}^{Te}$ and obtain the estimated accuracy as $\alpha^{\prime(Te)} = \frac{1}{M}\sum_{i=1}^{M}q_{i}^{Te}$ (Fig.~\ref{fig:preliminaries}).

However, directly computing confidence $q_{i}$ over the softmax function for estimation can be problematic as the predicted probability cannot reflect the real accuracy with modern neural networks~\citep{guo2017calibration}. Therefore, we need to adjust the confidence to obtain the modified confidence $\hat{q}_i$, a process commonly known as model calibration. To achieve this, additional parameters are introduced into the confidence calculation process, enabling fine-tuning. While methods like TS, DoC, and ATC define these parameters differently, they all share the same optimization function. Specifically, a common approach is to use the validation accuracy, defined as $\alpha^V=\frac{1}{N}\sum_{i=1}^{N}\mathbbm{1}{(y^{\prime}_i=y_i)}$\footnote{$\mathbbm{1}$ is an indicator function that equals 1 when the underlying criterion is met and 0 otherwise.}, as a surrogate objective for $\alpha'$ to find the solution~\citep{elsahar2019annotate, guillory2021predicting, garg2022leveraging}. Subsequently, the same calibration process is applied to the test dataset, with the expectation that the estimation results using the calibrated confidence $\hat{q}_i^{Te}$ will be more accurate.

\begin{figure*}[ht]
  \centering
   \includegraphics[width=\linewidth]{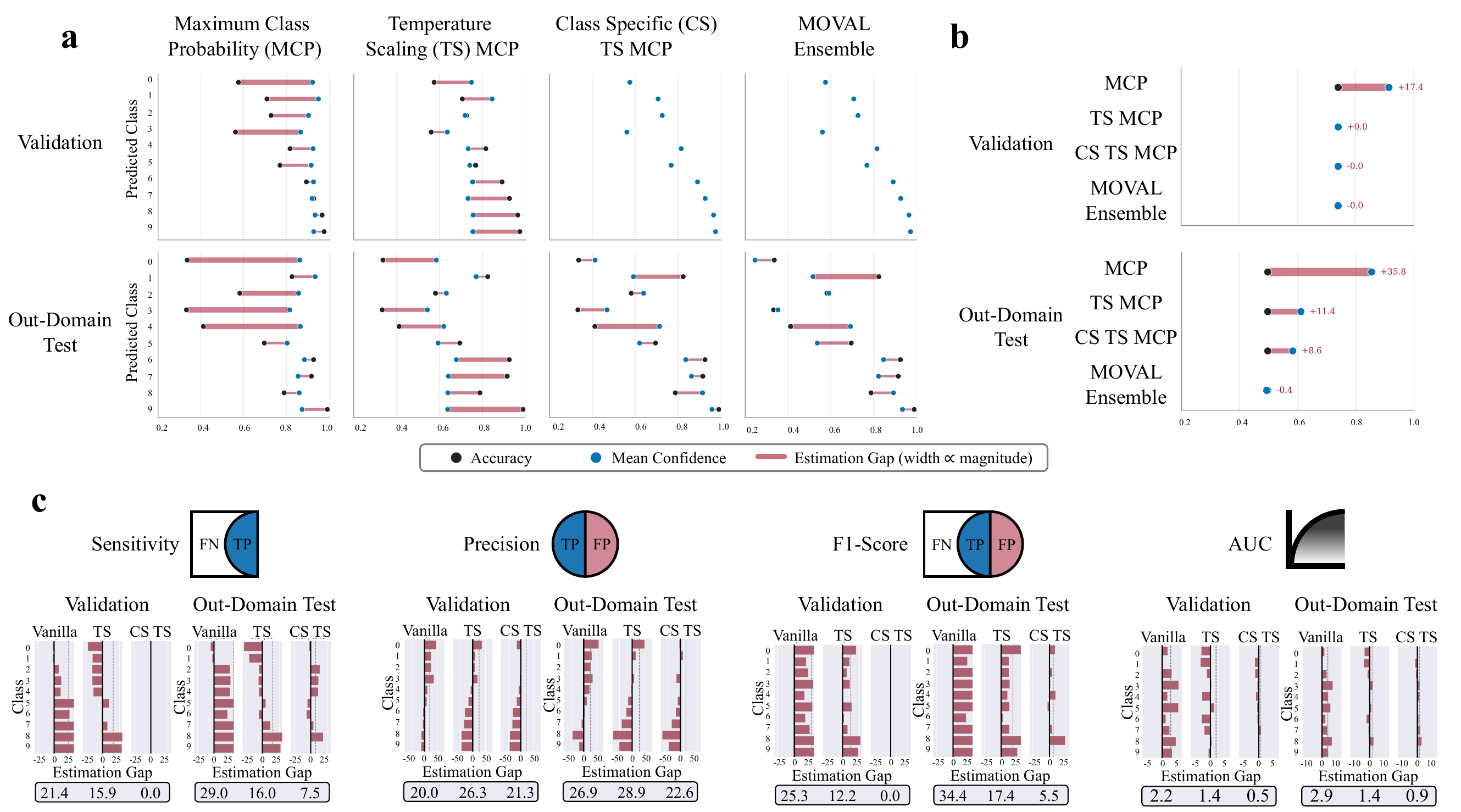}
   \caption{\textcolor{black}{\textcolor{black}{Class-wise calibration and multi-metric estimation with MOVAL.} \textcolor{black}{We show an estimation example on CIFAR10-LT, where sample counts decrease from class 0 to class 9.}}
   \textbf{(a)}: MOVAL facilitates class-wise calibration with class-specific parameters. Class-wise calibration allows for the alignment of estimated class-wise metrics and the true underlying counterparts, which can mitigate model biases caused by class imbalance.
   \textbf{(b)}: \textcolor{black}{Overall accuracy estimation across different methods. By aggregating multiple confidence scores, MOVAL Ensemble achieves the most accurate results.}
   \textbf{(c)}: In addition to accuracy estimation, which is provided by existing methods, class-wise alignment enables MOVAL to estimate class-wise metrics, thus providing a comprehensive evaluation.
   }
   \label{fig:method}
\end{figure*}

\begin{figure*}[ht]
  \centering
   \includegraphics[width=\linewidth]{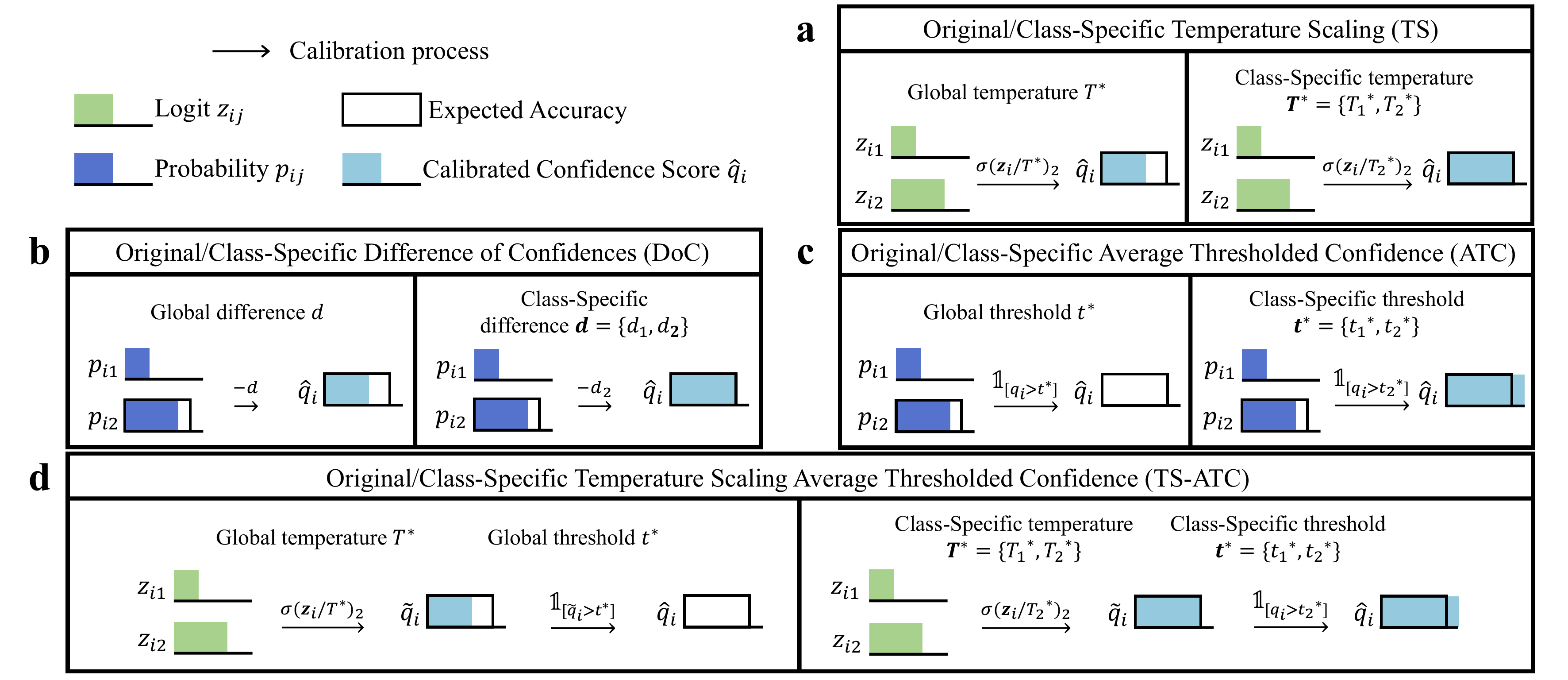}
   \caption{We extend existing model calibration methods to class-specific variants by introducing parameters for different classes. This enables class-wise calibration and improves estimation accuracy, especially when the model is trained with class-imbalanced data. 
   }
   \label{fig:method_details}
\end{figure*}

We argue that existing confidence-based accuracy estimation methods struggle to generalize effectively to imbalanced datasets, which are common in medical image classification and segmentation. Prior studies have demonstrated that models trained on imbalanced data are prone to misclassifying minority class samples as belonging to the majority class, but rarely the reverse~\citep{li2020analyzing}. \textcolor{black}{As a result, these models often achieve high overall accuracy but are underconfident in predicting minority class samples, while being overly confident for majority class predictions (Fig.~\ref{fig:method}c). Existing calibration methods apply a uniform probability-adjustment procedure across all classes. This uniformity leads to higher estimation error for minority-class samples (e.g., TS MCP exhibits substantial estimation error on minority classes, as shown in Fig.~\ref{fig:method}a), which in turn limits their ability to generalize effectively (Fig.~\ref{fig:method}b).} To address this, we introduce class-specific parameters into the performance estimation framework using adaptive optimization, reducing the confidence bias of neural networks toward minority classes (Fig.~\ref{fig:method_details}). Other group-wise metrics, such as sensitivity, precision,  F1-Score and AUC, can be formulated similarly. The details are provided in the Suppl. Note.

\section{Proposed Method: MOVAL}
\label{sec:method}

The MOVAL model consists of three key components. 1) Confidence Score Calculation: Given the network logits $\boldsymbol{z}_i$, the confidence score $q_i$ is computed. 
2) Model Calibration: The confidence score is adjusted to align estimated metrics with target metrics, producing the calibrated confidence score $\hat{q}_i$ from $q_i$. 
3) Metrics Computation: Using $\hat{q}_i$, evaluation metrics for the entire dataset are computed, including calibrated probabilities $\hat{\boldsymbol{p}}_i$ for metrics beyond accuracy.

In following sections, we will present our main contribution, class-wise model calibration, using MCP for accuracy estimation. Subsequent sections will explore its extension to various metrics and types of confidence scores.

\subsection{Class-Specific Temperature Scaling}

The TS method is one of the most widely used calibration methods~\citep{guo2017calibration, hinton2015distilling}. It introduces a temperature parameter to rescale the logits, with $q_{i}=\max_j(\sigma(\boldsymbol{z}_i/T)_j)$. \textcolor{black}{To find the value of $T$ that makes the estimated accuracy closely approximate the true accuracy, the optimal temperature $T^*$ is determined using the validation dataset $\mathcal{D}^V$ as:}

\begin{equation}
\begin{aligned}
    T^{*} = \argmin_T\Big|\frac{1}{N}\sum_{i=1}^{N}q_i-\alpha^V\Big| = \argmin_T\Big|\frac{1}{N}\sum_{i=1}^{N}\max_j (\sigma(\frac{\boldsymbol{z}_i}{T})_j)-\alpha^V\Big|.
    \label{eq:TS}
\end{aligned}
\end{equation}

To enable class-specific calibration, we extend the temperature parameter to a vector $\boldsymbol{T}$ with $c$ elements, where the $j$'th element $T_j$ corresponds to the $j$'th class. \textcolor{black}{The calibrated confidence score is then computed as $q_{i}=\max_j(\sigma(\boldsymbol{z}_i/T_{y^{\prime}_i})_j)$, with the optimal $T_j^*$ determined by:}

\begin{equation}
\begin{aligned}
    T_j^* = \argmin_{T_j}\Big|\frac{1}{N_j}\sum_{i=1}^{N}\mathbbm{1}_{ij}q_i-\alpha^V_j\Big| = \argmin_{T_j}\Big|\frac{1}{N_j}\sum_{i=1}^{N}\mathbbm{1}_{ij} \max_k(\sigma(\frac{\boldsymbol{z}_i}{T_{j}})_k)-\alpha^V_j\Big|,
    \label{eq:CSTS}
\end{aligned}
\end{equation}

\noindent \textcolor{black}{where} $N_j=\sum_{i=1}^{N}\mathbbm{1}_{ij}$ and $\alpha^V_j$ is the accuracy for the $j$'th class on the validation set which is calculated as $\alpha^V_j=\frac{1}{N_j}\sum_{i=1}^{N}\mathbbm{1}_{ij}\mathbbm{1}{(y^{\prime}_i=y_i)}$. This approach ensures that the optimized $\boldsymbol{T}^*$ results in $\hat{q}_i$ aligning with the class-wise accuracy (Fig.~\ref{fig:method_details}a).

\subsection{Class-Specific Difference of Confidences}

The DoC method~\citep{guillory2021predicting} adjusts $q_i$ by subtracting the difference \( d = \alpha^{\prime} - \alpha^V \), which is the gap between the real accuracy \(\alpha^V\) and the estimated results \(\alpha^{\prime}\):

\begin{equation}
\hat{q}_i = q_i -d.
\label{eq:DoC}
\end{equation}

A clipping operation between 0 and 1 may be necessary to ensure that the calibrated confidence is properly normalized. We introduce a class-specific DoC method, where we define a difference vector $\boldsymbol{d}$ with $c$ elements, with the $j$'th element \(d_j\) representing the difference for the $j$'th class. The class-wise calibrated confidence score $\hat{q}_{i}$ is then computed as follows:

\begin{equation}
\hat{q}_i = q_i - d_{y^{\prime}_i},
\label{eq:CSDoC}
\end{equation}

\noindent \textcolor{black}{where} $d_j = \frac{1}{N_j} \sum_{i=1}^{N} \mathbbm{1}_{ij}q_i  - \alpha^V_j$ is the difference between the average predicted probability for samples predicted that they belong to class $j$ and $\alpha^V_j$. This approach insulates the probabilities of samples predicted as minority classes from the influence of majority classes, reducing bias (Fig.~\ref{fig:method_details}b).

\subsection{Class-Specific Average Thresholded Confidence}

The ATC method~\citep{garg2022leveraging} is a heuristic yet effective approach for estimating model performance, \textcolor{black}{consistently delivering top results across multiple robustness evaluation benchmarks~\citep{garg2022leveraging, baek2022agreement}}. It predicts accuracy by calculating the proportion of samples with $q_i$ values exceeding a learned threshold \( t \). The threshold \( t \) is determined as follows, and the confidence score is then calculated as \(\hat{q}_i = \mathbbm{1}{(q_i > t^*)}\), where:

\begin{equation}
\begin{aligned}
    t^{*} = \argmin_t\Big|\frac{1}{N}\sum_{i=1}^{N}\mathbbm{1}{(q_i>t)} - \alpha^V\Big|.
    \label{eq:ATC}
\end{aligned}
\end{equation}

In the context of ATC, \(\hat{q}_i\) is no longer a standard calibrated confidence score but instead a discrete value \(\in \{0, 1\}\). Despite this, it can still be used for accuracy estimation as previously described. For class-specific calibration, we propose learning a set of thresholds represented by the vector \(\boldsymbol{t}\), which has \(c\) elements, one for each class. The \(j\)'th element is determined by:

\begin{equation}
\begin{aligned}
    t_j^* = \argmin_{t_j}\Big|\frac{1}{N_j}\sum_{i=1}^{N}\mathbbm{1}_{ij}\mathbbm{1}{(q_i>t_j)} - \alpha^V_j\Big|.
    \label{eq:CSATC}
\end{aligned}
\end{equation}

We then calibrate the probabilities as \(\hat{q}_i = \mathbbm{1}{(q_i > t_{y^{\prime}_i}^*)}\) based on the predicted class \(y^{\prime}_i\) (Fig.~\ref{fig:method_details}c). This approach discretizes the probabilities of samples predicted as different classes using distinct thresholds \(t_{y^{\prime}_i}^*\), effectively compensating for the bias introduced by class imbalance.

\subsection{Class-Specific with Average Thresholded Confidence with Temperature Scaling}

\textcolor{black}{Confidence scores can be first calibrated using TS before being applied in the ATC process, referred to as TS-ATC~\citep{garg2022leveraging}}. Specifically, the confidence score is initially calibrated with TS to obtain \(\tilde{q}_i\), which satisfies Eq.~\ref{eq:TS}, and is then used in the ATC calibration process. By incorporating our class-specific modifications to both TS and ATC, we similarly derive the class-specific variant, CS TS-ATC (Fig.~\ref{fig:method_details}d).

\subsection{Confidence Scores Calculation}

In addition to MCP, various other confidence scores can be derived from logits \(\boldsymbol{z}_i\), such as Entropy~\citep{steinhardt2016unsupervised}, Doctor~\citep{granese2021doctor}, and Energy~\citep{liu2020energy}. While these techniques have not been extensively explored in the context of model evaluation, \textcolor{black}{they have demonstrated improved performance in related tasks like anomaly detection~\citep{chandola2009anomaly} and failure detection~\citep{bernhardt2022failure}}. Therefore, we believe it would be valuable to extend the discussed model calibration methods to incorporate these confidence scores. To achieve this, Entropy and Doctor is calculated with:

\begin{equation}
    q_i^\text{Entropy} = 1 + \sum_j p_{ij} \log_c p_{ij}, \quad q_i^\text{Doctor} = \frac{\sum_j p^2_{ij}-1}{\sum_j p^2_{ij}}.
    \label{eq:entropy}
\end{equation}

The calibration of these two confidence scores is straightforward since they directly involve \(p_{ij}\) in their computation. As a result, TS calibration can be applied by adjusting \(T\) in the softmax process, where \(p_{ij} = \sigma(\boldsymbol{z}_i/T)_j\). In contrast, Energy is calculated as follows:

\begin{equation}
    q_i^\text{Energy} = T\log \sum_j e^{z_{ij}/T}.
    \label{eq:energy}
\end{equation}

It is important to note that we adjust the calculation to ensure alignment with the typical requirements of confidence scores, where a higher value indicates greater confidence. Additionally, we normalize the Doctor and Energy scores using \(\max_i q_i\) and \(\min_i q_i\)\footnote{The calibrated confidence scores are normalized based on the range of uncalibrated confidence scores rather than calibrated scores; otherwise, the calibration process would not affect the estimation results.}, then clip the values between 0 and 1 to give them quantitative significance.

\subsection{Calibration for Group-Wise Metrics}

As previously discussed, calculating group-wise metrics requires calibrated probabilities \(\hat{\boldsymbol{p}}_i\). Thus, we need to derive \(\hat{\boldsymbol{p}}_i\) from the confidence score \(\hat{q}_i\), which involves filling in the probabilities for non-maximum classes. To achieve this, we propose two approaches: 1) Generate the probabilities as if they originated from the TS process. Specifically, we find the parameter that best fits the maximum probabilities (i.e., find \(T^*\) such that \(\hat{q}_i = \max_j (\sigma\left(\frac{\boldsymbol{z}_i}{T^*}\right)_j)\)) on a per-case basis and use it to estimate the non-maximum probabilities. 2) Equally distribute the remaining probability across the non-maximum classes, i.e., \(\frac{1 - q_i}{c - 1}\)~\footnote{For calibrated probabilities under ATC, the non-maximal values are simply set to 0.}. Initial experiments suggest that neither solution consistently outperforms the other. We adopt the second approach for all experiments, as it is more efficient.

Since the class-specific parameters may influence the group-wise metrics estimation of other classes, we optimize each parameter sequentially, starting with the class that has the highest number of predictions and proceeding in descending order. This approach helps minimize the overall estimation error on the validation dataset.

\subsection{MOVAL Ensemble}

\textcolor{black}{Model ensembles, which average the probabilities from multiple models before making decisions, can effectively enhance model performance~\citep{kamnitsas2018ensembles}.} We apply this concept to our task, referring to it as MOVAL Ensemble. In this approach, \(L\) different models are calibrated separately, each yielding \(\hat{\boldsymbol{p}}_i^{(l)}\). The ensemble probabilities are then computed as follows:

\begin{equation}
    \hat{\boldsymbol{p}}_i^{\text{Ensemble}} = \frac{1}{L}\sum_{l=1}^L\hat{\boldsymbol{p}}_{i}^{(l)},
    \label{eq:ensemble}
\end{equation}

\noindent \textcolor{black}{which} are later used on test data to estimate metrics. We select different combinations for various tasks, summarized in Extended Table~\ref{tab:moval_ensemble}.

\subsection{Performance Estimation for Segmentation Tasks}
\label{sec:seg}

We propose extending confidence-based model evaluation methods to segmentation. Unlike classification, segmentation evaluation emphasizes group-wise metrics that compare volumes rather than the entire dataset.

To this end, we slightly adjust the notation and denote the validation dataset as consisting of \( Z \) segmentation cases, represented as \( \mathcal{D}^V = \{ \mathcal{S}_z \}_{z = 1}^Z \), where \( \mathcal{S}_z \) contains all the training pairs for case \( z \), with a total of \( n \) pixels. This is defined as \( \mathcal{S}_z = \{(\boldsymbol{x}_{zi}, y_{zi}) \}_{i = 1}^n \), where \( \boldsymbol{x}_{zi} \) is an image patch sampled from case \( z \) and \( y_{zi} \) is the ground truth class of its central pixel. We denote \( p_{zij} \) as the predicted probability of sample \( \boldsymbol{x}_{zi} \) for the \( j \)'th class, and its calibrated version as \( \hat{p}_{zij} \) (derived from the calibrated confidence score \( \hat{q}_{zi} \)). The estimated F1-Score is then computed using the predicted probabilities, analogous to the common soft DSC loss~\citep{milletari2016v}. Specifically, the real F1-Score of \( \mathcal{D}^V \) for the \( j \)'th class, denoted as \( \Phi^V_j \), is estimated as follows:

\begin{equation}
\Phi'_j = \frac{1}{Z}\sum_{z=1}^{Z}\frac{2\sum_{i=1}^{n}\mathbbm{1}_{zij} \hat{p}_{zij}}{\sum_{i=1}^{n}\mathbbm{1}_{zij} +\sum_{i=1}^{n}\hat{p}_{zij}},
    \label{eq:sDSC}
\end{equation}

\noindent \textcolor{black}{where} now $\mathbbm{1}_{zij} = \mathbbm{1}{(\argmax_k (p_{zik})=j)}$. Notably, we do not calibrate the background class, as its performance remains consistently high due to the relative ease of classifying the majority of background samples. Adjusting the estimation to match this performance is difficult and can interfere with the calibration of foreground classes, as optimizing for the background class was found to destabilize the overall results.

\subsection{Group-Wise Calibration}

\textcolor{black}{Since our approach performs calibration on a per-class basis, a natural concern is the requirement for a sufficient number of samples in each class, especially for minority classes. To address this limitation, we introduce a group-wise calibration strategy that allows class-wise calibration methods to operate at the group level, sharing class-specific parameters across classes and aligning metrics across multiple predicted classes.} This approach, also used in~\citep{li2024learning} for a different application, reduces reliance on validation datasets and proves more robust to various advanced training strategies.

As an example, we consider the CS TS of accuracy based on MCP (i.e., Eq.~\ref{eq:CSTS}). For group-wise calibration, we assume a unique parameter $\widetilde{T}$ for calibrating all the samples belonging to a batch of $B$ classes, $\boldsymbol{b} = [b_1, \cdots, j, \cdots, b_B]^\intercal$. The construction of $\boldsymbol{b}$ can be user-defined, but by default, the classes are grouped based on the number of samples predicted as class $j$, i.e., $N_j$. Recall that the accuracy for the $j$'th class can be estimated as $\alpha^{\prime}_j = \frac{1}{N_j}\sum_{i=1}^{N}\mathbbm{1}_{ij} \max_k(\sigma\left(\frac{\boldsymbol{z}_i}{\widetilde{T}}\right)_k)$. The group-wise calibration of CS TS is optimized with:

\begin{equation}
\widetilde{T}^* = \argmin_{\widetilde{T}}\Big|\sum_{j \in \boldsymbol{b}}N_j(\alpha^{\prime}_j-\alpha^V_j)\Big|.
\label{eq:CSTS_rewrite}
\end{equation}

The estimation of other metrics (e.g., replacing $\alpha'_j$ with $\Phi'_j$ for F1-Score estimation) or the use of alternative estimation methods can be done in very similar ways. Notably, if \( B \) equals 1, group-wise calibration is equivalent to the vanilla CS calibration we introduced beforehand; if \( B \) equals the class number \( C \) when estimating accuracy, it is equivalent to the original class-agnostic calibration. All results in this paper are reported without group-wise calibration unless specified.

\subsection{Optimization}

We optimize the parameters of the MOVAL models using bound-constrained minimization algorithms. By default, we apply the Nelder-Mead simplex algorithm~\citep{gao2012implementing} through the SciPy toolbox~\citep{virtanen2020scipy}. The parameters are initially set to 1.0, and to improve the robustness of the optimization, we perform 5 additional initializations with different starting points if the initial optimization results are unsatisfactory. In practice, we optimize the class-specific parameter for class $j$ only when there are both predicted instances of class $j$ and ground truth instances labeled as class $j$ in the validation dataset. Otherwise, optimization would be meaningless. 

\section{Experiments}
\label{sec:experiments}

\subsection{Experimental Setup}

In our experiments, we first train a classifier $f$ with $\mathcal{D}^{Tr}$, then we make predictions on $\mathcal{D}^{V}$ and optimize the calibration model based on model output and the validation labels. Afterwards, we evaluate our model on $\mathcal{D}^{Te}$ and estimate the quality of the predicted results based on the pre-softmax network outputs $f(\boldsymbol{x}_i^{Te})$.

We train the classification and segmentation models using a well-established process based on ResNet~\citep{he2016deep} and U-Net~\citep{ronneberger2015u}, following previous successful related efforts~\citep{isensee2021nnu, russakovsky2015imagenet, marrakchi2021fighting}. Detailed training configurations are provided in Suppl. Note.

\subsection{Datasets}

\noindent\textbf{CIFAR10-LT.} We resample the CIFAR10 training set~\citep{krizhevsky2009learning} using an exponential class ratio to create an imbalanced dataset with $c=10$ classes and an imbalance ratio of 100, following~\citep{cao2019learning}. The validation and test sets remain unchanged. To introduce domain shifts, we apply 95 synthetic corruptions from CIFAR10-C~\citep{hendrycks2019robustness}. Dataset details are provided in Extended Fig.~\ref{fig:data}, with visualizations shown in Extended Fig.~\ref{fig:data_visual}.

\noindent\textbf{CIFAR100-LT.} We adopt similar strategies to create an imbalanced CIFAR100 dataset, which also has an imbalance ratio of 100. This dataset has $c=100$ classes.

\noindent\textbf{Skin lesion classification.} The skin lesion datasets have $c=7$ classes. We randomly select 70$\%$ of the data from HAM10000~\citep{tschandl2018ham10000} for the training dataset and reserve the rest for validation. We additionally collect 7 publicly available skin lesion datasets which were collected in different institutions including BCN~\citep{combalia2019bcn20000}, VIE~\citep{rotemberg2021patient}, MSK~\citep{gutman2016skin}, UDA~\citep{gutman2016skin},
OTH~\citep{rotemberg2021patient}, D7P~\citep{kawahara2018seven} and PH2~\citep{mendoncca2013ph} mainly following ~\citep{yoon2019generalizable}. \textcolor{black}{The split is performed at the dataset level: training and validation use HAM10000 exclusively, and each of the seven test datasets is an independently collected cohort from a different institution, with no patient or image overlap with the source data.}

\noindent\textbf{Brain lesion MRI segmentation.} We evaluate the proposed methods with brain lesion segmentation based on T1-weighted MR images using data from Anatomical Tracings of Lesions After Stroke (ATLAS)~\citep{liew2022large}. This dataset is for binary segmentation ($c = 2$), with highly imbalanced classes. Images are registered to the MNI152 template~\citep{brett2002problem} with a spacing of $1.00\ \text{mm} \times 1.00\ \text{mm} \times 1.00\ \text{mm}$.

This dataset contains 653 samples collected using 13 different scanners of different field strengths and manufacturers. \textcolor{black}{The split is performed at the scanner level: all 88 cases from the chosen scanner are used exclusively for training and validation, and the remaining 565 cases, acquired on the other 12 scanners, are used exclusively as target test data, with no patient or scanner overlap between source and test.} We create synthetic domains shifts by applying 83 distinct transformations, including spatial, appearance and noise operations to the validation data. We show some visualization of this dataset in Extended Fig.~\ref{fig:example}. The detailed dataset statistics and transformation implementation are summarized in Extended Table~\ref{tabs} and Extended Table~\ref{tabteaug}. 

\noindent\textbf{Prostate MRI segmentation.} We also evaluate on the task of cross-site prostate segmentation based on T2-weighted MR images following~\citep{liu2020saml}. This is also a binary segmentation task ($c = 2$). We resample the images to a common resolution of $0.80\ \text{mm} \times 0.80\ \text{mm} \times 1.50\ \text{mm}$.

We use 30 cases collected by~\citep{bloch2015nci} as the source domain. For the target domain, we utilize five datasets described in \citep{lemaitre2015computer, bloch2015nci, litjens2014evaluation}, which were collected from various institutions using scanners of different field strengths and manufacturers. \textcolor{black}{As with the brain lesion data, the split is at the institution/scanner level.} Synthetic domain shifts are generated using the same 83 transformations applied to brain lesions.

\noindent\textbf{Cardiac MRI segmentation.} We utilize cardiac MR images from the Multi-Centre, Multi-Vendor, and Multi-Disease Cardiac Segmentation (M$\&$Ms) Challenge~\citep{campello2021multi}. The task is to segment the images into three classes ($c = 3$), including the left and right ventricle cavities and the left ventricle myocardium. The image volume is anisotropic and is therefore evaluated slice by slice, that is, in 2D. We resample the images to a common resolution of $1.25\ \text{mm} \times 1.25\ \text{mm} \times 10.00\ \text{mm}$.

This dataset contains short-axis cardiac MR images from 345 samples across 5 different sites. Following the clinical protocol, we segment two frames of the MRI, including the end-diastolic (ED) and end-systolic (ES) phases, totaling 690 images. We use data from one site (190 images) as the source domain and validate the trained model on other domains and synthesized domain shifts. \textcolor{black}{The split is again at the site level.} Compared with synthesized domain shifts in 3D segmentation, we exclude those that operate in the z-axis due to its anisotropic nature, resulting in 65 test conditions. We report the average estimation error across all three classes.

\subsection{Class-Wise Performance Alignment}

\begin{figure*}[ht]
  \centering
   \includegraphics[width=0.93\linewidth]{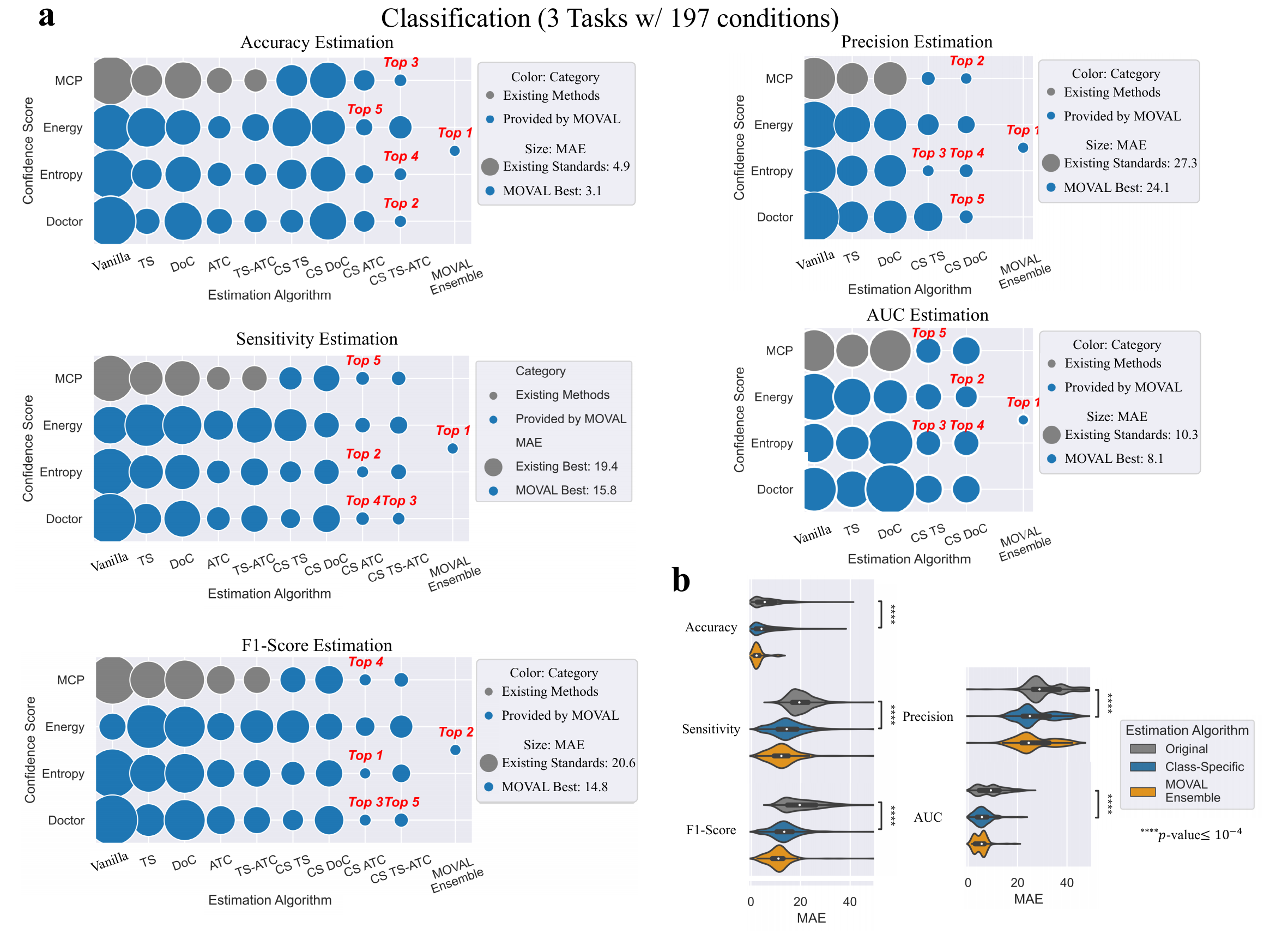}
   \caption{\textcolor{black}{Class-wise calibration improves the estimation results for all metrics in classification tasks.}
   \textbf{(a)}: The mean absolute error (MAE) between real metrics and estimated results based on different estimation methods for classification tasks. The size of the circles represents the MAE, with smaller circles indicating better performance. We report the average MAE for various tasks here. 
   \textbf{(b)}: MAE comparison between the original estimation methods and our proposed class-specific variants for classification tasks. Paired t-tests were used for statistical analysis.
   }
   \label{fig:cls_results}
\end{figure*}

We first demonstrate the effectiveness of class-wise calibration by extending existing methods including TS, DoC, ATC and TS-ATC (which is a combination of TS and ATC) to the variants where performances are aligned in class-wise manner with class-specific (CS) parameters. These proposed variants are denoted as CS TS, CS DoC, CS ATC and CS TS-ATC, respectively.

We use CIFAR10-LT to illustrate the calibration process for CS TS, as shown in Fig.~\ref{fig:method}b. TS adjusts confidence based on validation performance to align global accuracy with global confidence. However, this is achieved by sacrificing class-wise performance, where the majority classes remain overconfident (i.e., the mean confidence is higher than the actual accuracy), while the minority classes are underconfident.

We evaluate the estimation results across 3 classification tasks, covering 197 types of domain shifts, which include both synthetic shifts for CIFAR~\citep{krizhevsky2009learning} and real-world acquisition shifts for skin lesion classification where images are gathered from various sites~\citep{tschandl2018ham10000} (Fig.~\ref{fig:ensemble}a, Extended Fig.~\ref{fig:detailed_cls}a). To note, we resample the CIFAR training dataset to create a long-tailed distribution, making the evaluation more realistic, as this reflects the nature of real-world datasets~\citep{liu2019large, cao2019learning}. The models are based on a standardized version of ResNet~\citep{he2016deep}. In order to evaluate the estimation results, we calculate the Mean Absolute Error (MAE) between the predicted and the real accuracy and report the overall results as the mean MAE across three tasks (Fig.~\ref{fig:cls_results}a). Our proposed CS variants significantly improve the estimation results, reducing the overall MAE from 6.64 to 5.81 (Fig.~\ref{fig:cls_results}b, Extended Fig.~\ref{fig:detailed_cls}b, paired t-test, $p=1.01\times10^{-48}$). Specifically, the CS variants reduce the MAE for TS from 7.33 to 7.06 ($p=0.02$), from 10.40 to 9.76 for DoC ($p=1.71\times10^{-9}$), from 4.38 to 3.50 for ATC ($p=6.42\times10^{-15}$) and from 4.45 to 2.93 for TS-ATC ($p=1.94\times10^{-42}$). These results clearly highlight the importance of addressing class imbalance in the design of performance-prediction methods to mitigate biases introduced by the data.

Another advantage of class-wise alignment is its robustness to label shifts in test datasets, as it tracks performance separately for each class. This is especially beneficial when there are more samples from minority classes at test time. For example, when deploying the model on the CIFAR10-LT dataset, which includes only the two most minority classes, CS variants can reduce the MAE from 27.0 to 21.7 (Extended Table~\ref{tab:cls_acc_imbalanced}, paired t-test, $p=2.40\times10^{-120}$). Additionally, it is clear that the proposed CS variants provide more accurate performance estimation on the subsets defined by the model's predictions (Fig~\ref{fig:method}b).

\subsection{Multi-Metric Estimation}

The MOVAL framework allows to estimate other metrics beyond accuracy, delivering a thorough assessment based solely on the network outputs. Those metrics like the F1-Score are typically calculated on a per-class basis. Consequently, we compute the MAE for each individual class and report the MAE for a case along with its average MAE across all classes (Extended Fig.~\ref{fig:full_res}a).

Class-wise performance alignment naturally suits the estimation of group-wise metrics, as demonstrated with CIFAR10-LT in Fig.~\ref{fig:method}a. Existing calibration methods focused on accuracy can reduce overconfidence and, in turn, lower estimation error for group-wise metrics~\citep{nannyml}, but they do not achieve optimal results for those metrics. Although our CS variants may not fully align performance for precision and AUC, they greatly enhance calibration, thereby reducing estimation errors during testing.

Similarly to the accuracy estimation evaluation, we report results across three classification tasks (Fig.~\ref{fig:cls_results}a). Our CS variants show clear advantage than existing metrics (Fig.~\ref{fig:cls_results}b, Extended Fig.~\ref{fig:detailed_cls}b), reducing the MAE from 21.25 to 13.95 for F1-Score (paired t-test, $T(3151)=79.16, p\leq10^{-4}$), from 20.40 to 14.96 for sensitivity ($T(3151)=46.89, p\leq10^{-4}$), from 31.23 to 26.99 for precision ($T(1575)=64.76, p\leq10^{-4}$), and from 9.19 to 6.04 for AUC ($p=1.25\times10^{-125}$).

\subsection{MOVAL for Segmentation Tasks}

\begin{figure*}[ht]
  \centering
   \includegraphics[width=0.93\linewidth]{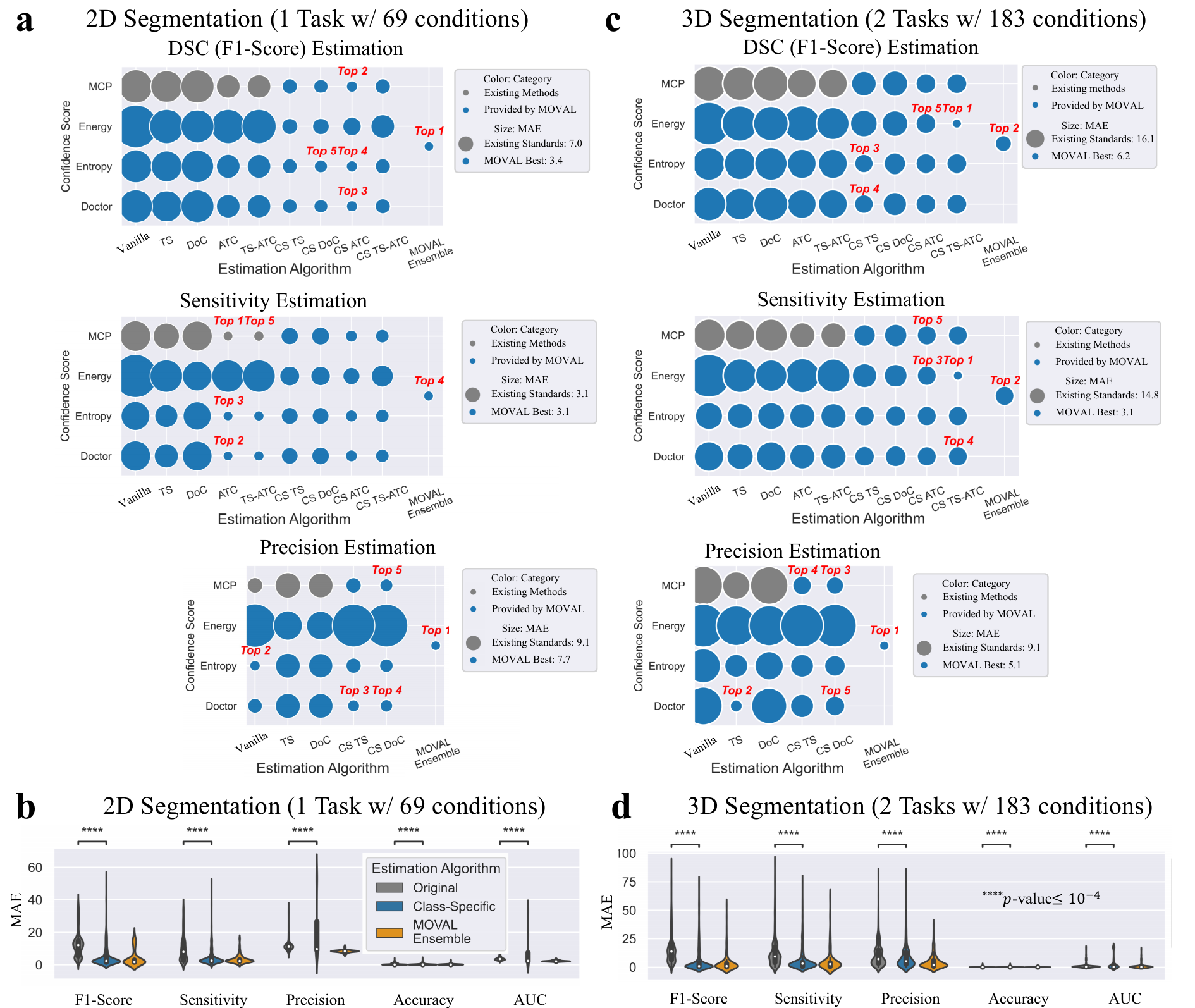}
   \caption{\textcolor{black}{Class-wise calibration is even more crucial for segmentation tasks.}
   \textbf{(a, c)}: MAE between real metrics and estimated results from different methods for 2D and 3D segmentation tasks.
   \textbf{(b, d)}: MAE comparison between the original estimation methods and proposed class-specific variants for 2D and 3D segmentation tasks. Statistical tests are conducted using paired t-tests.
   }
   \label{fig:seg_results}
\end{figure*}

\begin{figure*}[ht]
  \centering
   \includegraphics[width=\linewidth]{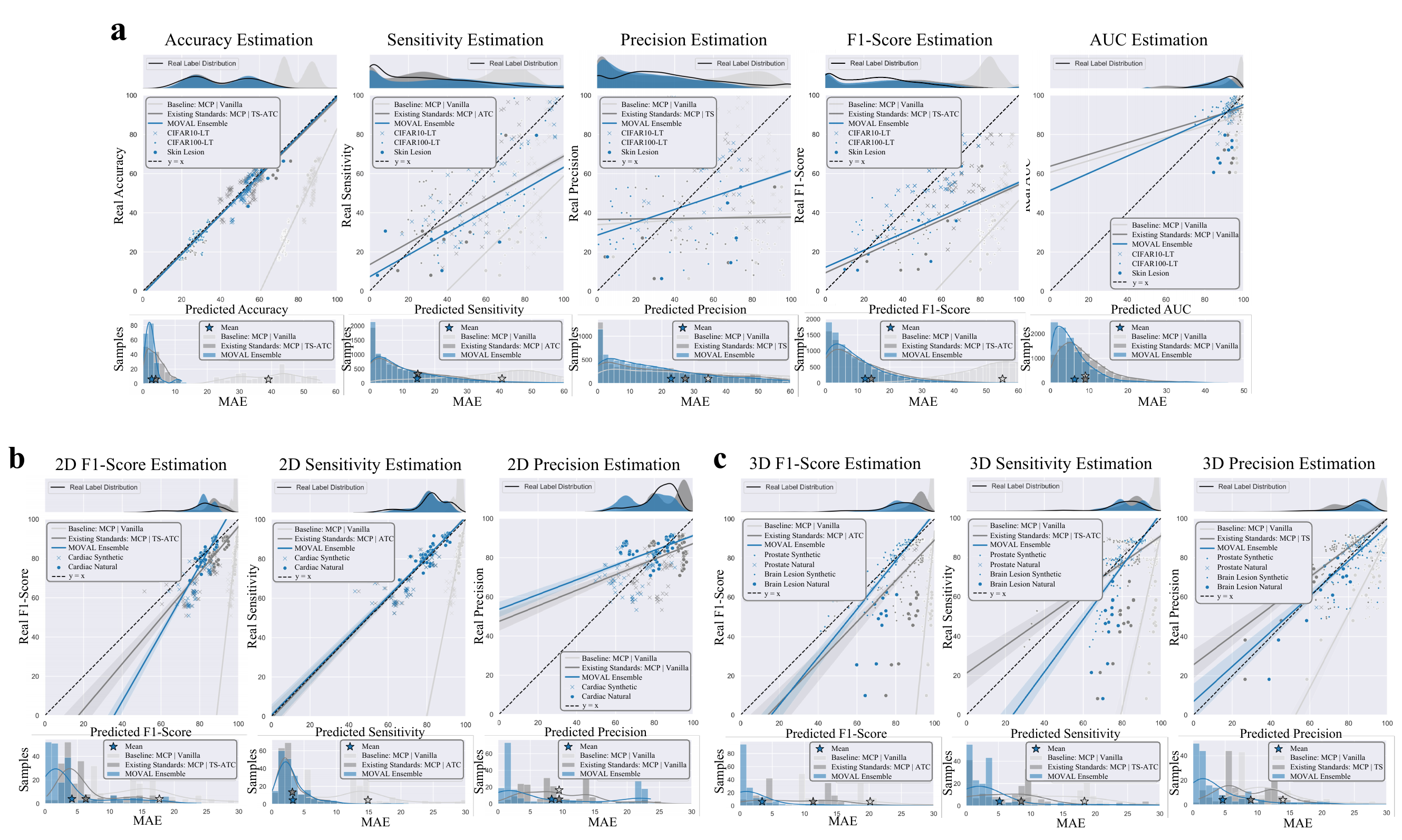}
   \caption{\textcolor{black}{Agreement between predicted and real metrics for different tasks.} For each subfigure: Top: Predicted and real label distribution. We benchmark MOVAL ensemble (blue) against MCP baselines (light gray) and the best-performing existing method (dark gray). Middle:  Predicted versus real metrics. Each point represents the model's performance on unseen target data under specific conditions. Bottom: Histogram of MAE for different estimation methods. 
   \textbf{(a)}: Classification tasks.
   \textbf{(b)}: 2D segmentation tasks. 
   \textbf{(c)}: 3D segmentation tasks. 
   }
   \label{fig:scatter_results}
\end{figure*}

The MOVAL framework supports the evaluation of both 2D and 3D segmentation tasks. It extends to semantic segmentation by treating it as pixel-level dense classification. A key distinction between evaluating segmentation and classification is that segmentation involves comparing 2D slices or 3D volumes, rather than assessing the entire dataset. With this adjustment, all the previously mentioned methods can be easily adapted for segmentation tasks.

We evaluate 2D segmentation on the task of cardiac MRI~\citep{campello2021multi} segmentation and 3D segmentation on prostate MRI~\citep{liu2020saml} and brain lesion MRI~\citep{liew2022large} segmentation. For all segmentation tasks, we apply transformations as synthetic domain shifts (Extended Table~\ref{tabteaug}) and collect images from different scanners to represent real domain shifts. The models are built upon a standardized version of 3D U-Net~\citep{ronneberger2015u}.

We present the evaluation results of the average MAE between real and synthesized scenarios (Fig.~\ref{fig:seg_results}a, Fig.~\ref{fig:seg_results}c). We omit accuracy and AUC evaluations, as these metrics often approach 100$\%$ for segmentation tasks (Extended Fig.~\ref{fig:detailed_seg_acc}c, Extended Fig.~\ref{fig:detailed_seg_acc}d). Our CS variants demonstrate even greater advantages over existing standards compared to classification results (Fig.~\ref{fig:seg_results}b, Fig.~\ref{fig:seg_results}d). This may be due to previous model calibration methods being heavily biased toward the background class, a consequence of class imbalance.

We focus on evaluating the F1-Score (i.e., DSC), the most commonly used metric for segmentation tasks (Fig~\ref{fig:ensemble}c, Extended Fig.~\ref{fig:detailed_seg}a, Extended Fig.~\ref{fig:detailed_seg}b). CS variants greatly enhance confidence-based estimation of the F1-Score, lowering the MAE from 12.88 to 4.73 for 2D segmentation (paired t-test, $p=1.03\times10^{-227}$) and from 15.41 to 4.69 for 3D segmentation ($T(2927)=72.55$, $p\leq10^{-4}$). Similar conclusions also apply to sensitivity and precision (Extended Fig.~\ref{fig:full_res}b, Extended Fig.~\ref{fig:full_res}c). For sensitivity, CS variants decrease the MAE from 9.55 to 4.31 for 2D segmentation ($p=2.57\times10^{-145}$) and from 12.78 to 5.75 for 3D segmentation($p=1.59\times10^{-320}$). For precision, they reduce the MAE from 11.25 to 9.36 for 2D segmentation ($p=1.94\times10^{-47}$) and from 9.38 to 5.76 for 3D segmentation ($p=2.97\times10^{-29}$). Note that we exclude energy-based evaluation for precision due to its consistently poor performance in segmentation tasks.

\subsection{Confidence Scores Ensemble}

\begin{figure*}[ht]
  \centering
   \includegraphics[width=\linewidth]{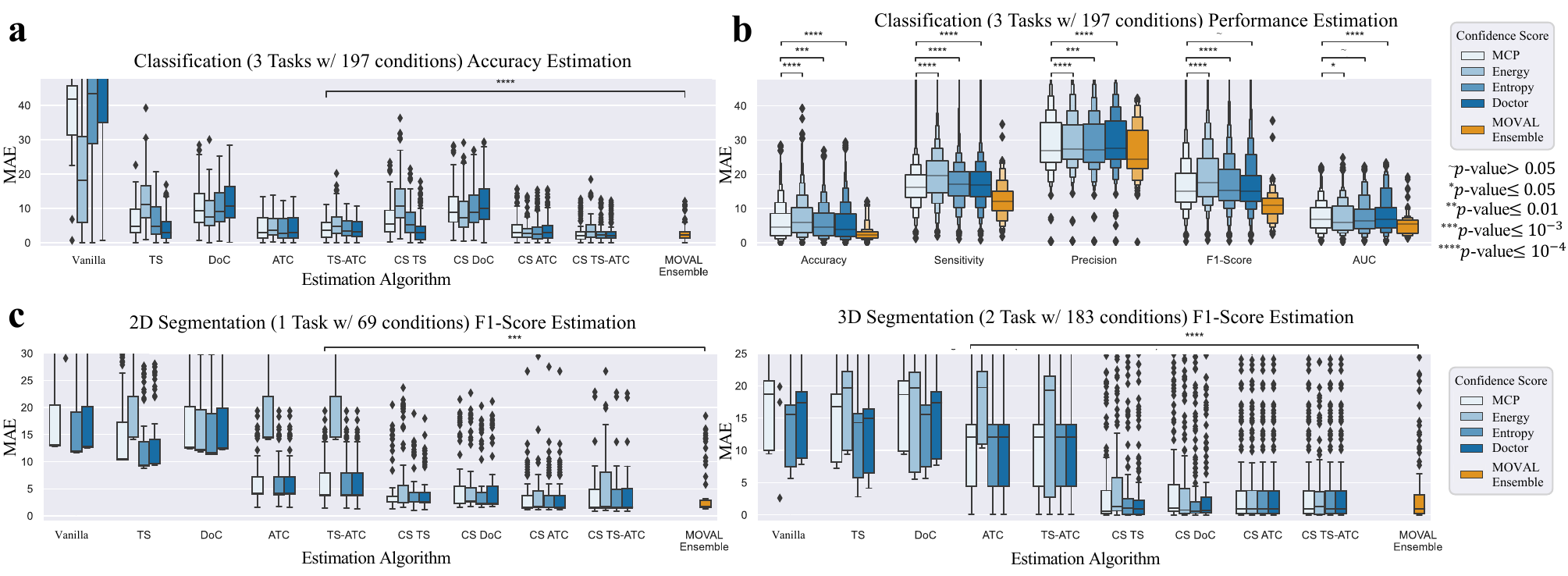}
   \caption{MOVAL Ensemble achieves state-of-the-art performance.
   \textbf{(a, c)}: MAE comparison between proposed algorithms and existing standards (Best-performing existing method versus MOVAL Ensemble, paired t-test,  $p=1.05\times10^{-7}$ for accuracy estimation of classification tasks, $p=3.54\times10^{-4}$ for F1-Score estimation of 2D segmentation tasks, $p=4.28\times10^{-27}$ for F1-Score estimation of 3D segmentation tasks, see the results of other metrics in Extended Fig.~\ref{fig:full_res} and the results of individual datasets metrics in Extended Fig.~\ref{fig:detailed_cls} and Fig.~\ref{fig:detailed_seg}).
   \textbf{(b)}: MAE comparison between estimation algorithms based on MCP and other confidence scores for classification tasks (see Extended Fig.~\ref{fig:confidence} for other tasks).
   }
   \label{fig:ensemble}
\end{figure*}

MOVAL enhances existing model calibration methods by integrating confidence scores beyond MCP and utilizing ensemble techniques for improved results. While traditional model evaluation methods rely on MCP for its simplicity and straightforward quantitative meaning, MOVAL advances this by integrating other confidence scores, such as Entropy~\citep{steinhardt2016unsupervised}, Doctor~\citep{granese2021doctor}, and Energy~\citep{liu2020energy}. These confidence scores have shown enhanced outlier detection capabilities due to their improved ability to capture information from non-maximum predictions (Fig.~\ref{fig:method}a). Additionally, akin to model ensembles in discriminative models, we investigate the ensemble of the best methods to deliver robust and effective solutions.

We begin by investigating the discrepancies between different confidence scores and estimation algorithms, as model ensembles are effective only when the methods exhibit substantial diversity~\citep{kamnitsas2018ensembles, ganaie2022ensemble}. We find that MCP yields significantly different results compared to other confidence scores in 13 out of 15 conditions for classification tasks (Fig.~\ref{fig:ensemble}b), 8 out of 9 conditions for 2D segmentation, and 9 out of 9 conditions for 3D segmentation (Extended Fig.~\ref{fig:confidence}b). Additionally, we observe that different existing estimation algorithms also display clear differences under varying conditions (Extended Fig.~\ref{fig:confidence}a).

Next, we combine three estimation methods into an ensemble, known as MOVAL Ensemble, and apply it to all the test conditions previously mentioned. Notably, we select different combinations for various tasks and metrics, with most of the chosen methods being from our proposed CS variants (Extended Table~\ref{tab:moval_ensemble}). The resulting MOVAL Ensemble consistently ranks among the top performers across all tasks and metrics, achieving the best results in 7 out of 11 scenarios (Fig.~\ref{fig:cls_results}a, Fig.~\ref{fig:seg_results}a, Fig.~\ref{fig:seg_results}c). Compared to state-of-the-art methods, the MOVAL Ensemble demonstrates significant advantages in most cases (Fig.~\ref{fig:ensemble}a, Fig.~\ref{fig:ensemble}c, Extended Fig.~\ref{fig:full_res}a, Extended Fig.~\ref{fig:full_res}b, Extended Fig.~\ref{fig:full_res}c). 

For classification tasks, compared to the best-performing existing methods, MOVAL Ensemble reduces the MAE from 4.86 to 3.09 for accuracy (paired t-test, $p=1.05\times10^{-7}$), from 20.63 to 14.81 for F1-Score ($p=6.04\times10^{-18}$), from 19.37 to 15.84 for sensitivity ($p=3.54\times10^{13}$), from 27.32 to 24.10 for precision ($p=1.43\times10^{-6}$), and from 10.33 to 8.08 for AUC ($p=8.18\times10^{-7}$). Specifically, we observe that MOVAL Ensemble significantly enhances performance especially when there are fewer classes (e.g., 10 classes, Extended Fig.~\ref{fig:detailed_cls}a). For example, MOVAL Ensemble reduces the MAE from 18.85 to 9.45 for sensitivity estimation on CIFAR10-LT (Extended Table~\ref{tab:cls_sensitivity_precision},  $p=1.53\times10^{-93}$).

For segmentation tasks, MOVAL Ensemble consistently enhances F1-Score estimation accuracy across various tasks (Extended Fig.~\ref{fig:detailed_seg}a). Specifically, compared to the best-performing existing methods, MOVAL Ensemble reduces the MAE from 7.05 to 3.44 for 2D segmentation (paired t-test, $p=3.54\times10^{-4}$) and from 16.13 to 7.74 for 3D segmentation ($p=4.28\times10^{-27}$). For precision estimation, MOVAL Ensemble decreases the MAE from 9.06 to 7.68 for 2D segmentation ($p=8.54\times10^{-7}$) and from 9.09 to 5.12 for 3D segmentation ($p=1.53\times10^{-10}$). MOVAL Ensemble also proves effective for sensitivity estimation in 3D segmentation, reducing the MAE from 14.76 to 10.43 ($p=1.09\times10^{-4}$). Although the improvement in sensitivity for 2D segmentation is not significant (3.07 versus 3.09, $p=0.80$), MOVAL Ensemble demonstrates notably better results for the most minority class (i.e., left cavities segmentation), reducing the MAE from 2.93 to 2.87 ($p=1.4\times10^{-2}$). In summary, MOVAL Ensemble serves as a reliable tool for estimating key metrics such as F1-Score, sensitivity, and precision in segmentation tasks.

To gain a deeper understanding of the estimation results, we examine the actual metrics and estimation outcomes of MOVAL Ensemble, the best-performing existing methods and the MCP baselines (Fig.~\ref{fig:scatter_results}a, Fig.~\ref{fig:scatter_results}b, Fig.~\ref{fig:scatter_results}c). The predicted performance by the MCP baseline consistently exceeds the actual performance across all metrics due to the common over-confidence in neural networks~\citep{guo2017calibration}. Existing standards can mitigate the discrepancy between estimated and actual metrics, while MOVAL Ensemble performs better at estimating the underlying label distribution (Top Fig.~\ref{fig:scatter_results}a). The estimation results of MOVAL Ensemble are closer to the $"\text{predicted metric}=\text{real metric}$" line over a broad range, indicating that it produces less estimation error across varying levels of difficulty (Middle Fig.~\ref{fig:scatter_results}c). The distribution of MAE produced by MOVAL Ensemble is more concentrated at lower values, indicating that it achieves smaller estimation errors in a greater number of cases (Bottom Fig.~\ref{fig:scatter_results}c).

\subsection{Robustness under Training Conditions}

\begin{figure*}[ht]
  \centering
   \includegraphics[width=\linewidth]{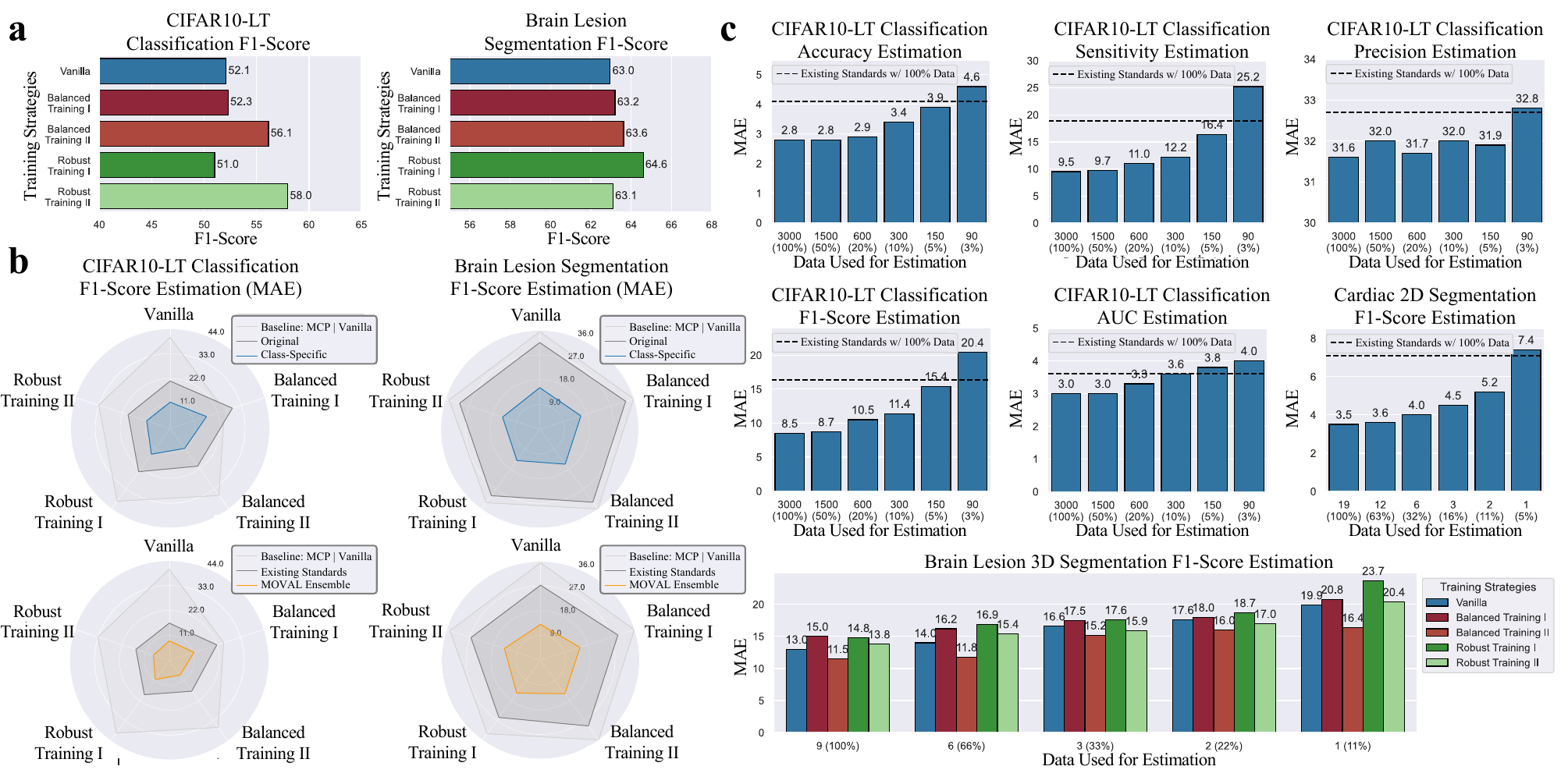}
   \caption{MOVAL Ensemble maintains stability in realistic settings.
   \textbf{(a)}: Classification and segmentation results when models are trained with advanced training strategies designed for class imbalance or domain generalization (Classification, Balanced Training \uppercase\expandafter{\romannumeral1}: label-distribution-aware margin (LDAM) loss~\citep{cao2019learning}, Balanced Training \uppercase\expandafter{\romannumeral2}: deferred re-weighting (DRW) training schedule~\citep{cao2019learning}, Robust Training \uppercase\expandafter{\romannumeral1}: cutout~\citep{devries2017improved}, Robust Training \uppercase\expandafter{\romannumeral2}: RandAugment~\citep{cubuk2020randaugment}; Segmentation, Balanced Training \uppercase\expandafter{\romannumeral1}: asymmetric large margin loss~\citep{li2020analyzing}, Balanced Training \uppercase\expandafter{\romannumeral2}: asymmetric focal loss~\citep{li2020analyzing}, Robust Training \uppercase\expandafter{\romannumeral1}: global intensity non-linear augmentation (GIN) technique~\citep{ouyang2022causality}, Robust Training \uppercase\expandafter{\romannumeral2}: mixup~\citep{zhang2017mixup}).
   \textbf{(b)}: Top: MAE comparison between the original estimation methods and our proposed class-specific variants with various training strategies. Bottom: MAE comparison between best-performing existing method and MOVAL Ensemble with various training strategies.
   \textbf{(c)}: MAE achieved by MOVAL Ensemble when optimized with different amounts of validation data.
   }
   \label{fig:ablation}
\end{figure*}

We evaluate the compatibility of MOVAL Ensemble with models trained using advanced strategies, specifically Balanced Training to address class imbalance~\citep{cao2019learning, li2020analyzing} and Robust Training to improve model generalization~\citep{devries2017improved, cubuk2020randaugment, ouyang2022causality, zhang2017mixup}. These strategies are commonly used in real-world applications (Suppl. Note) and are applied to CIFAR10-LT and brain lesion segmentation. Most strategies improve model performance on unseen test data across various domain shifts (Fig.~\ref{fig:ablation}a).

We then apply the MOVAL framework to these models to evaluate its effectiveness in these realistic settings (Fig.~\ref{fig:ablation}b). Previous conclusions remain valid for most scenarios involving advanced training strategies. For example, the proposed CS variants yield better results than the original estimation algorithms for F1-Score (Top Fig.~\ref{fig:ablation}c). Additionally, MOVAL Ensembles consistently outperform best-performing existing methods by a substantial margin when models are trained using various strategies (Bottom Fig.~\ref{fig:ablation}c). 

In practice, certain balanced training strategies can significantly alter the distribution of predicted probabilities and complicate the confusion matrix. This creates challenges for performance estimation, particularly accuracy estimation, since calibration methods rely on these probabilities and metrics. To address this, we propose group-wise calibration, which aligns accuracy across groups of samples from different classes which share similar statistics instead of focusing on individual classes. This approach can stabilize accuracy estimation in advanced training settings (Extended Table~\ref{tab:cls_acc_bl},~\ref{tab:cls_acc_bl_batch}).

\subsection{Stability with Varying Amounts of Data}

We evaluate MOVAL Ensemble with varying amounts of available validation data to provide practical application guidelines. We optimize MOVAL Ensemble for default-trained models on CIFAR10-LT and cardiac segmentation with different data quantities, and for all advanced training strategies in brain lesion segmentation. This simulates a realistic setting, and we present the average results from five runs.

We find that MOVAL Ensemble is not sensitive to the amount of validation data, performing well even with limited data (Fig.~\ref{fig:ablation}c). For classification tasks like CIFAR10-LT, it requires as little as 5$\%$ of the validation data to achieve performance comparable to best-performing existing methods for most metrics. Approximately 60 samples per case are needed to obtain the most stable results. A similar conclusion applies to segmentation tasks, where 12 slices are required for 2D segmentation and 6 volumes for 3D segmentation to achieve stable results. \textcolor{black}{MOVAL Ensemble performs best with Balanced Training II (asymmetric focal loss)~\citep{li2020analyzing}}, likely due to reduced variance in predicted probabilities for background samples, minimizing probability shifts under domain changes and enhancing estimation effectiveness.

Additionally, as previously mentioned, we provide group-wise calibration for classification, which assumes that classes with similar statistics should share similar optimal calibration parameters, thereby reducing the number of parameters. This approach could further decrease the need for validation samples (Extended Fig.~\ref{fig:group_cal}).

\subsection{Informative Indicators at Different Levels}

\begin{figure*}[ht]
  \centering
   \includegraphics[width=\linewidth]{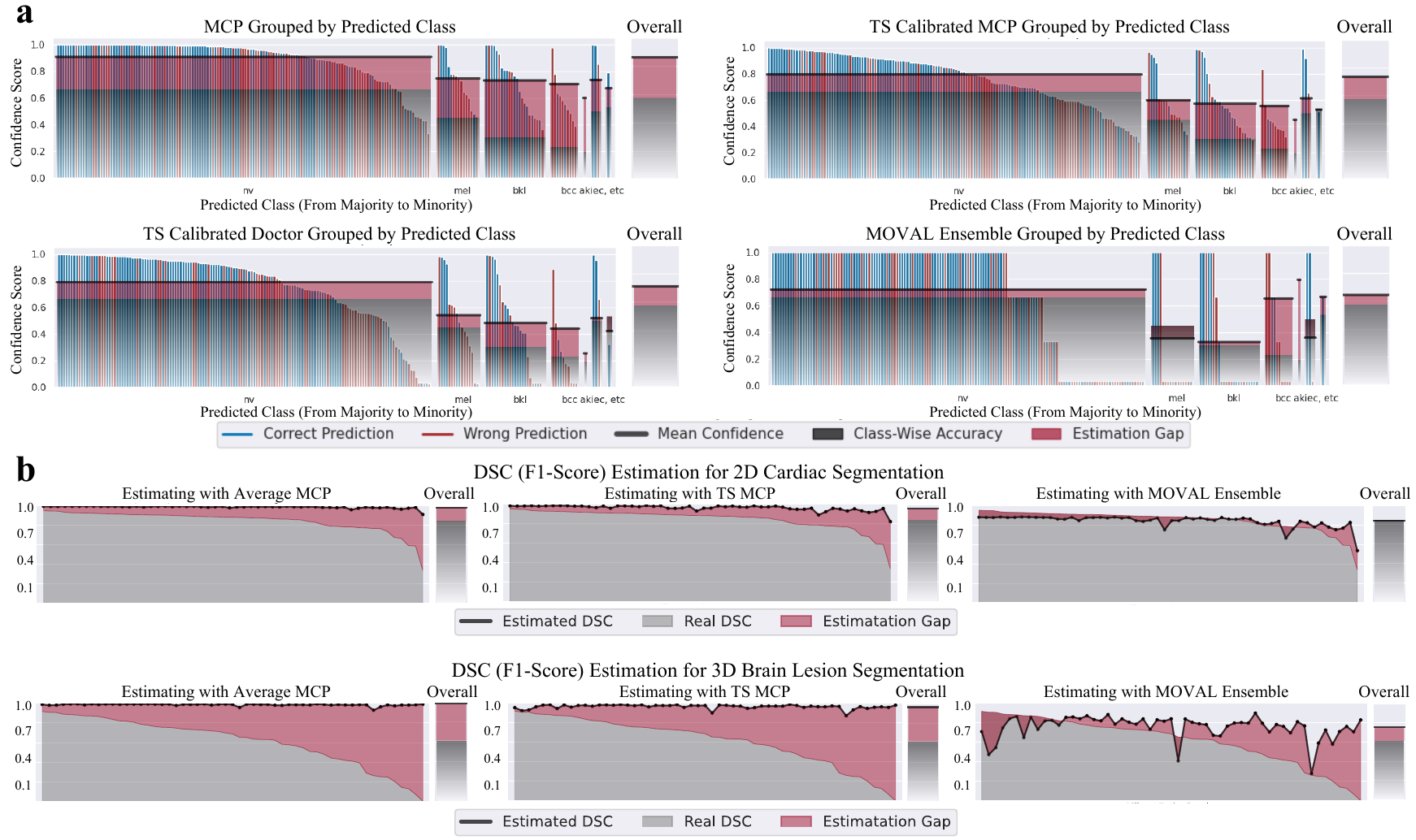}
   \caption{Fine-grained performance indicators at class and instance levels.
   \textbf{(a)}: Alignment of confidence scores and class-wise correctness for skin lesion classification.
   \textbf{(b)}: Top: Alignment of instance-wise DSC estimation and real DSC for 2D cardiac segmentation. Bottom: Alignment for 3D brain lesion segmentation.
   }
   \label{fig:probability}
\end{figure*}

Lastly, we test if the calibrated confidence of MOVAL Ensemble can be informative except the use of estimating global metrics. Specifically, the confidence score is calibrated to match accuracy for skin lesion classification and to match the F1-Score for cardiac and brain lesion segmentation.

MOVAL provides more precise feedback at the class-wise level (Fig.~\ref{fig:probability}a). By evaluating model performance on a per-class basis, the MOVAL Ensemble not only offers an overall estimation advantage but also demonstrates improved alignment with class-wise performance. This allows for a more accurate assessment of sample groups and facilitates comparisons between different groups.

MOVAL also provides more precise feedback at the instance-wise level for segmentation tasks (Fig.~\ref{fig:probability}b). By optimizing MOVAL parameters using averages from various cases, we can also evaluate performance for individual cases. While existing accuracy-calibrated methods tend to be overly optimistic for segmentation tasks, MOVAL Ensemble offers a more accurate reflection of segmentation results across different cases.

\begin{figure*}[ht]
  \centering
   \includegraphics[width=\linewidth]{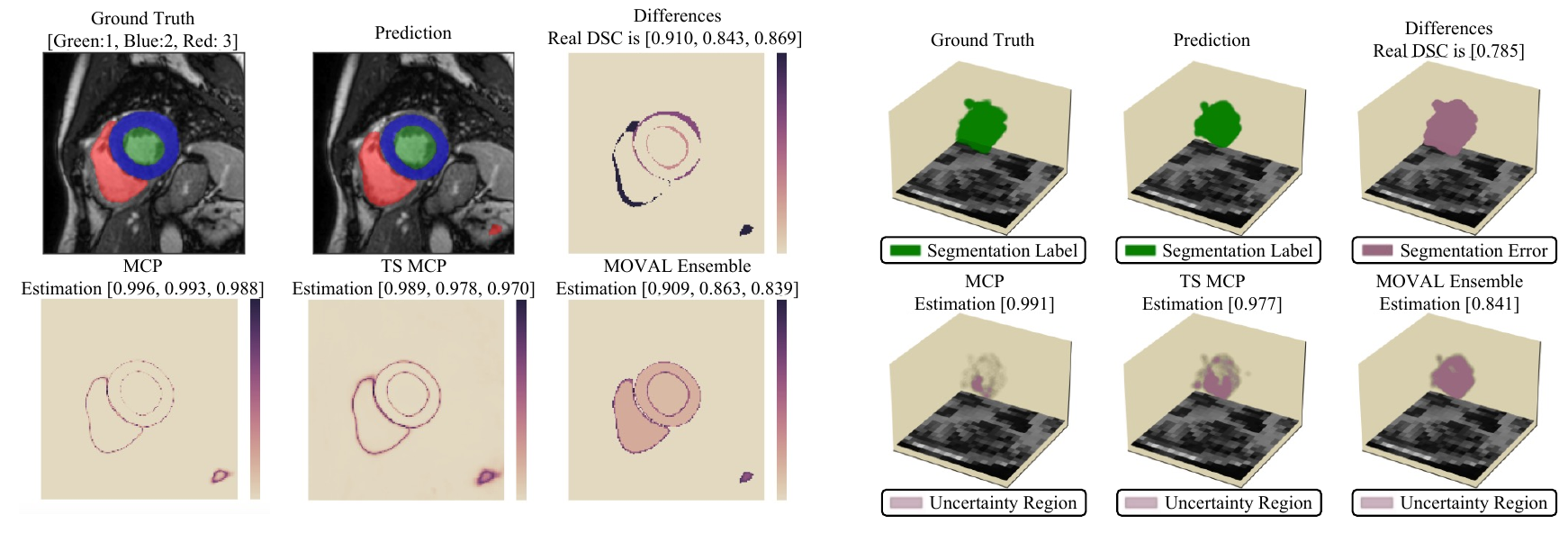}
   \caption{Fine-grained performance indicators at pixel level.
   Left: Visualization of calibrated confidence scores for 2D cardiac segmentation. Right: 3D scatter plots of segmentation error and uncertainty regions from calibrated confidence scores.
   }
   \label{fig:probability_pixel}
\end{figure*}

MOVAL provides prompt performance feedback at the pixel-wise level (Fig.~\ref{fig:probability_pixel}). We can assess model performance using the calibrated confidence map. For example, MOVAL Ensemble can highlight classes with relatively poor performance (Left Fig.~\ref{fig:probability_pixel}). It also offers quantitative insights into uncertain regions and emphasizes areas that need closer attention (Right Fig.~\ref{fig:probability_pixel}). \textcolor{black}{This behavior arises because MOVAL Ensemble integrates multiple confidence scores, whose combined confidence values tend to concentrate near 0 and 1 (see Extended Fig.~\ref{fig:conf_kde}), potentially making it easier to detect anomalies without carefully crafting a threshold.}

\section{Discussion and Conclusion}
\label{sec:discussion}

\textcolor{black}{Reliable deployment is crucial for fully realizing the potential of modern machine learning models in real-world applications. Efforts to regulate the development of machine learning models to support safer deployment have been on the rise~\citep{rivera2020guidelines, vasey2022reporting}, yet a standardized tool to easily monitor performance without labels is still lacking. To fill this gap, we introduce MOVAL, a new confidence-based model evaluation tool that estimates model performance without needing ground truth. By extending confidence-based approaches to a variety of confidence scores, metrics, and tasks, we achieve promising results in diverse scenarios, from cross-site skin lesion classification to cross-scanner MRI segmentation. We believe this tool can enhance existing machine learning workflows with its easy-to-use and transparent design, as a practical post-training monitoring and performance-estimation aid that can inform, rather than by itself guarantee, safe deployment of machine learning in real-world settings. For instance, it can help determine whether a pre-trained model should be deployed by evaluating if its estimated performance on target data meets the acceptance criteria for the deployment environment, complementing rather than replacing standard validation and regulatory processes.}

The key feature of MOVAL is its ability to track a wide range of complementary and informative metrics, including accuracy, F1-Score, sensitivity, precision, and AUC, all within a unified framework. Confidence scores have shown notable accuracy results for balanced benchmark datasets like CIFAR10, but they do not address class imbalance, a common issue in medical imaging databases. MOVAL's unique contribution is its extension of the standard calibration process to incorporate class-wise alignment. This approach not only reduces calibration bias caused by class imbalance but also naturally adapts the framework to estimate group-wise metrics beyond accuracy (Fig.~\ref{fig:method}). Furthermore, unlike existing methods, class-wise calibration enables MOVAL to track performance at the class level, making it less sensitive to variations in label distribution during testing compared to the training distribution.

MOVAL is highly versatile and can be easily applied to a variety of tasks, including classification, 2D segmentation, and 3D segmentation. Through extensive evaluation across six tasks and 6,572 test scenarios (Fig.~\ref{fig:cls_results}, Fig.~\ref{fig:seg_results}), we demonstrate that MOVAL offers significant advantages over existing standards. Designed with distinct modular components, the framework is flexible enough to integrate additional confidence scores or other performance estimation methods. In the future, we plan to include more recent successes in confidence-based accuracy estimation, such as Agreement-on-the-Line~\citep{baek2022agreement}, ATC-distance~\citep{roschewitz2023distance}, Confidence Optimal Transport (COT)~\citep{lu2023characterizing} and Weighted Flips (WF)~\citep{press2024entropy} into our framework.

Building on our successful approaches, we introduce MOVAL Ensemble, which combines the strengths of various estimation methods by averaging the calibrated probabilities. This ensemble achieves top performance in most of the scenarios we tested. Specifically, compared to the best-performing existing methods, it reduces classification accuracy estimation error by 36.4$\%$, 2D segmentation F1-Score estimation error by 51.2$\%$, and 3D segmentation F1-Score estimation error by 52.0$\%$. While we provide MOVAL Ensemble as a comprehensive solution, users can also customize it by selecting the confidence scores they wish to combine for their specific tasks.

We illustrate the utility and effectiveness of MOVAL Ensemble in realistic settings and use cases. Our findings indicate that MOVAL Ensemble performs well with various advanced model training strategies designed to enhance model robustness against class imbalance and domain shifts. Despite these strategies significantly altering model behavior, our method consistently produced stable estimation results across most scenarios. We also demonstrate that MOVAL Ensemble has minimal requirements for the validation set and offer practical suggestions for designing experiments in realistic settings (Fig.~\ref{fig:ablation}). Furthermore, we show that calibrated confidence can be valuable for fine-grained performance analysis (Fig.~\ref{fig:probability}).

MOVAL is primarily designed to estimate model performance under acquisition shifts in medical imaging, where images are typically obtained under varying conditions~\citep{castro2020causality}. While it can be used in other domain shift scenarios, users should ensure the following for optimal performance (though not all can be avoided): 1) No annotation shifts between the validation and test sets, meaning the labeling concept remains consistent across datasets. 2) No population shifts between validation and test sets, meaning there should be no population-related covariate shifts in the input features. This requires ensuring that samples are collected from similar regions and demographics, with comparable disease patterns or evolving environments. \textcolor{black}{3) Although MOVAL is not highly sensitive to prevalence shifts (e.g., differing class ratios between validation and test), performance is generally optimal when these ratios align because the learned calibration parameters are closely tied to the confusion matrix, which is more likely to remain correlated between validation and test with identical prevalence.} 

\textcolor{black}{The precision estimate is inherently limited as it can only be calibrated from positive predictions, ignoring the informative but large pool of false negatives. This makes it less reliable than sensitivity or F1-Score, which leverage a wider range of prediction data for calibration. Consequently, MOVAL should not be the sole metric for applications prioritizing precision.} \textcolor{black}{This limitation matters most in clinical use cases where false positives drive unnecessary interventions or follow-up procedures (e.g., biopsy referral, repeat scans). Based on our experiments (Fig.~\ref{fig:cls_results}, Fig.~\ref{fig:ensemble}), MOVAL's precision estimate is reasonably reliable when validation and target prevalence are similar and the positive class is not extremely rare; otherwise, we recommend treating it as indicative only and confirming with a modest labeled sample or the more robust F1-Score or sensitivity estimates.}

\textcolor{black}{MOVAL relies on a calibration parameter learned from the validation dataset and applied, without adjustment, to unlabeled test data—an assumption, not a guarantee, since the ideal parameters would require test-domain labels. We therefore describe MOVAL as a practical monitoring and performance-estimation tool that can support deployment decisions, rather than one that ensures safe deployment under arbitrary shift. Its validity depends on the shift type. Under acquisition shifts, where the scanner, protocol, or site changes but the model's input-output relationship is largely preserved, we find empirically that validation-domain calibration remains informative (Section~\ref{sec:experiments}). Under prevalence or class-ratio shifts, MOVAL's class-wise design is comparatively robust. Under more severe shifts, where the confidence-correctness relationship changes qualitatively, calibration may not transfer, and we recommend a small labeled test sample to check and, if necessary, recalibrate.}

\textcolor{black}{Estimating performance from unlabeled test data is not identifiable in general without assumptions linking validation-domain calibration to test-domain correctness, and no distribution-free guarantee should be expected under arbitrary shift. In the acquisition-shift regime we target, a bounded divergence between validation and target confidence-score distributions, a stable class-conditional label distribution, or an identifiable label-shift structure would let one bound the estimation error; access to multiple labeled source domains would offer a further, more permissive route to a worst-case bound. We leave a formal treatment of these assumptions to future work.} Several other directions for future work remain to enhance MOVAL's applicability and robustness. First, integrating test-time calibration, where the model is adjusted on-the-fly using target domain data~\citep{wang2020learning, hu2024pseudo} or information from multiple domains~\citep{yu2022robust}, could improve adaptability, though such methods typically require retraining or access to multiple source domains that may not be available at deployment; we view them as complementary rather than as direct substitutes for MOVAL's post-training, single-source setting. \textcolor{black}{Second, more extensive evaluation across a wider range of applications (e.g., ultrasound), tasks (e.g., 3D classification), and model architectures is warranted. In particular, broader validation on modern foundation models, multimodal models, and additional modalities such as CT, ultrasound, and video remains future work.}

\textcolor{black}{For practitioners, we summarize the practical guidance that follows from our ablations (Fig.~\ref{fig:ablation}). Calibration set size: as few as 5$\%$ of the validation data (approximately 60 samples per class for classification) is sufficient for stable estimates; for segmentation, about 12 slices (2D) or 6 volumes (3D) are typically needed. Class-wise versus group-wise calibration: class-wise calibration is preferred when each class has enough validation samples (rule of thumb, tens of samples per class); when some classes are rare, group-wise calibration, which shares parameters across a small number of groups, achieves comparable accuracy with substantially fewer validation samples per group. Interpreting disagreement across confidence scores: when individual confidence scores (MCP, Entropy, Doctor, Energy) diverge substantially in their estimates, this itself is a useful diagnostic that the shift may be more severe than an acquisition shift, and we recommend defaulting to the MOVAL Ensemble estimate, which is more robust than any single score, in these cases. Triggering manual review: because MOVAL cannot certify that validation-domain calibration transfers, we recommend manual review or additional labeling whenever the estimated performance drops below the acceptance threshold for the application, when confidence-score disagreement is high, or when the estimated prevalence distribution departs substantially from the validation distribution.}

MOVAL is parameter-efficient, at most requiring only $c$ (the number of classes) × $L$ (the number of ensemble choices) parameters. This leads to computationally efficient and reproducible optimization and inference processes, especially compared to methods that require references during inference~\citep{valindria2017reverse} or an additional neural network~\citep{robinson2018real}. We provide a Python API for all the algorithms described above, designed for use and enhancement with every trained model. \textcolor{black}{Given the importance of time efficiency in machine learning applications like autonomous driving and surgical image analysis, we believe our approach can serve as a practical, real-time monitoring aid that supports, but does not by itself guarantee, safe deployment in safety-critical applications.}

\section*{Acknowledgments}
\label{sec:ack}

This project received funding from the ERC under the EU's Horizon 2020 research and innovation programme (grant No 757173). MR is funded by an Imperial College President’s PhD scholarship and a Google PhD Fellowship. 

\section*{Author contributions}
  \textbf{Conceptualization:} ZL, MI, BG.
  \textbf{Methodology:} ZL, MR, CO.
  \textbf{Software:} ZL.
  \textbf{Validation:} ZL.
  \textbf{Formal analysis:} ZL.
  \textbf{Data curation:} ZL, CO.
  \textbf{Writing -- original draft:} ZL.
  \textbf{Writing -- review \& editing:} ZL, KK, CC, MR, BG.
  \textbf{Supervision:} BG.
  \textbf{Project administration:} BG.
  \textbf{Funding acquisition:} BG.

\appendix

\section{Quantitative Results}
\label{app:results}

\begin{table*}[ht]
\centering
 \resizebox{\linewidth}{!}{%
    \begin{tabular}{ccccccccccccc}
    \toprule
        \multicolumn{3}{c}{\multirow{3}{*}{\shortstack{Dataset \\ and Training \\ Strategy}}} & &  \multicolumn{9}{c}{Confidence Score}  \\
        & & & \multirow{2}{*}{\shortstack{Estimation \\ Algorithm}} & \multicolumn{4}{c}{Class-Agnostic} & \multicolumn{4}{c}{Class-Specific} & \multirow{2}{*}{\shortstack{MOVAL \\ Ensemble}} \\
        & & & & MCP & Energy & Entropy & Doctor & MCP & Energy & Entropy & Doctor \\
        \cmidrule(r){1-3} \cmidrule(r){4-4} \cmidrule(lr){5-8} \cmidrule(lr){9-12} \cmidrule(lr){13-13}
        
        \multirow{5}{*}{\rotatebox{90}{\shortstack{CIFAR10}}} & \multirow{5}{*}{\rotatebox{90}{\shortstack{Synthetic-95}}} & \multirow{5}{*}{\rotatebox{90}{\shortstack{Vanilla \\ Training}}} & Vanilla  & \ms{32.6}{6.2} & \ms{29.4}{5.2} & \ms{30.5}{6.1} & \ms{36.3}{6.5} & --- & --- & --- & --- & \multirow{5}{*}{\msb{2.8}{2.1}} \\
        & & & TS  & \ms{6.5}{5.0} & \ms{13.5}{6.7} & \ms{5.5}{5.0} & \ms{4.9}{4.0} & \ms{7.1}{5.4} & \ms{12.4}{7.6} & \ms{5.9}{4.9} & \ms{4.0}{3.7} \\
        & & & DoC  & \ms{11.7}{6.2} & \ms{8.5}{5.3} & \ms{10.9}{6.1} & \ms{12.6}{6.5} & \ms{10.9}{6.5} & \ms{8.5}{5.9} & \ms{10.3}{6.1} & \ms{11.6}{6.7}  \\
        & & & ATC  & \ms{4.1}{3.7} & \ms{6.2}{3.7} & \ms{4.2}{3.7} & \ms{4.1}{3.7} & \ms{3.5}{3.5} & \ms{3.2}{1.8} & \ms{3.3}{3.2} & \ms{3.4}{3.4} \\
        & & & TS-ATC  & \ms{5.0}{2.6} & \ms{6.7}{3.7} & \ms{5.4}{3.1} & \ms{5.3}{2.9} & \msb{2.8}{2.7} & \ms{3.6}{2.2} & \msb{2.8}{2.4} & \msbb{3.0}{2.2} \\

        \midrule

        \multirow{5}{*}{\rotatebox{90}{\shortstack{CIFAR100}}} & \multirow{5}{*}{\rotatebox{90}{\shortstack{Synthetic-95}}} & \multirow{5}{*}{\rotatebox{90}{\shortstack{Vanilla \\ Training}}} & Vanilla  & \ms{46.7}{4.3} & \ms{5.4}{3.1} & \ms{55.6}{4.6} & \ms{64.8}{4.9} & --- & --- & --- & --- & \multirow{5}{*}{\msbb{2.5}{2.3}} \\
        & & & TS  & \ms{6.1}{3.8} & \ms{10.4}{4.8} & \ms{6.9}{4.2} & \ms{3.3}{2.9} & \ms{5.8}{3.2} & \ms{10.5}{3.9} & \ms{6.5}{3.6} & \ms{2.8}{2.8} \\
        & & & DoC  & \ms{9.0}{4.3} & \ms{8.7}{4.5} & \ms{9.7}{4.6} & \ms{11.0}{4.9} & \ms{8.5}{3.6} & \ms{7.8}{3.8} & \ms{9.0}{3.8} & \ms{10.3}{4.1} \\
        & & & ATC  & \ms{4.2}{3.1} & \ms{3.3}{2.4} & \ms{3.8}{3.0} & \ms{4.2}{3.1} & \ms{3.9}{2.7} & \ms{2.9}{2.6} & \ms{3.3}{2.7} & \ms{3.8}{2.7} \\
        & & & TS-ATC  & \ms{2.7}{4.2} & \ms{4.2}{3.3} & \ms{3.0}{2.2} & \ms{2.7}{1.7} & \msb{2.2}{2.1} & \ms{3.5}{3.2} & \ms{2.6}{2.4} & \msb{2.2}{1.8} \\

        \midrule

        \multirow{5}{*}{\rotatebox{90}{\shortstack{HAM10000}}} & \multirow{5}{*}{\rotatebox{90}{\shortstack{Natural-7}}} & \multirow{5}{*}{\rotatebox{90}{\shortstack{Vanilla \\ Training}}} & Vanilla  & \ms{20.8}{11.6} & \ms{41.8}{12.2} & \ms{15.9}{9.4} & \ms{20.1}{10.8} & --- & --- & --- & --- & \multirow{5}{*}{\msb{3.9}{3.8}} \\
        & & & TS  & \ms{11.0}{6.6} & \ms{20.1}{10.2} & \ms{9.8}{6.3} & \ms{8.9}{5.4} & \ms{11.1}{6.7} & \ms{19.3}{10.4} & \ms{9.8}{6.3} & \ms{8.2}{5.5} \\
        & & & DoC  & \ms{15.3}{8.3} & \ms{14.0}{7.8} & \ms{12.6}{7.3} & \ms{14.7}{8.1} & \ms{15.1}{8.5} & \ms{13.8}{7.8} & \ms{12.6}{7.4} & \ms{14.5}{8.2} \\
        & & & ATC  & \ms{8.3}{5.0} & \ms{4.9}{3.0} & \ms{7.4}{4.2} & \ms{8.6}{5.0} & \ms{6.0}{4.3} & \ms{5.1}{3.5} & \ms{5.4}{4.0} & \ms{6.2}{4.3} \\
        & & & TS-ATC  & \ms{6.9}{6.3} & \ms{7.7}{6.3} & \ms{5.5}{2.8} & \ms{6.5}{3.2} & \ms{4.7}{4.0} & \ms{7.4}{5.8} & \msbb{4.3}{3.8} & \msbb{4.3}{3.9} \\
        
        \bottomrule
    \end{tabular}%
}
\caption{Related to Fig.~\ref{fig:cls_results}. Evaluation of accuracy estimation for different classification tasks under various types of domain shifts based on MAE. The best and second-best results within the same category are in \textbf{bold} and \underline{underlined} for each configuration.}
\label{tab:cls_acc}
\end{table*}

\begin{table*}[ht]
\centering
 \resizebox{\linewidth}{!}{%
    \begin{tabular}{cccccccccccc}
    \toprule
        \multicolumn{2}{c}{\multirow{3}{*}{\shortstack{Dataset}}} & &  \multicolumn{9}{c}{Confidence Score}  \\
        & & \multirow{2}{*}{\shortstack{Estimation \\ Algorithm}} & \multicolumn{4}{c}{Class-Agnostic} & \multicolumn{4}{c}{Class-Specific} & \multirow{2}{*}{\shortstack{MOVAL \\ Ensemble}} \\
        & & &  MCP & Energy & Entropy & Doctor & MCP & Energy & Entropy & Doctor \\
        \cmidrule(r){1-2} \cmidrule(r){3-3} \cmidrule(lr){4-7} \cmidrule(lr){8-11} \cmidrule(lr){12-12}
        
        \multirow{10}{*}{\rotatebox{90}{\shortstack{Cardiac-2D}}} & \multirow{5}{*}{\rotatebox{90}{\shortstack{Synthetic-65}}} & Vanilla  & \ms{17.8}{7.7} & \ms{52.5}{9.0} & \ms{16.3}{7.2} & \ms{17.5}{7.5} & --- & --- & --- & ---& \multirow{5}{*}{\ms{4.3}{5.0}} \\
        & & TS  & \ms{14.8}{7.1} & \ms{19.8}{8.4} & \ms{13.3}{6.7} & \ms{13.7}{6.7} & \ms{5.1}{5.0} & \ms{6.3}{7.0} & \ms{4.7}{4.2} & \ms{4.5}{3.9} \\
        & & DoC  & \ms{17.4}{7.7} & \ms{17.0}{8.0} & \ms{15.9}{7.2} & \ms{17.1}{7.5} & \ms{5.6}{6.0} & \ms{5.9}{6.1} & \ms{5.2}{5.5} & \ms{5.5}{5.9} \\
        & & ATC  & \ms{6.5}{4.5} & \ms{19.8}{8.4} & \ms{6.4}{4.4} & \ms{6.5}{4.5} & \msb{3.6}{4.2} & \ms{4.5}{7.7} & \msb{3.6}{4.3} & \msb{3.6}{4.2} \\
        & & TS-ATC  & \ms{6.2}{6.7} & \ms{19.8}{8.4} & \ms{6.2}{4.4} & \ms{6.2}{4.4} & \msbb{4.0}{6.2} & \ms{6.2}{10.1} & \msbb{4.0}{6.2} & \msbb{4.0}{6.2} \\

        \cmidrule(r){2-12}

        & \multirow{5}{*}{\rotatebox{90}{\shortstack{Natural-4}}} & Vanilla  &  \ms{15.9}{0.8} & \ms{52.3}{1.8} & \ms{14.8}{0.8} & \ms{15.7}{0.8} & --- & --- & --- & --- & \multirow{5}{*}{\msbb{2.6}{0.4}} \\
        & & TS  & \ms{13.8}{0.8} & \ms{17.3}{0.9} & \ms{12.8}{0.8} & \ms{13.1}{0.8} & \ms{3.4}{0.6} & \msbb{2.6}{0.5} & \ms{3.8}{0.4} & \ms{3.8}{0.4} \\
        & & DoC  & \ms{15.5}{0.8} & \ms{15.4}{1.0} & \ms{14.5}{0.8} & \ms{15.3}{0.8} & \msb{2.4}{0.4} & \ms{2.9}{0.7} & \msb{2.4}{0.4} & \msb{2.4}{0.4} \\
        & & ATC  & \ms{7.8}{0.4} & \ms{17.3}{0.9} & \ms{7.8}{0.4} & \ms{7.8}{0.4} & \ms{3.6}{0.2} & \ms{5.5}{1.2} & \ms{3.6}{0.2} & \ms{3.6}{0.2} \\
        & & TS-ATC  & \ms{7.9}{0.8} & \ms{17.3}{0.9} & \ms{7.9}{0.4} & \ms{7.9}{0.4} & \ms{4.5}{0.8} & \ms{8.1}{2.0} & \ms{4.5}{0.8} & \ms{4.5}{0.8} \\

        \midrule

        \multirow{10}{*}{\rotatebox{90}{\shortstack{Prostate-3D}}} & \multirow{5}{*}{\rotatebox{90}{\shortstack{Synthetic-83}}} & Vanilla  & \ms{12.7}{8.0} & \ms{61.4}{5.3} & \ms{8.6}{4.8} & \ms{10.7}{6.4} & --- & --- & --- & --- & \multirow{5}{*}{\ms{1.7}{2.2}} \\
        & & TS  & \ms{10.4}{7.0} & \ms{13.1}{9.0} & \ms{6.1}{3.4} & \ms{7.2}{4.2} & \ms{2.3}{6.5} & \ms{4.0}{9.0} & \msbb{1.4}{1.6} & \msb{1.3}{1.7} \\
        & & DoC  & \ms{12.5}{8.0} & \ms{9.1}{7.8} & \ms{8.5}{4.8} & \ms{10.6}{6.4} & \ms{3.4}{7.8} & \ms{2.8}{7.6} & \ms{1.7}{4.2} & \ms{2.4}{6.1} \\
        & & ATC  & \ms{3.5}{1.6} & \ms{14.1}{9.1} & \ms{3.5}{1.6} & \ms{3.5}{1.6} & \ms{3.1}{5.3} & \ms{3.2}{5.4} & \ms{3.1}{5.3} & \ms{3.1}{5.3} \\
        & & TS-ATC  & \ms{3.5}{3.4} & \ms{3.1}{3.4} & \ms{3.5}{1.6} & \ms{3.5}{1.6} & \ms{3.1}{5.3} & \ms{3.6}{5.9} & \ms{3.1}{5.3} & \ms{3.1}{5.3} \\

        \cmidrule(r){2-12}

        & \multirow{5}{*}{\rotatebox{90}{\shortstack{Natural-5}}} & Vanilla  & \ms{18.9}{6.6} & \ms{57.2}{5.3} & \ms{12.9}{4.3} & \ms{16.0}{5.4} & --- & --- & --- & --- & \multirow{5}{*}{\msb{1.6}{1.4}} \\
        & & TS  & \ms{16.0}{5.9} & \ms{20.0}{7.4} & \ms{9.7}{3.5} & \ms{11.2}{4.0} & \ms{7.4}{5.8} & \ms{11.0}{7.4} & \ms{2.2}{1.8} & \ms{2.5}{2.1} \\
        & & DoC  & \ms{18.8}{6.6} & \ms{15.2}{6.6} & \ms{12.8}{4.3} & \ms{15.9}{5.3} & \ms{9.5}{6.4} & \ms{8.7}{6.5} & \ms{5.4}{4.0} & \ms{7.4}{5.1} \\
        & & ATC  & \ms{5.2}{1.9} & \ms{21.1}{7.4} & \ms{5.2}{1.9} & \ms{5.2}{1.9} & \ms{2.7}{2.1} & \ms{2.7}{2.1} & \ms{2.7}{2.1} & \ms{2.7}{2.1} \\
        & & TS-ATC  & \ms{5.2}{3.5} & \msbb{2.1}{0.9} & \ms{5.2}{1.9} & \ms{5.2}{1.9} & \ms{2.7}{2.1} & \ms{2.7}{2.0} & \ms{2.7}{2.1} & \ms{2.7}{2.1} \\

        \midrule

        \multirow{10}{*}{\rotatebox{90}{\shortstack{Brain Lesion-3D}}} & \multirow{5}{*}{\rotatebox{90}{\shortstack{Synthetic-83}}} & Vanilla  & \ms{23.2}{7.7} & \ms{73.0}{8.2} & \ms{18.7}{5.9} & \ms{21.1}{6.7} & --- & --- & --- & --- & \multirow{5}{*}{\msbb{2.1}{3.3}} \\
        & & TS  & \ms{20.9}{7.2} & \ms{24.8}{8.4} & \ms{17.3}{5.6} & \ms{18.0}{5.8} & \ms{4.4}{7.2} & \ms{5.1}{8.3} & \ms{2.2}{4.1} & \ms{2.3}{4.2} \\
        & & DoC  & \ms{23.1}{7.7} & \ms{24.8}{8.4} & \ms{18.7}{5.9} & \ms{21.0}{6.7} & \ms{4.3}{7.3} & \ms{5.1}{8.3} & \ms{2.8}{5.4} & \ms{3.4}{6.2} \\
        & & ATC  &  \ms{15.3}{4.6} & \ms{24.8}{8.4} & \ms{15.3}{4.6} & \ms{15.3}{4.6} & \ms{2.3}{4.0} & \ms{2.3}{4.0} & \ms{2.3}{4.0} & \ms{2.3}{4.0} \\
        & & TS-ATC  & \ms{15.3}{5.6} & \ms{24.8}{8.4} & \ms{15.3}{4.6} & \ms{15.3}{4.6} & \ms{2.3}{4.0} & \msb{1.8}{2.6} & \ms{2.3}{4.0} & \ms{2.3}{4.0} \\

        \cmidrule(r){2-12}

        & \multirow{5}{*}{\rotatebox{90}{\shortstack{Natural-12}}}  & Vanilla  & \ms{51.8}{15.0} & \ms{39.2}{16.7} & \ms{45.5}{12.6} & \ms{48.5}{13.4} & --- & --- & --- & --- & \multirow{5}{*}{\msbb{25.5}{12.3}} \\
        & & TS  & \ms{49.3}{14.8} & \ms{54.0}{15.9} & \ms{44.0}{12.4} & \ms{44.9}{12.6} & \ms{34.4}{16.2} & \ms{35.7}{17.1} & \ms{28.7}{12.7} & \ms{28.9}{12.8} \\
        & & DoC  &  \ms{51.7}{15.0} & \ms{54.0}{15.9} & \ms{45.5}{12.6} & \ms{48.5}{13.4} & \ms{33.9}{15.8} & \ms{35.6}{17.1} & \ms{30.3}{13.3} & \ms{31.8}{14.1} \\
        & & ATC  & \ms{40.5}{11.0} & \ms{54.0}{15.9} & \ms{40.5}{11.0} & \ms{40.5}{11.0} & \ms{28.8}{12.4} & \ms{28.7}{12.3} & \ms{28.8}{12.4} & \ms{28.8}{12.4} \\
        & & TS-ATC  & \ms{40.5}{12.4} & \ms{54.0}{15.9} & \ms{40.5}{11.0} & \ms{40.5}{11.0} & \ms{28.8}{12.4} & \msb{16.8}{10.7} & \ms{28.8}{12.4} & \ms{28.8}{12.4} \\
        
        \bottomrule
    \end{tabular}%
}
\caption{Related to Fig.~\ref{fig:seg_results}. Evaluation of DSC (F1-Score) estimation for different segmentation tasks under various types of domain shifts based on MAE. The best and second-best results within the same category are in \textbf{bold} and \underline{underlined} for each configuration.}
\label{tab:seg}
\end{table*}

\clearpage
\newpage

\section{MOVAL API and Example Usage}
\label{app:api}

We offer a Python implementation of MOVAL that supports input in the form of NumPy arrays~\citep{harris2020array}~\footnote{Classification datasets and models:~\url{https://github.com/ZerojumpLine/Robust-Skin-Lesion-Classification}. Segmentation datasets and models:~\url{https://github.com/ZerojumpLine/Robust-Medical-Segmentation}. 
Project page:~\url{https://zerojumpline.github.io/moval/}. Code:~\url{https://github.com/ZerojumpLine/MOVAL}. Documentation:~\url{https://moval.readthedocs.io/en/latest/index.html}. Code (and data) to reproduce the figures:~\url{https://github.com/ZerojumpLine/moval_scripts}.}. \textcolor{black}{The \texttt{moval} package (installable via \texttt{pip install moval}) implements only the MOVAL estimation algorithms (confidence scores, calibration, metric computation) for use with any trained model; it does not itself contain the paper's experiments. The separate \texttt{moval\_scripts} repository, together with the dataset and model repositories linked above, contains the scripts and configurations needed to reproduce the main classification and segmentation experiments reported in this paper.} The provided API is compatible with scikit-learn~\citep{pedregosa2011scikit}, a widely used machine learning library. This compatibility allows users to leverage scikit-learn tools for downstream processing of the computed probabilities, such as classification, clustering, and dimensionality reduction~\citep{schneider2023learnable}.

In a classification task with \(c=10\) classes, the validation set contains 100 samples, while the test dataset consists of 200 samples:

\begin{mdframed}
\begingroup
\fvset{fontsize=\scriptsize}
\begin{Verbatim}[commandchars=\\\{\}]
\PYG{k+kn}{import} \PYG{n+nn}{numpy} \PYG{k}{as} \PYG{n+nn}{np}
\PYG{n}{C} \PYG{o}{=} \PYG{l+m+mi}{10}
\PYG{n}{logits} \PYG{o}{=} \PYG{n}{np}\PYG{o}{.}\PYG{n}{random}\PYG{o}{.}\PYG{n}{randn}\PYG{p}{(}\PYG{l+m+mi}{100}\PYG{p}{,} \PYG{n}{C}\PYG{p}{)}
\PYG{n}{gts} \PYG{o}{=} \PYG{n}{np}\PYG{o}{.}\PYG{n}{random}\PYG{o}{.}\PYG{n}{randint}\PYG{p}{(}\PYG{l+m+mi}{0}\PYG{p}{,} \PYG{n}{C}\PYG{p}{,} \PYG{p}{(}\PYG{l+m+mi}{100}\PYG{p}{))}
\PYG{n}{l\PYGZus{}test} \PYG{o}{=} \PYG{n}{np}\PYG{o}{.}\PYG{n}{random}\PYG{o}{.}\PYG{n}{randn}\PYG{p}{(}\PYG{l+m+mi}{200}\PYG{p}{,} \PYG{n}{C}\PYG{p}{)}
\end{Verbatim}

\endgroup
\end{mdframed}

For example, a MOVAL classification model calibrated for accuracy using TS with Entropy can be defined as:

\begin{mdframed}
\begingroup
\fvset{fontsize=\scriptsize}
\begin{Verbatim}[commandchars=\\\{\}]
\PYG{k+kn}{import} \PYG{n+nn}{moval}
\PYG{n}{moval\PYGZus{}model} \PYG{o}{=} \PYG{n}{moval}\PYG{o}{.}\PYG{n}{MOVAL}\PYG{p}{(}
    \PYG{n}{mode} \PYG{o}{=} \PYG{l+s+s2}{\PYGZdq{}classification\PYGZdq{}}\PYG{p}{,}
    \PYG{n}{confidence\PYGZus{}scores} \PYG{o}{=} \PYG{l+s+s2}{\PYGZdq{}entropy\PYGZhy{}conf\PYGZdq{}}\PYG{p}{,}
    \PYG{n}{estim\PYGZus{}algorithm} \PYG{o}{=} \PYG{l+s+s2}{\PYGZdq{}ts\PYGZhy{}model\PYGZdq{}}\PYG{p}{,}
    \PYG{n}{class\PYGZus{}specific} \PYG{o}{=} \PYG{k+kc}{False}
\PYG{p}{)}
\end{Verbatim}

\endgroup
\end{mdframed}

Alternatively, as we suggest, a MOVAL Ensemble model can be constructed with:

\begin{mdframed}
\begingroup
\fvset{fontsize=\scriptsize}
\begin{Verbatim}[commandchars=\\\{\}]
\PYG{n}{moval\PYGZus{}model} \PYG{o}{=} \PYG{n}{moval}\PYG{o}{.}\PYG{n}{MOVAL}\PYG{p}{(}
    \PYG{n}{mode} \PYG{o}{=} \PYG{l+s+s2}{\PYGZdq{}classification\PYGZdq{}}\PYG{p}{,}
    \PYG{n}{estim\PYGZus{}algorithm} \PYG{o}{=} \PYGZbs{}
    \PYG{l+s+s2}{\PYGZdq{}moval\PYGZhy{}ensemble\PYGZhy{}cls\PYGZhy{}accuracy\PYGZdq{}}
\PYG{p}{)}
\end{Verbatim}

\endgroup
\end{mdframed}

For a 3D segmentation task with \(c=2\) classes, the validation set consists of 5 samples, and the test set includes 1 sample, defined as:

\begin{mdframed}
\begingroup
\fvset{fontsize=\scriptsize}
\begin{Verbatim}[commandchars=\\\{\}]
\PYG{n}{C} \PYG{o}{=} \PYG{l+m+mi}{2} \PYG{p}{;} \PYG{n}{H} \PYG{o}{=} \PYG{l+m+mi}{50} \PYG{p}{;} \PYG{n}{W} \PYG{o}{=} \PYG{l+m+mi}{50} \PYG{p}{;} \PYG{n}{D} \PYG{o}{=} \PYG{l+m+mi}{80}
\PYG{n}{logits} \PYG{o}{=} \PYG{p}{[]} \PYG{p}{;} \PYG{n}{gts} \PYG{o}{=} \PYG{p}{[]}
\PYG{k}{for} \PYG{n}{\PYGZus{}} \PYG{o+ow}{in} \PYG{n+nb}{range}\PYG{p}{(}\PYG{l+m+mi}{5}\PYG{p}{):}
    \PYG{n}{logit} \PYG{o}{=} \PYG{n}{np}\PYG{o}{.}\PYG{n}{random}\PYG{o}{.}\PYG{n}{randn}\PYG{p}{(}\PYG{n}{C}\PYG{p}{,} \PYG{n}{H}\PYG{p}{,} \PYG{n}{W}\PYG{p}{,} \PYG{n}{D}\PYG{p}{)}
    \PYG{n}{gt} \PYG{o}{=} \PYG{n}{np}\PYG{o}{.}\PYG{n}{random}\PYG{o}{.}\PYG{n}{randint}\PYG{p}{(}\PYG{l+m+mi}{0}\PYG{p}{,} \PYG{n}{C}\PYG{p}{,} \PYG{p}{(}\PYG{n}{H}\PYG{p}{,} \PYG{n}{W}\PYG{p}{,} \PYG{n}{D}\PYG{p}{))}
    \PYG{n}{logits}\PYG{o}{.}\PYG{n}{append}\PYG{p}{(}\PYG{n}{logit}\PYG{p}{)}
    \PYG{n}{gts}\PYG{o}{.}\PYG{n}{append}\PYG{p}{(}\PYG{n}{gt}\PYG{p}{)}
\PYG{n}{l\PYGZus{}test} \PYG{o}{=} \PYG{p}{[}\PYG{n}{np}\PYG{o}{.}\PYG{n}{random}\PYG{o}{.}\PYG{n}{randn}\PYG{p}{(}\PYG{n}{C}\PYG{p}{,} \PYG{n}{H}\PYG{p}{,} \PYG{n}{W}\PYG{p}{,} \PYG{n}{D}\PYG{p}{)]}
\end{Verbatim}

\endgroup
\end{mdframed}

Similarly, a MOVAL segmentation model calibrated for F1-Score using CS ATC with Doctor can be defined as:

\begin{mdframed}
\begingroup
\fvset{fontsize=\scriptsize}
\begin{Verbatim}[commandchars=\\\{\}]
\PYG{n}{moval\PYGZus{}model} \PYG{o}{=} \PYG{n}{moval}\PYG{o}{.}\PYG{n}{MOVAL}\PYG{p}{(}
    \PYG{n}{mode} \PYG{o}{=} \PYG{l+s+s2}{\PYGZdq{}segmentation\PYGZdq{}}\PYG{p}{,}
    \PYG{n}{metric} \PYG{o}{=} \PYG{l+s+s2}{\PYGZdq{}f1score\PYGZdq{}}\PYG{p}{,}
    \PYG{n}{confidence\PYGZus{}scores} \PYG{o}{=} \PYG{l+s+s2}{\PYGZdq{}doctor\PYGZhy{}conf\PYGZdq{}}\PYG{p}{,}
    \PYG{n}{estim\PYGZus{}algorithm} \PYG{o}{=} \PYG{l+s+s2}{\PYGZdq{}atc\PYGZhy{}model\PYGZdq{}}\PYG{p}{,}
    \PYG{n}{class\PYGZus{}specific} \PYG{o}{=} \PYG{k+kc}{True}
\PYG{p}{)}
\end{Verbatim}

\endgroup
\end{mdframed}

The fitting and inference processes for the MOVAL model in both tasks are identical:

\begin{mdframed}
\begingroup
\fvset{fontsize=\scriptsize}
\begin{Verbatim}[commandchars=\\\{\}]
\PYG{n}{moval\PYGZus{}model}\PYG{o}{.}\PYG{n}{fit}\PYG{p}{(}\PYG{n}{logits}\PYG{p}{,} \PYG{n}{gts}\PYG{p}{)}
\PYG{n}{estim} \PYG{o}{=} \PYG{n}{moval\PYGZus{}model}\PYG{o}{.}\PYG{n}{estimate}\PYG{p}{(}\PYG{n}{l\PYGZus{}test}\PYG{p}{)}
\PYG{c+c1}{\PYGZsh{} estim is of shape (1, ) for accuracy,}
\PYG{c+c1}{\PYGZsh{} while (c, ) for group\PYGZhy{}wise metrics}
\end{Verbatim}

\endgroup
\end{mdframed}

After optimization, we can obtain instance-wise estimation for the validation data using:

\begin{mdframed}
\begingroup
\fvset{fontsize=\scriptsize}
\begin{Verbatim}[commandchars=\\\{\}]
\PYG{n}{estim\PYGZus{}val} \PYG{o}{=} \PYG{n}{moval\PYGZus{}model}\PYG{o}{.}\PYG{n}{fitted\PYGZus{}perf}
\PYG{c+c1}{\PYGZsh{} estim\PYGZus{}val has N values for classification}
\PYG{c+c1}{\PYGZsh{} while Z values for segmentation}
\end{Verbatim}

\endgroup
\end{mdframed}

In addition to the overall estimated metrics, we can also derive the calibrated probabilities \(\hat{p}_i\):

\begin{mdframed}
\begingroup
\fvset{fontsize=\scriptsize}
\begin{Verbatim}[commandchars=\\\{\}]
\PYG{n}{prob} \PYG{o}{=} \PYG{n}{moval\PYGZus{}model}\PYG{o}{.}\PYG{n}{get\PYGZus{}probability}\PYG{p}{(}\PYG{n}{l\PYGZus{}tests}\PYG{p}{)}
\end{Verbatim}

\endgroup
\end{mdframed}

We recommend supplying the MOVAL model with all pre-trained models. To achieve this, the parameters can be saved and loaded using:

\begin{mdframed}
\begingroup
\fvset{fontsize=\scriptsize}
\begin{Verbatim}[commandchars=\\\{\}]
\PYG{n}{moval\PYGZus{}model}\PYG{o}{.}\PYG{n}{save}\PYG{p}{(}\PYG{l+s+s1}{\PYGZsq{}./m.pkl\PYGZsq{}}\PYG{p}{)}
\PYG{n}{loaded\PYGZus{}model} \PYG{o}{=} \PYG{n}{moval}\PYG{o}{.}\PYG{n}{MOVAL}\PYG{o}{.}\PYG{n}{load}\PYG{p}{(}\PYG{l+s+s1}{\PYGZsq{}./m.pkl\PYGZsq{}}\PYG{p}{)}
\end{Verbatim}

\endgroup
\end{mdframed}

\section{Supplementary Figures}
\label{sec_supp_figures}

\begin{figure*}[ht]
  \centering
   \includegraphics[width=\linewidth]{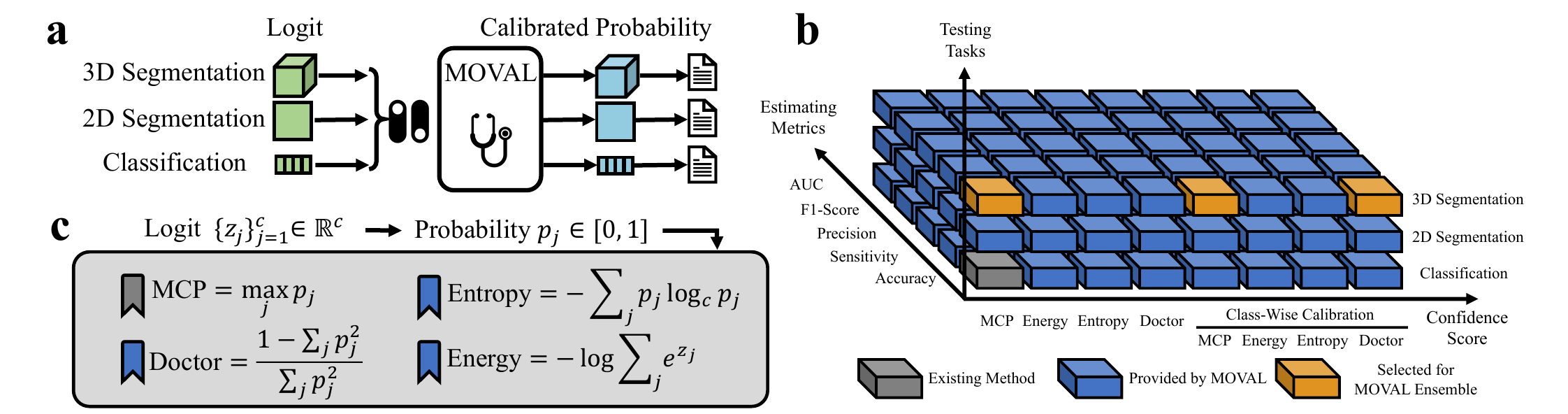}
   \caption{Comparison of MOVAL with existing standards.
   \textbf{(a)}: Within a unified framework, MOVAL can be readily applied to classification, 2D, and 3D segmentation tasks. 
   \textbf{(b)}: After extending existing methods in many dimensions, we find that an ensemble of multiple confidence scores leads to more robust and state-of-the-art results.
   \textbf{(c)}: MOVAL explores performance estimation based on confidence scores rather than maximum class probability (MCP), which is the basis of all existing methods.
   }
   \label{fig:method_partone}
\end{figure*}

\begin{figure*}[ht]
  \centering
   \includegraphics[width=\linewidth]{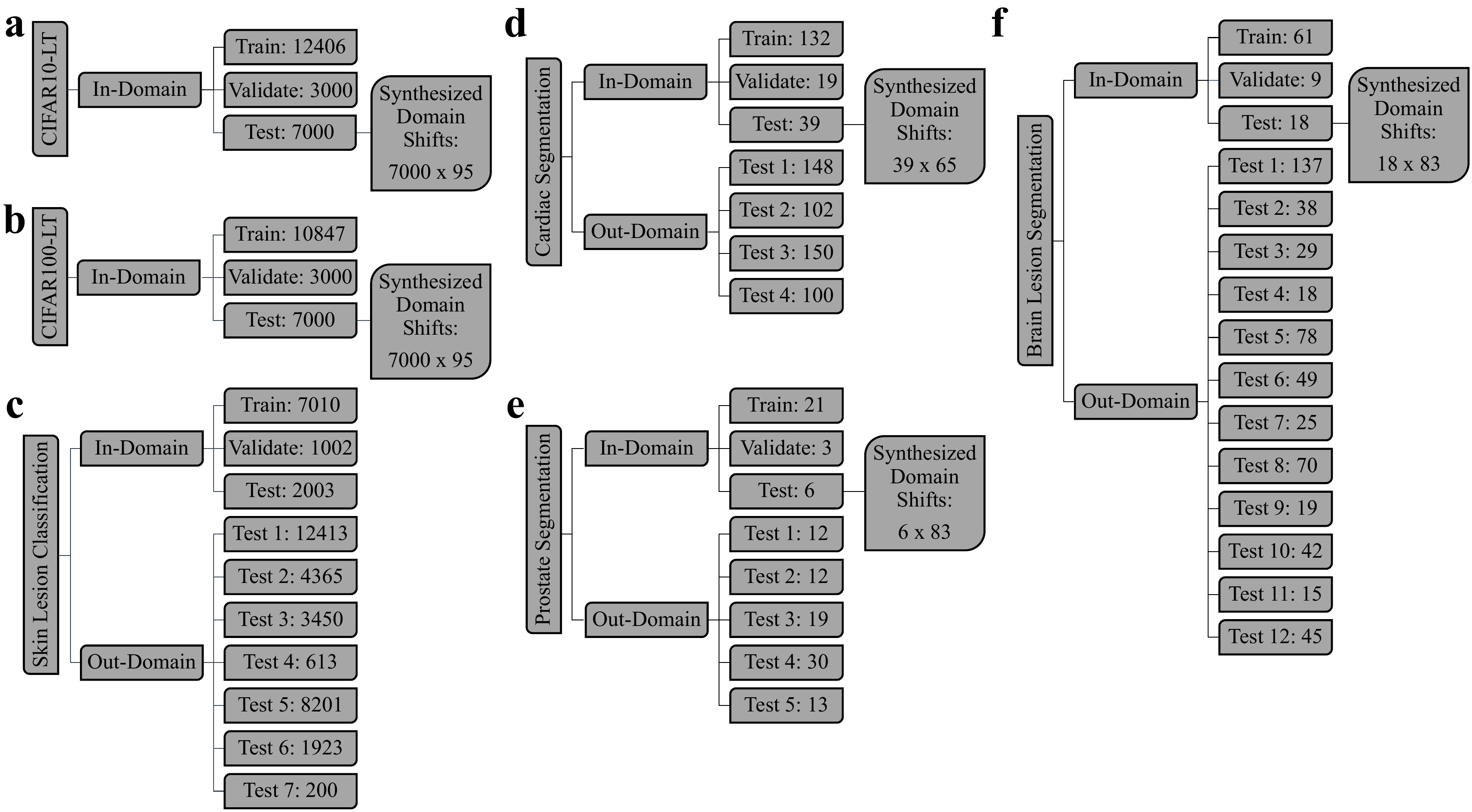}
   \caption{Summary of the datasets used in this study.
   \textbf{(a)}: Classification task \uppercase\expandafter{\romannumeral1}, Long-tailed version of the CIFAR10 dataset. 
   \textbf{(b)}: Classification task \uppercase\expandafter{\romannumeral2}, Long-tailed version of the CIFAR10 dataset. 
   \textbf{(c)}: Classification task \uppercase\expandafter{\romannumeral3}, skin lesion classification from 8 sources of data.
   \textbf{(d)}: 2D segmentation task \uppercase\expandafter{\romannumeral1}, cardiac MRI segmentation from 5 different scanners. 
   \textbf{(e)}: 3D segmentation task \uppercase\expandafter{\romannumeral1}, prostate MRI segmentation from 6 different scanners.
   \textbf{(f)}: 3D segmentation task \uppercase\expandafter{\romannumeral2}, brain lesion segmentation from 13 different scanners.
   }
   \label{fig:data}
\end{figure*}

\begin{figure*}[ht]
  \centering
   \includegraphics[width=\linewidth]{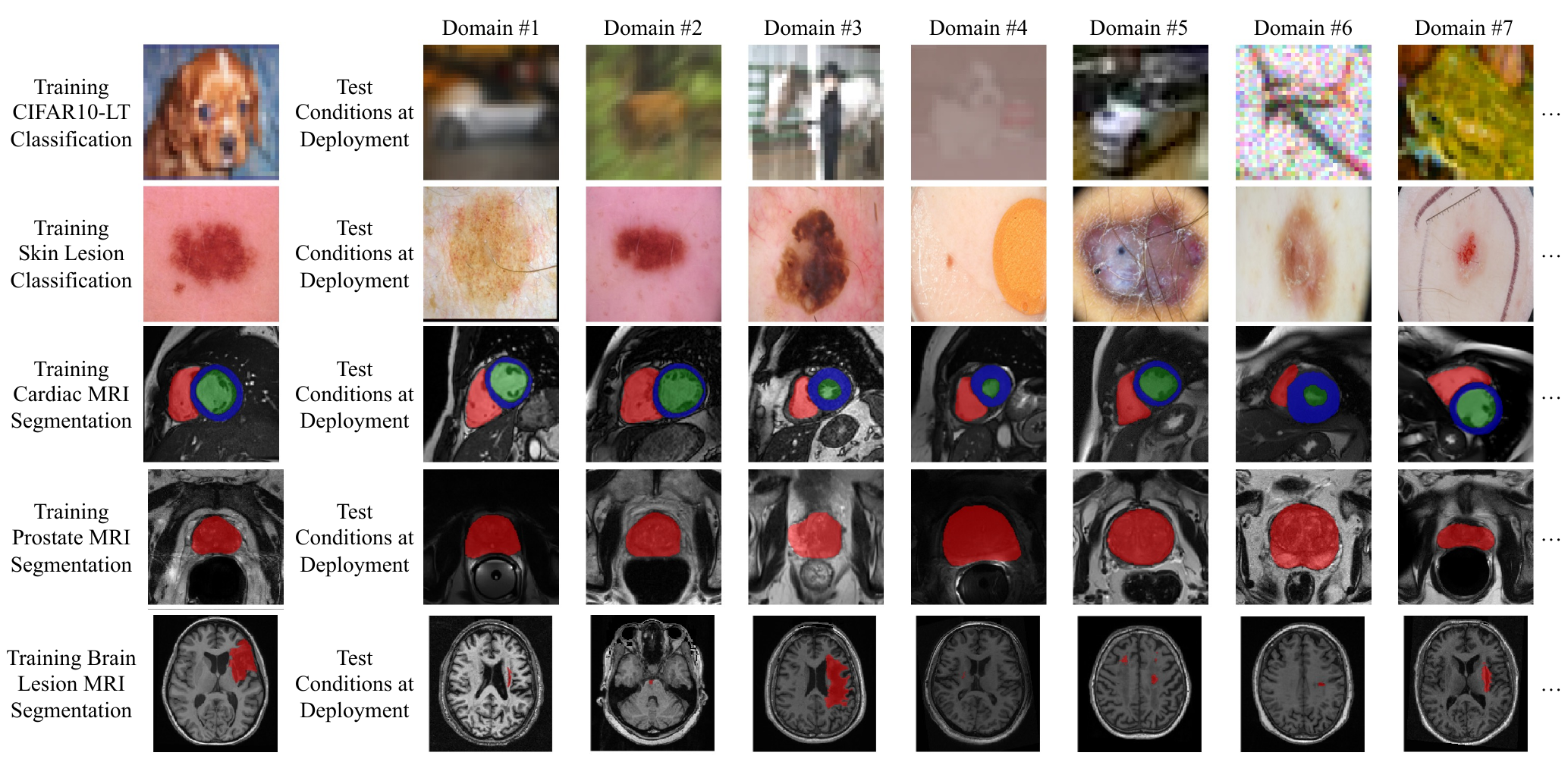}
   \caption{In real-world scenarios, significant discrepancies often exist between training and testing environments, affecting the distribution of acquired images. For instance, in natural images, such discrepancies can be caused by image corruptions like blur or noise~\citep{hendrycks2019benchmarking}. In medical images, these can arise from scanner effects or variations in imaging protocols~\citep{castro2020causality}. For segmentation tasks, regions of interest are highlighted with different colors.
   }
   \label{fig:data_visual}
\end{figure*}

\begin{figure*}[ht]
  \centering
   \includegraphics[width=\linewidth]{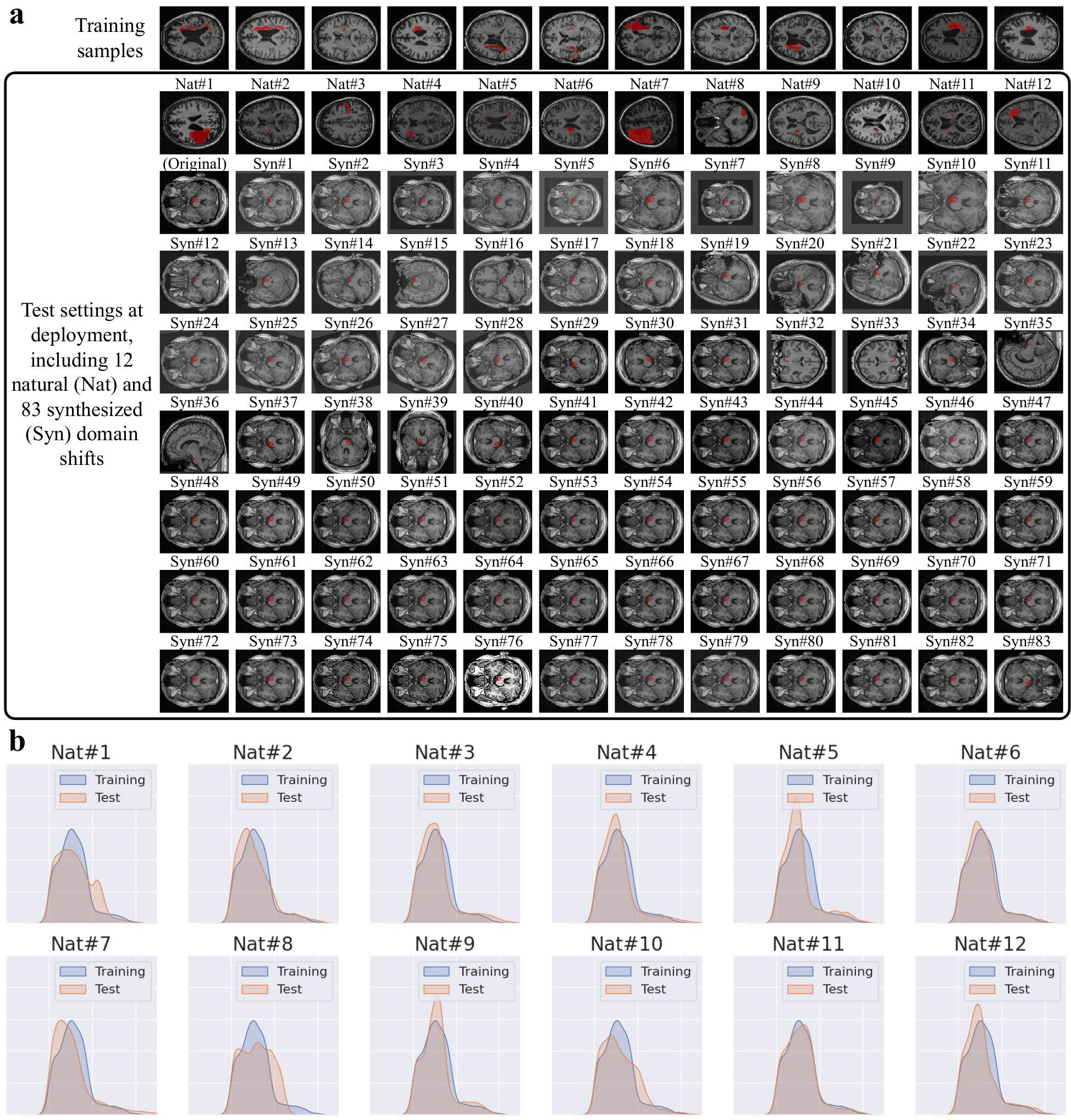}
   \caption{Examples of brain lesion segmentation datasets.
   \textbf{(a)}: Visualization of Brain MR images of In-Domain and Out-Domain Examples. Top: Brain MR images used for training. Brain lesion regions are highlighted with red color. Bottom: Brain MR images with natural and synthesized domain shifts.
   \textbf{(b)}: Intensity histograms of training and test data for 12 test conditions. 
   }
   \label{fig:example}
\end{figure*}

\begin{figure*}[ht]
  \centering
   \includegraphics[width=\linewidth]{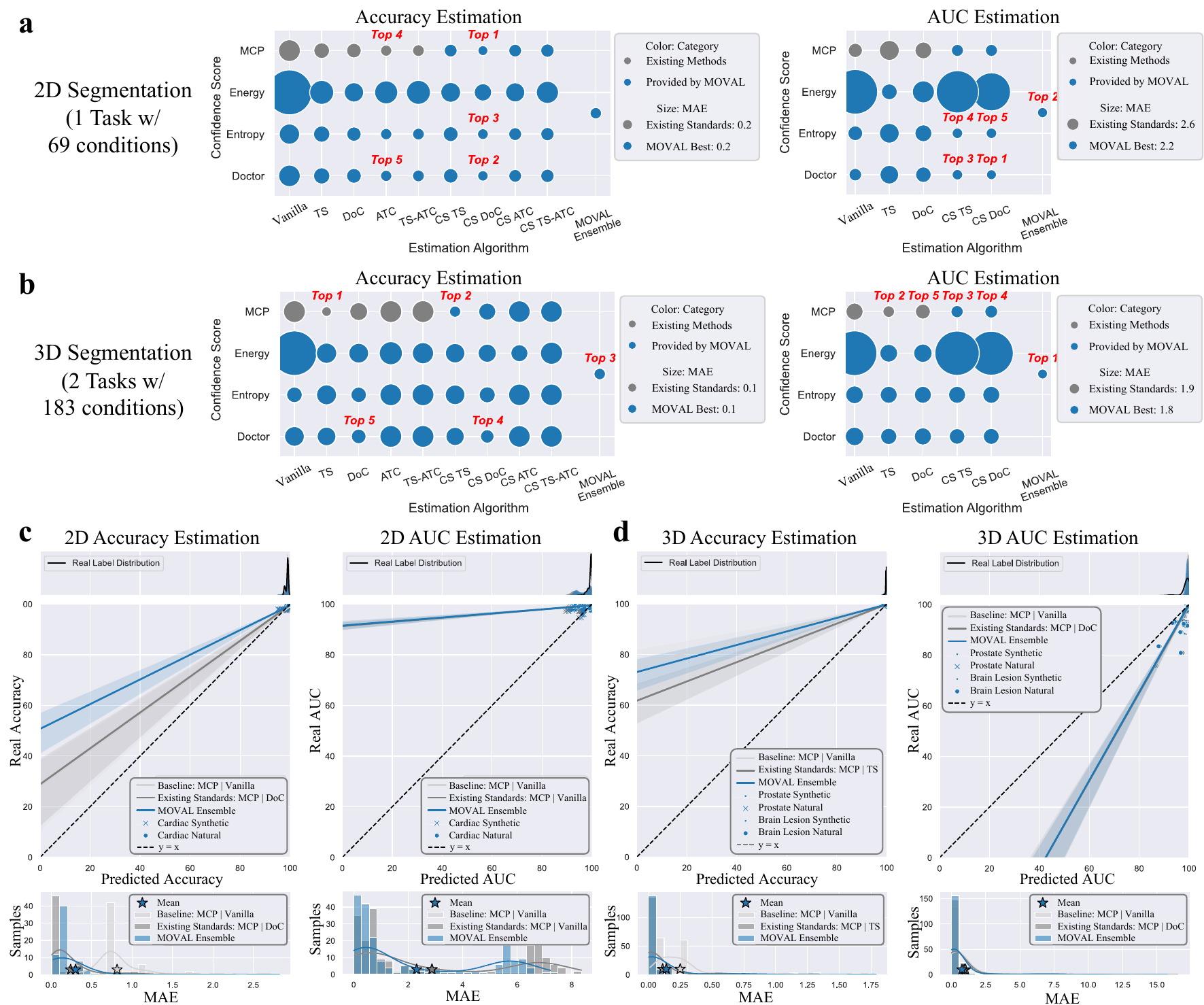}
   \caption{Accuracy and AUC are less useful for segmentation tasks.
   \textbf{(a, b)}: MAE between real metrics and estimated results based on different estimation methods for 2D and 3D segmentation tasks. We also report the average MAE for natural and synthetic domain shifts here. The smaller the MAE, the better. 
   \textbf{(c, d)}: Top: Predicted and real label distribution. We benchmark MOVAL ensemble (blue) against MCP baselines (light gray) and the best-performing existing method (dark gray). Middle:  Predicted versus real metrics for 2D and 3D segmentation tasks. Bottom: Histogram of MAE for different estimation methods. As the class ratios are extremely imbalanced for segmentation tasks, accuracy and AUC are less informative as they are close to 100$\%$.
   }
   \label{fig:detailed_seg_acc}
\end{figure*}

\begin{figure*}[ht]
  \centering
   \includegraphics[width=\linewidth]{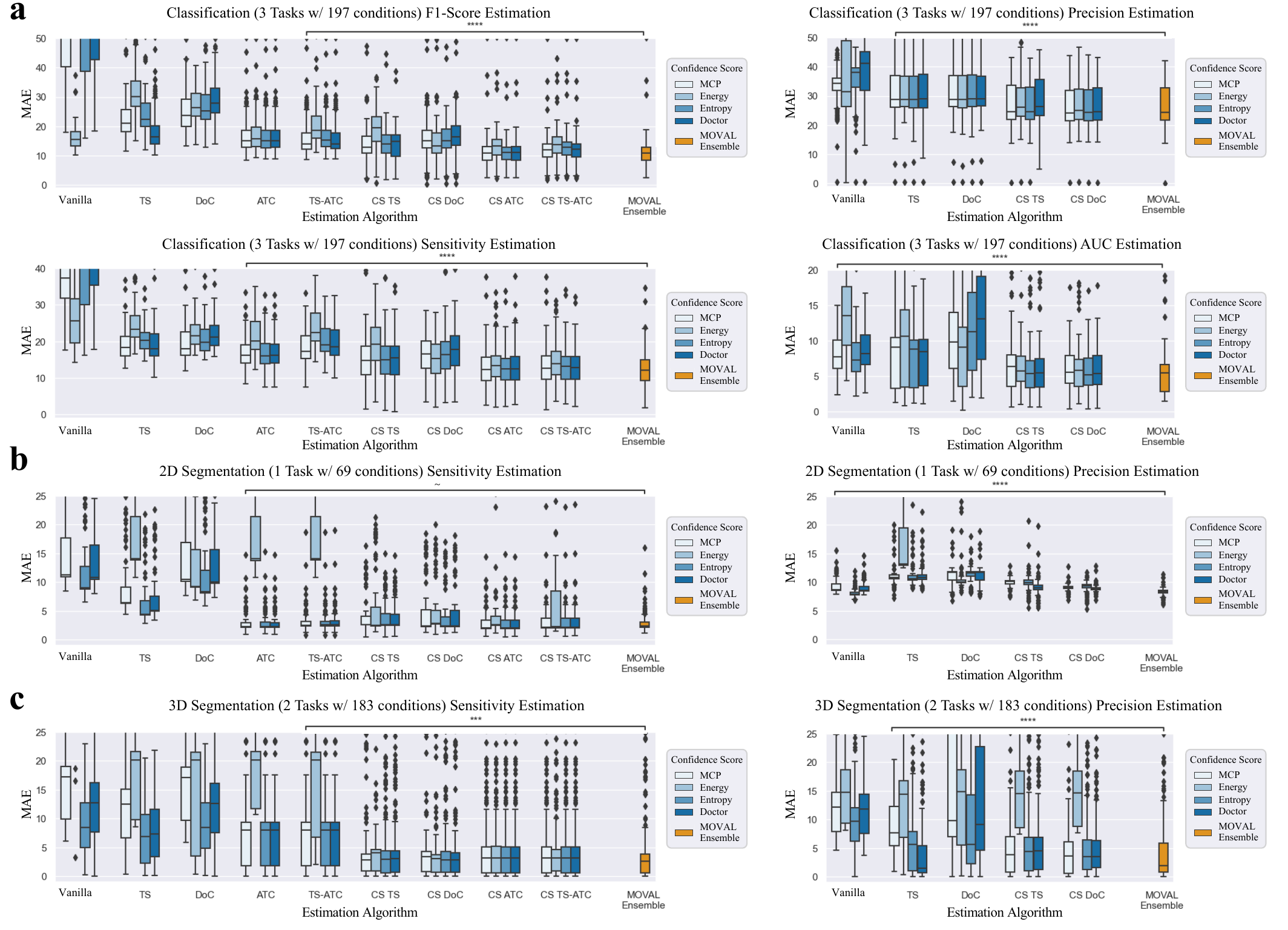}
   \caption{MOVAL Ensemble achieves significant improvements on major evaluation metrics.
   \textbf{(a, b, c)}: MAE comparison between proposed algorithms and existing methods (Existing best method versus MOVAL Ensemble, paired t-test,  $T(196)=9.54, p=6.04\times10^{-18}$ for F1-Score estimation of classification tasks, $T(196)=4.97, p=1.43\times10^{-6}$ for precision estimation of classification tasks, $T(196)=7.80, p=3.54\times10^{13}$ for sensitivity estimation of classification tasks, $T(196)=5.10, p=8.18\times10^{-7}$ for AUC estimation of classification tasks, $T(68)=-0.26, p=0.80$ for sensitivity estimation of 2D segmentation tasks, $T(68)=5.42, p=8.54\times10^{-7}$ for precision estimation of 2D segmentation tasks, $T(182)=3.96, p=1.09\times10^{-4}$ for sensitivity estimation for 3D segmentation tasks and $T(182)=6.79, p=1.53\times10^{-10}$ for precision estimation for 3D segmentation tasks.
   }
   \label{fig:full_res}
\end{figure*}

\begin{figure*}[ht]
  \centering
   \includegraphics[width=\linewidth]{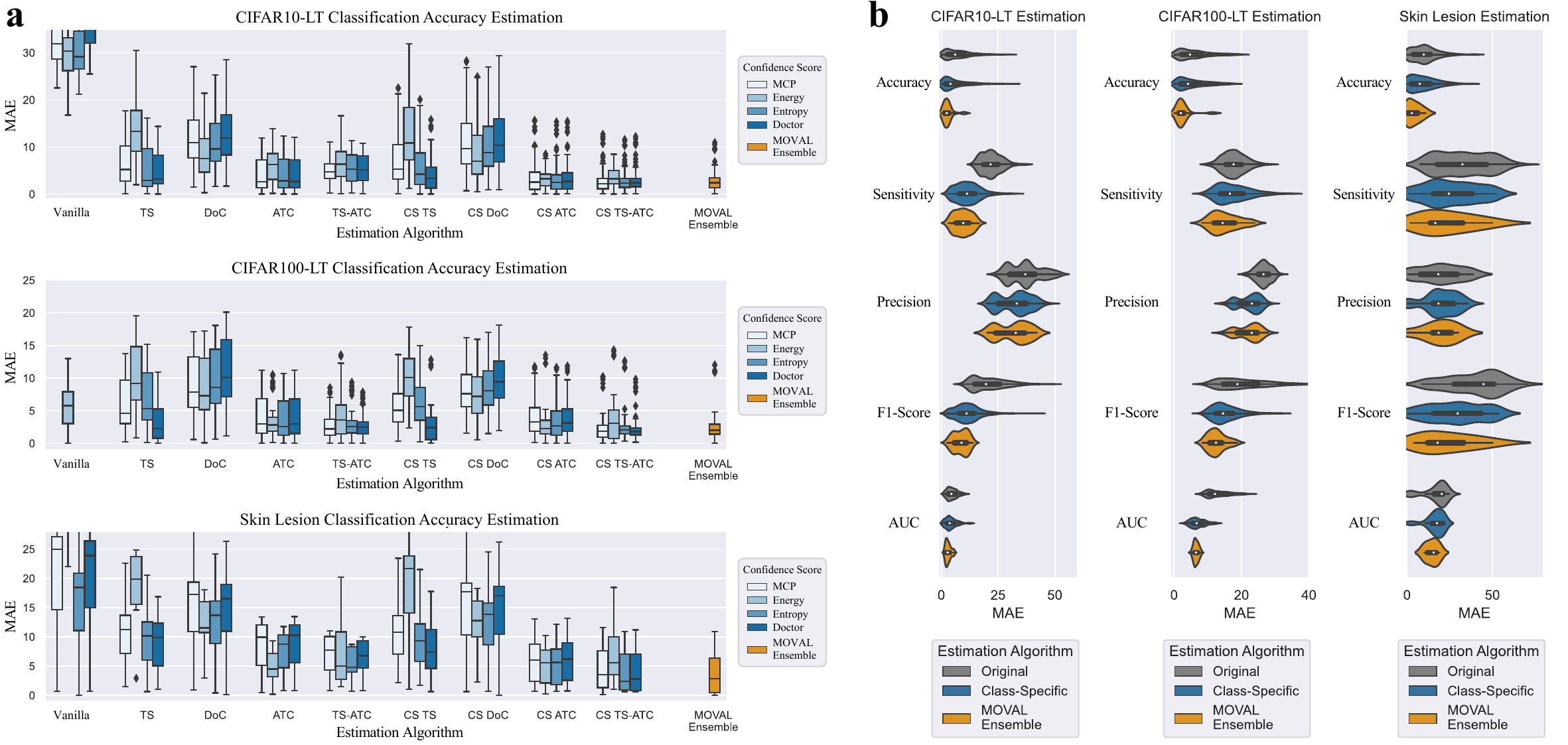}
   \caption{MOVAL Ensemble achieves state-of-the-art performance for several classification tasks.
   \textbf{(a)}: MAE comparison between proposed algorithms and existing methods for different classification tasks.
   \textbf{(b)}: MAE comparison between the original estimation methods and our proposed class-specific variants for different classification tasks.
   }
   \label{fig:detailed_cls}
\end{figure*}

\begin{figure*}[ht]
  \centering
   \includegraphics[width=\linewidth]{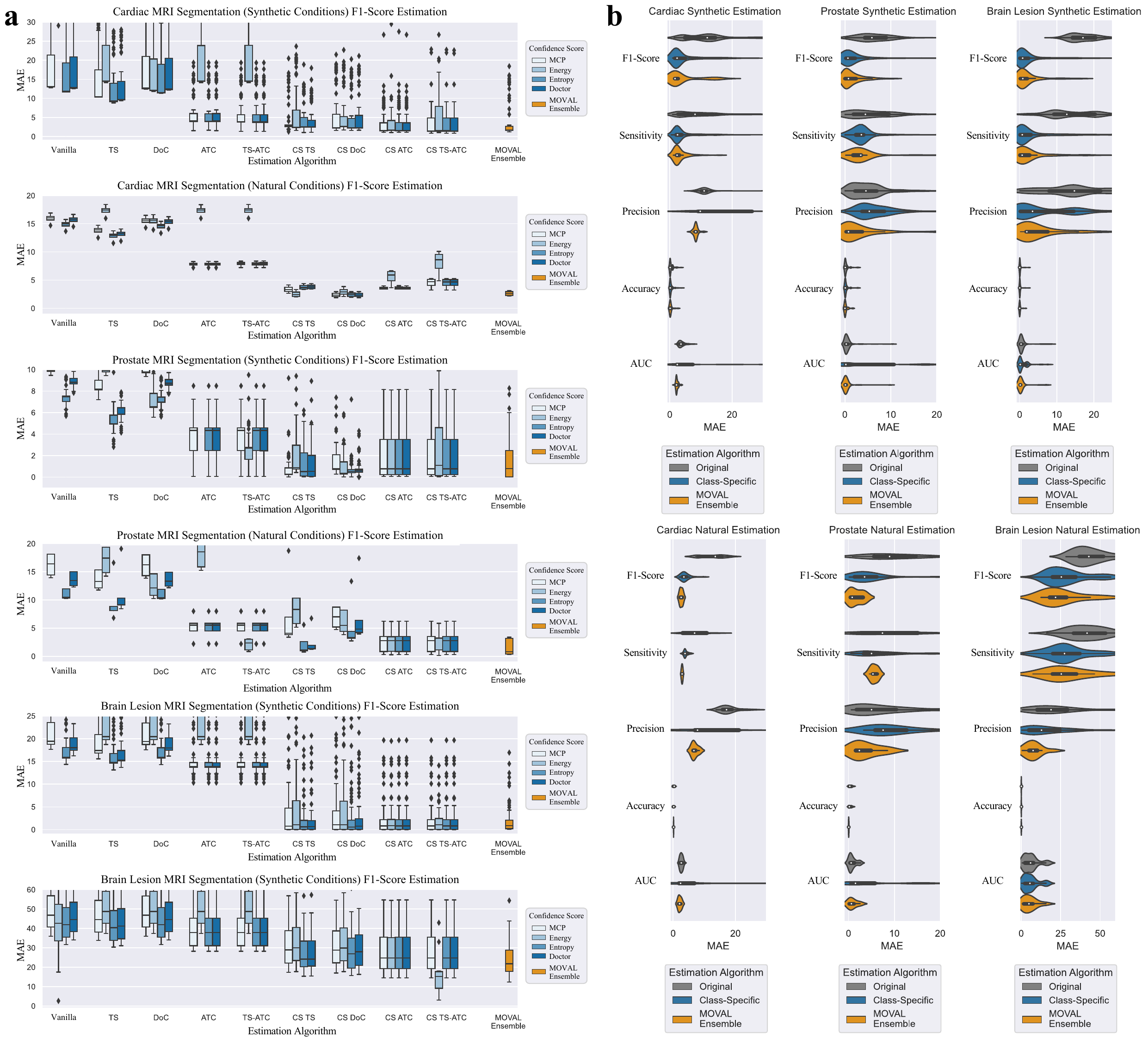}
   \caption{MOVAL Ensemble achieves state-of-the-art performance for several segmentation tasks.
   \textbf{(a)}: MAE comparison between proposed algorithms and existing methods for different segmentation tasks.
   \textbf{(b)}: MAE comparison between the original estimation methods and our proposed class-specific variants for different segmentation tasks.
   }
   \label{fig:detailed_seg}
\end{figure*}

\begin{figure*}[ht]
  \centering
   \includegraphics[width=\linewidth]{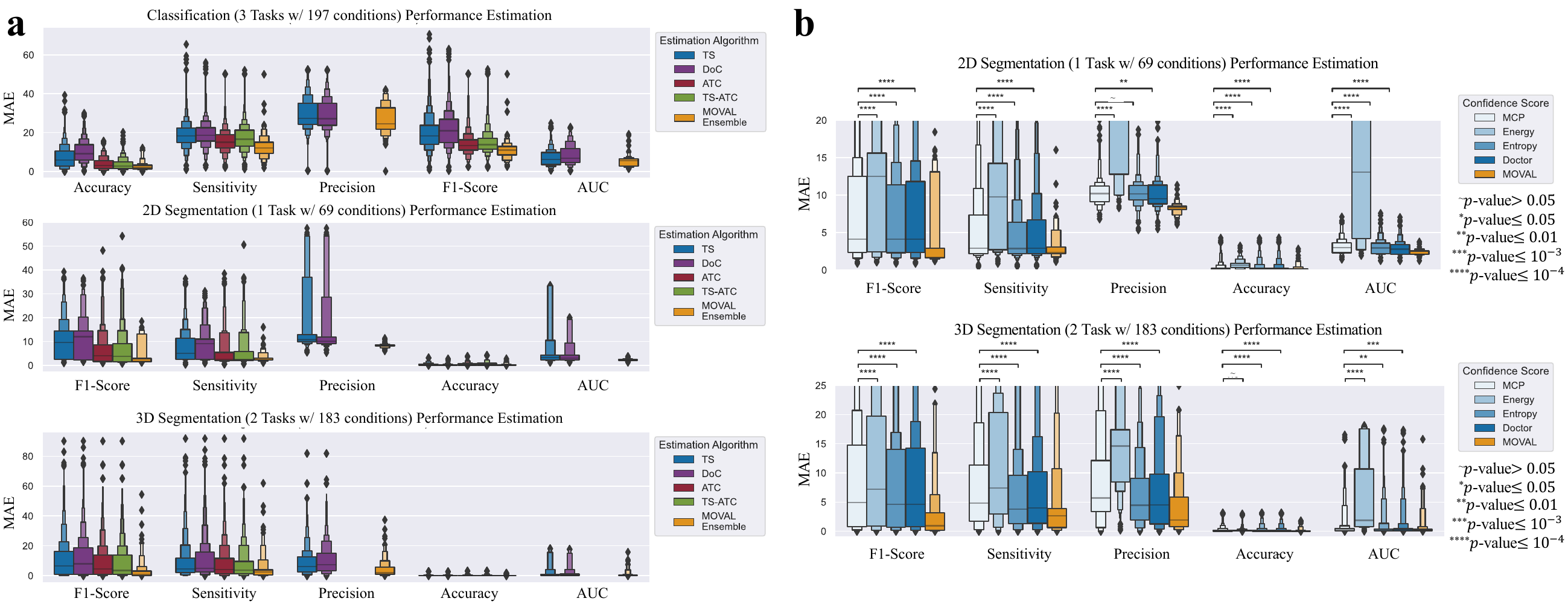}
   \caption{MOVAL Ensemble combines advanced calibration and confidence scores for better performance.
   \textbf{(a)}: MAE comparison between estimation algorithms based on different confidence calibration algorithms.
   \textbf{(b)}: MAE comparison between estimation algorithms based on MCP and other confidence scores for segmentation tasks.
   }
   \label{fig:confidence}
\end{figure*}

\begin{figure*}[ht]
  \centering
   \includegraphics[width=\linewidth]{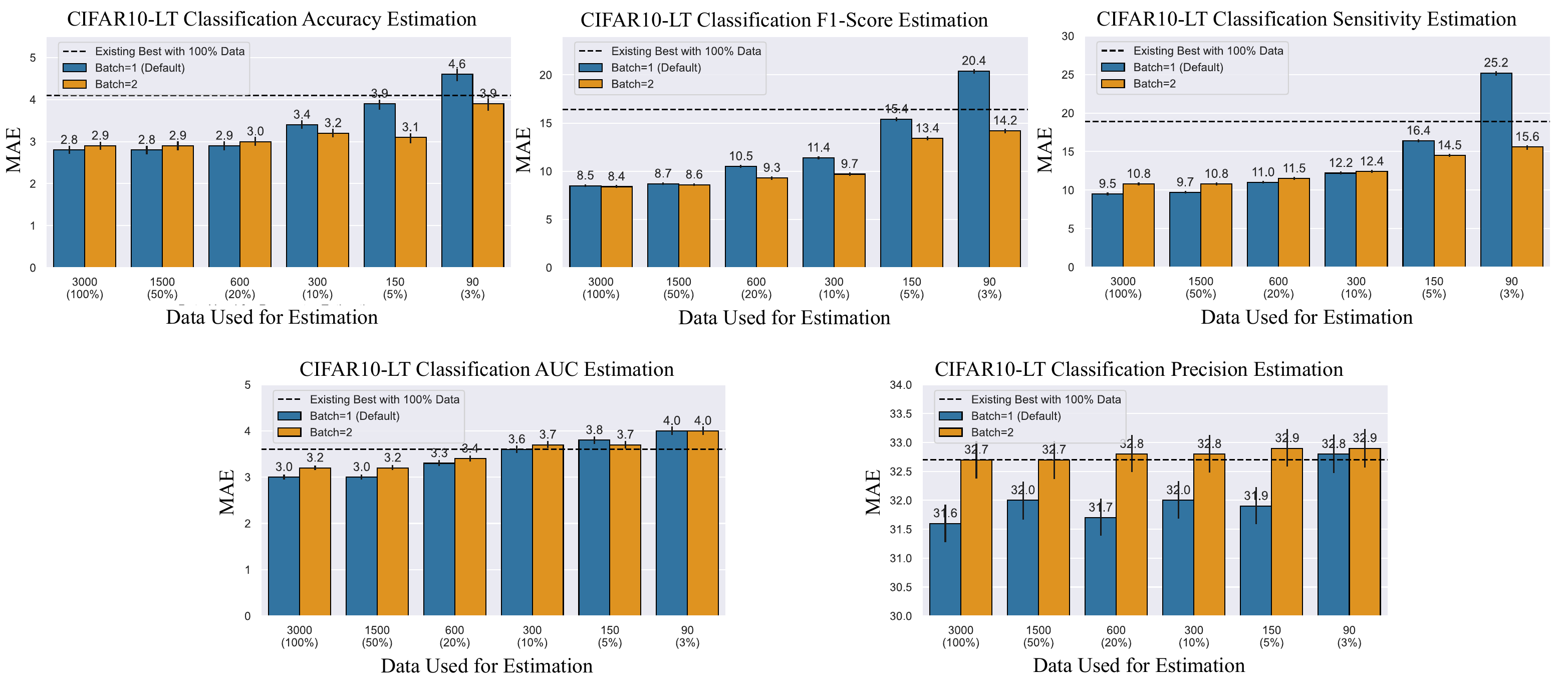}
   \caption{Group-wise calibration enhances the robustness of the MOVAL Ensemble against a reduction in validation samples.
   The proposed group-wise calibration method generally requires fewer validation samples because it has fewer learnable parameters. However, it may negatively impact the estimation of false positive predictions, which require class-specific adjustments. This could potentially affect the estimation of AUC and precision when data are sufficient. The standard error is shown alongside the bar plots.
   }
   \label{fig:group_cal}
\end{figure*}

\begin{figure*}[ht]
  \centering
   \includegraphics[width=\linewidth]{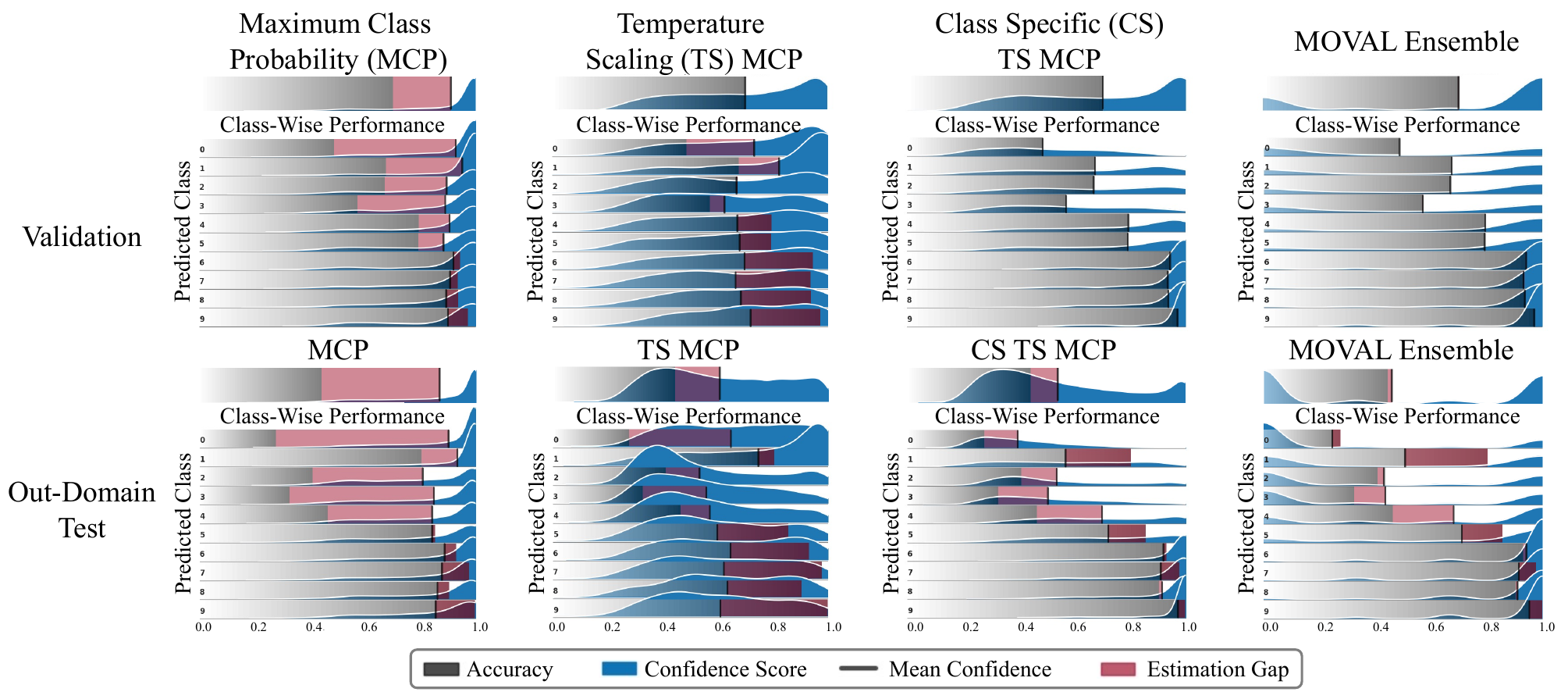}
   \caption{\textcolor{black}{Confidence score distributions of different calibration methods on CIFAR10-LT.
   For each class and calibration method, we show the estimated accuracy (gray gradient bar), the true accuracy (black tick), and the gap between them (red), together with the underlying confidence-score distribution (blue). Confidence distributions are shown as kernel density estimates (KDE) of the per-sample confidence values, for visualization only, to make the shift in confidence across classes easier to see. MOVAL Ensemble, which combines multiple calibrated confidence scores, shows a distinctly bimodal distribution concentrated near 0 and 1, providing a more distinct indicator of prediction correctness than the single-score methods.}}
   \label{fig:conf_kde}
\end{figure*}

\clearpage
\newpage

\section{Supplementary Note}
\label{sec_supp_note}

\subsection{\textcolor{black}{Group-Wise and Class-Specific Metrics Estimation}}

Apart from accuracy, we can also estimate group-wise metrics that measure model performance from different perspectives based on the predicted probability $\boldsymbol{p}_i$ instead of the confidence score $q_i$. This is because these metrics also incorporate information from non-maximum classes. This approach is related to efforts that formulate these metrics as loss functions~\citep{gao2013one, milletari2016v}. However, to enable estimation without ground truth at test time, we use the predicted label $y^{\prime}_i$ as a substitute for the ground truth in our calculations. This can be viewed as an extension of the confidence-based accuracy framework, which assumes that all predictions are correct at some level of confidence. Specifically, the estimated F1-Score is defined as:

\begin{equation}
\Phi'_j = \frac{2\sum_{i=1}^N\mathbbm{1}_{ij}p_{ij}}{\sum_{i=1}^N\mathbbm{1}_{ij}+\sum_{i=1}^Np_{ij}},
    \label{eq:f1c}
\end{equation}

\noindent \textcolor{black}{where} $\mathbbm{1}_{ij} = \mathbbm{1}{(y^{\prime}_i=j)}$ is 1 if the predicted most probable class is $j$ and $0$ otherwise. In a similar manner, sensitivity and precision can be calculated as follows:

\begin{equation}
    \rho'_j = \frac{\sum_{i=1}^N\mathbbm{1}_{ij}p_{ij}}{\sum_{i=1}^Np_{ij}}, \quad \pi'_j = \frac{\sum_{i=1}^N\mathbbm{1}_{ij}p_{ij}}{\sum_{i=1}^N\mathbbm{1}_{ij}}.
    \label{eq:sensprec}
\end{equation}

The AUC calculation uses the One-vs-Rest strategy, which involves comparing one class against all other classes combined in multi-class problems. Notably, this is equivalent to the One-vs-One strategy for binary classification, which involves comparing every unique pairwise combination. The estimation is given by:

\begin{equation}
    \Lambda'_j = \int_{0}^{1}\text{TPR}'_j(t)d(\text{FPR}'_j(t)),
    \label{eq:auc}
\end{equation}

\noindent \textcolor{black}{where} the true positive rate (TPR) and false positive rate (FPR) are calculated as follows:

\begin{equation}
    \text{TPR}'_j(t) = \frac{\sum_{i=1}^N\mathbbm{1}{(p_{ij}>t)}p_{ij}}{\sum_{i=1}^N\mathbbm{1}{(p_{ij}>t)}}, \quad \text{FPR}'_j(t) = \frac{\sum_{i=1}^N\mathbbm{1}{(p_{ij} \leq t)} \sum_{k \neq j}p_{ik}}{\sum_{i=1}^N\mathbbm{1}{(p_{ij} \leq t)}}.
    \label{eq:tprfpr}
\end{equation}

\textcolor{black}{Throughout, $i$ indexes a sample for classification and a pixel (2D) or voxel (3D) for segmentation; for segmentation, $z$ additionally indexes the case (volume), so the per-pixel probability is written $p_{zij}$. Eqs.~\ref{eq:f1c}-\ref{eq:tprfpr} are written with the raw softmax output ($p_{ij}$ for classification, $p_{zij}$ for segmentation) to introduce the estimators; as described below, all reported estimates are computed after MOVAL's class-wise calibration, i.e., with the raw probability replaced by its calibrated counterpart ($\hat{p}_{ij}$ or $\hat{p}_{zij}$).} In this manner, we can derive all the metrics using only the network outputs. However, like accuracy estimation, metrics derived from uncalibrated probabilities may be inaccurate. Although some studies suggest that accuracy-calibrated models perform better~\citep{nannyml}, we believe that this approach may be suboptimal, as accuracy calibration does not necessarily improve all metrics. 

Instead, we offer a novel solution for calibrating confidence to align with these metrics. For each metric, we optimize a specific model tailored to that metric. It directly adjusts parameters to ensure that the estimated metrics align closely with the actual metrics on the validation set. This alignment is made possible by our unique MOVAL framework, which incorporates class-wise calibration with class-specific parameters.

After obtaining the calibrated probability $\hat{\boldsymbol{p}}_i$, the optimization process closely follows the accuracy calibration procedure, as class-specific parameters can naturally be extended to other metrics. Using F1-Score as an example, we perform separate calibration for the $j$'th class by adjusting all $\boldsymbol{p}_i$ predicted as $j$ to $\hat{\boldsymbol{p}}_i$ with class-specific parameters, ensuring that the estimation from Eq.~\ref{eq:f1c} using $\hat{\boldsymbol{p}}_i$ aligns with the actual metrics. This process is carried out for each class individually. All the other group-wise metrics, including sensitivity, precision, and AUC, can be calibrated similarly using Eq.~\ref{eq:sensprec} and Eq.~\ref{eq:auc}, respectively. Regarding AUC, we compute five thresholds at equal intervals based on the range of probabilities $\hat{p}_{ij}$ for all samples of class $j$ (denoted as $t$ in Eq.~\ref{eq:tprfpr}). Using these thresholds, we then calculate the areas under the $\text{FPR}'_j$–$\text{TPR}'_j$ curve to estimate the AUC of class $j$, i.e, $\Lambda'_j$.

\subsection{Network Configurations}

Here, we detail the network configurations and training procedures used to develop the model for classification and segmentation tasks. This includes specifics on neural network training and advanced strategies commonly employed to tackle class imbalance and domain generalization. These methods are established techniques widely adopted as best practices.

\subsubsection{Classification}

\noindent\textbf{Network architecture.} We train a ResNet-32~\citep{he2016deep} for CIFAR10 and CIFAR100 experiments following~\citep{cao2019learning}. We train a ResNet-50 for skin lesion lesion classification following~\citep{russakovsky2015imagenet, marrakchi2021fighting}.

\noindent\textbf{Network training.} We trained for 200 epochs on the CIFAR experiments and 20,000 epochs for skin lesion classification. The networks are trained using stochastic gradient descent (SGD). The learning rate was set to 0.1 and reduced using step decay. The images are normalized based on the statistics of the training dataset. For the CIFAR baseline experiments, we employed data augmentation techniques such as random flipping and color transformations. For the skin lesion classification baseline experiments, we applied data augmentation strategies including cropping to $224\times224$ and random flipping.

Given a sample with label $y \in \{1, 2, \ldots, c\}$, the neural network generates $\boldsymbol{z}$, which is subsequently transformed into a probability vector $\boldsymbol{p}$, where the $j$th element $p_j$ represents the probability of the $j$th class. Unless specified otherwise, the network is trained using cross-entropy (CE) loss, which is defined as:

\begin{equation}
    \mathcal{L}_{CE} = -\log p_{y} = -\log \frac{e^{z_y}}{\sum_k e^{z_k}}.
    \label{eq:cls_ce}
\end{equation}

\noindent\textbf{Balanced training \uppercase\expandafter{\romannumeral1}: label-distribution-aware margin (LDAM) loss~\citep{cao2019learning}.} LDAM applies unequal margins to the logits of different classes to enhance the robustness of predictions for minority classes. The calculation is defined as:

\begin{equation}
    \mathcal{L}_{CE}^{LDAM} = -\log \frac{e^{z_y-\Delta_y}}{e^{z_y-\Delta_y}+\sum_{k \neq y} e^{z_k}},
    \label{eq:ldam}
\end{equation}

\noindent \textcolor{black}{where} the margins for different classes are calculated as:

\begin{equation}
    \Delta_j = \frac{C}{n_j^{1/4}},
    \label{eq:ldam_margin}
\end{equation}

\noindent \textcolor{black}{where} $C$ is a hyper-parameter and $n_j$ is the number of samples for class $j$. This encourages the network to focus on minority classes, as they require larger margins.

\noindent\textbf{Balanced training \uppercase\expandafter{\romannumeral2}: deferred re-weighting (DRW) weighting schedule~\citep{cao2019learning}.} DRW adjusts the weight of samples belonging to different classes. To learn good representations and achieve unbiased performance across classes, it applies uniform weighting at the beginning and then reweights the samples based on class frequency (i.e., minority classes receive larger weights) in the later stages of network training. Specifically, it optimizes with: 

\begin{equation}
    \mathcal{L}_{CE}^{DRW} = -w_y\log p_{y},
    \label{eq:cls_ce_weight}
\end{equation}

\noindent \textcolor{black}{where} $w_y$ is the weight for class $y$, which starts at 1 and later changes to $1/n_y$.

\noindent\textbf{Robust training \uppercase\expandafter{\romannumeral1}: cutout~\citep{devries2017improved}.} Cutout is a data augmentation strategy that modifies the input sample by randomly cropping out a square in the image and replacing it with black patches. This helps networks become robust to local features and improves generalization. Cutout is commonly adopted in neural networks to enhance model generalizability. We found that it significantly improves model accuracy on in-domain samples but fails to enhance performance on most out-of-domain samples, possibly because it causes neural networks to rely more on the original intensity distributions. We choose to report its performance because it is a default technique for training neural networks in classification.

\noindent\textbf{Robust training \uppercase\expandafter{\romannumeral2}: RandAugment~\citep{cubuk2020randaugment}.} RandAugment is a data augmentation strategy that applies multiple stacked transformations with predefined magnitudes, unlike baseline methods. It introduces transformations from a larger collection, improving network performance under these conditions.

\subsubsection{Segmentation}

\noindent\textbf{Network architecture.} We train a 3D U-Net~\citep{ronneberger2015u} for segmentation using the nnU-Net~\citep{isensee2021nnu} framework, with a base feature map size of 30 and 4 downsampling steps at the bottleneck.

\noindent\textbf{Network training.} 
The networks are trained using patches sampled from whole images, as the images are typically too large to be input into the network due to memory limitations. For cardiac MRI segmentation, we train using patches of size $128 \times 128 \times 8$ with a batch size of 30. For prostate MRI segmentation, we use patches of size $64 \times 64 \times 32$ with a batch size of 32. For brain lesion segmentation, we train with patches of size $128 \times 128 \times 128$ and a batch size of 2.

We train neural networks for 1000 epochs, in each epoch we train with patches for 100 iterations. The networks are trained using SGD, with learning rates linearly decreased from 0.01 to 0. By default, we apply Z-score normalization to MR images, ensuring that each MRI volume has a mean of zero and unit variance. Given an image patch, the label $y$ corresponds to the center pixel and is represented as a one-hot vector. In this vector, $e_j$ equals 1 if $j$ is the true class, and 0 otherwise. For baseline experiments, the networks are trained with a balanced sum of cross-entropy (CE) loss and sample-wise soft DSC loss, which are defined as:

\begin{equation}
    \mathcal{L}_{CE} = -\sum_{j=1}^c e_j \log p_j = -\sum_{j=1}^ce_j \log \frac{e^{z_j}}{\sum_k e^{z_k}},
    \label{eq:seg_ce}
\end{equation}

\noindent \textcolor{black}{and}

\begin{equation}
    \mathcal{L}_{DSC}= \sum_{j=1}^c \frac{(1-e_j)p_j + e_j(1-p_j)}{p_j + e_j}.
    \label{eq:seg_dsc}
\end{equation}

\noindent\textbf{Balanced training \uppercase\expandafter{\romannumeral1}: asymmetric large margin loss~\citep{li2020analyzing}.} Asymmetric large margin loss, similar to LDAM, is better suited for segmentation tasks and has been validated with soft DSC. Specifically, we represent the rare class with a one-hot vector $\boldsymbol{r}$, where the $j$th element $r_j \in \{0, 1\}$ is 1 if class $j$ is rare. For example, for brain lesion segmentation, we set $[0, 1]$ to indicate that the brain lesion class is considered rare. Based on this, the CE and soft DSC loss are modified to:

\begin{equation}
    \mathcal{L}_{CE}^{ALM} = -\sum_{j=1}^c e_j \log h_j,
    \label{eq:ce_asy_large}
\end{equation}

\noindent \textcolor{black}{and}

\begin{equation}
    \mathcal{L}_{DSC}^{ALM} = \sum_{j=1}^c \frac{(1-e_j)h_j + e_j(1-h_j)}{h_j + e_j},
    \label{eq:dsc_asy_large}
\end{equation}

\noindent \textcolor{black}{where} $h_j$ is the modified probability:

\begin{equation}
    h_j = \frac{e^{z_j}-e_j\Delta_j}{\sum_k e^{z_k-e_j\Delta_j}},
    \label{eq:asy_large_p}
\end{equation}

\noindent \textcolor{black}{where}

\begin{equation}
    \Delta_j = r_j C',
    \label{eq:asy_large_delta}
\end{equation}

\noindent \textcolor{black}{where} $C'$ is a hyper-parameter. The neural network is then optimized using the equal sum of the CE and soft DSC losses, and the same applies to the later asymmetric focal loss.

\noindent\textbf{Balanced training \uppercase\expandafter{\romannumeral2}: asymmetric focal loss~\citep{li2020analyzing}.} Asymmetric focal loss builds upon the concept of focal loss but is better suited for long-tail learning, as it removes the attenuation term for rare classes. This adjustment helps prevent the network from under-fitting minority classes. Specifically, the CE and soft DSC losses are modified as follows:

\begin{equation}
   \mathcal{L}_{CE}^{AF} = -\sum_{j=1}^c r_j e_j \log p_j - \sum_{j=1}^c (1-r_j)(1-p_j)^{\gamma} e_j \log p_j,
    \label{eq:ce_asy_focal}
\end{equation}

\noindent \textcolor{black}{and}

\begin{equation}
    \mathcal{L}_{DSC}^{AF} = \sum_{j=1}^c \frac{(1-e_j)p_j + r_je_j(1-p_j)}{p_j + e_j} + \sum_{j=1}^c \frac{(1-e_j)p_j + (1-r_j)(1-p_j)^{\gamma}e_j(1-p_j)}{p_j + e_j}.
    \label{eq:dsc_asy_focal}
\end{equation}

\noindent\textbf{Robust training \uppercase\expandafter{\romannumeral1}: global intensity non-linear augmentation (GIN) technique~\citep{ouyang2022causality}.} GIN is a random kernel-based domain generalization method that transforms images into random styles. Given an image patch $\boldsymbol{x}$, it transforms it into $\tilde{\boldsymbol{x}}$ using:

\begin{equation}
    \tilde{\boldsymbol{x}} = \frac{\beta g(\boldsymbol{x} + (1-\beta)\boldsymbol{x})}{\|\beta g(\boldsymbol{x} + (1-\beta)\boldsymbol{x}) \|_F} \| \boldsymbol{x} \|_F,
    \label{eq:GIN}
\end{equation}

\noindent \textcolor{black}{where} $\| \cdot \|_F$ denotes the Frobenius norm, $g(\cdot)$ represents a shadow CNN with randomly initialized parameters, and $\beta$ is a parameter sampled from a uniform distribution. In this way, the network can be robust to various kinds of intensity transformations. We find that GIN helps neural networks generalize better across different appearances but performs worse with spatial transformations. Therefore, to ensure a fair comparison, we exclude all spatial transformation-based synthetic domain shifts when evaluating advanced training strategies, including GIN.

\noindent\textbf{Robust training \uppercase\expandafter{\romannumeral2}: mixup~\citep{zhang2017mixup}.} Mixup extends the distributions of the training dataset by combining two individual samples. In addition to $\{\boldsymbol{x}, \boldsymbol{e}\}$, it requires another pair $\{\boldsymbol{x}', \boldsymbol{e}'\}$ and creates a new sample with:

\begin{equation}
    \tilde{\boldsymbol{x}} = \lambda \boldsymbol{x} + (1-\lambda) \boldsymbol{x}', \tilde{\boldsymbol{e}} = \lambda \boldsymbol{e} + (1-\lambda) \boldsymbol{e}', 
    \label{eq:mixup}
\end{equation}

\noindent \textcolor{black}{where} $\lambda$ is a parameter sampled from a  beta distribution.

\subsection{Fine-Grained Performance Analysis}

To evaluate the proposed algorithms, we compare the estimated results with the actual metrics and compute the MAE. For class-wise metrics beyond accuracy, we report the average MAE across different classes. To assess the significance between methods, we perform a paired t-test on the MAE values obtained under different conditions using each method. We exclude the evaluation of classes that are absent during testing, particularly in the task of skin lesion classification, where some classes may be missing at test time. For segmentation, we exclude the background class when evaluating the model's performance.

For a comprehensive analysis of the estimation results, we provide examples at the class, instance, and pixel levels. For class-wise performance, we group the confidence scores by the predicted class, calculate their averages, and compare these averages to the corresponding accuracy of the samples. In instance-wise analysis for segmentation tasks, we evaluate performance on a case-by-case basis rather than as a whole. For pixel-wise analysis, we visualize the calibrated confidence scores adjusted by the F1-Score, highlighting uncertain regions as \(1 - \hat{q}_i\). For 3D confidence score visualization, we display pixels with uncertainty greater than 0.1 and set the cluster transparency to 0.005.

\clearpage
\newpage

\section{Supplementary Tables}
\label{sec_supp_tables}

\begin{table}[ht]
\centering
\resizebox{\linewidth}{!}{%
\begin{tabular}{c<{\centering}|c<{\centering}c<{\centering}c<{\centering}c<{\centering}c<{\centering}m{18mm}<{\centering}m{16mm}<{\centering}c<{\centering}}
\toprule
 & Accuracy & F1-Score & Sensitivity & Precision & AUC & Advanced Training & Reduced Data & Total \\
 \midrule
Classification & 197 & 197 & 197 & 197 & 197 & 380 & 2375 & 3740 \\
2D Segmentation & 69 & 69 & 69 & 69 & 69 & 0 & 276 & 621 \\
3D Segmentation & 183 & 183 & 183 & 183 & 183 & 216 & 1080 & 2211 \\
Total & 449 & 449 & 449 & 449 & 449 & 596 & 3731 & 6572 \\
\bottomrule
\end{tabular}
}
\caption{Summary of experimental results covered in the main body.}
\label{tabs:exp}
\end{table}

\begin{table}[ht]
\resizebox{\linewidth}{!}{%
\centering
\begin{tabular}{m{10mm}<{\centering}|m{18mm}<{\centering}m{18mm}<{\centering}m{18mm}<{\centering}m{15mm}<{\centering}m{18mm}<{\centering}m{13mm}<{\centering}m{15mm}<{\centering}m{10mm}<{\centering}}
\toprule
 & nv (\%) & mel (\%) & bcc (\%) & df (\%) & bkl (\%) & vasc (\%) & akiec (\%) & Total \\
 \midrule
HAM & 6705 (67) & 1113 (11) & 514 (5) & 115 (1) & 1099 (11) & 142 (1) & 327 (3) & 10015 \\
BCN & 4206 (34) & 2857 (23) & 2809 (23) & 124 (1) & 1138 (9) & 111 (1) & 1168 (9) & 12413 \\
VIE & 4331 (99) & 34 (1) & 0 & 0 & 0 & 0 & 0 & 4365 \\
MSK & 2202 (62) & 826 (23) & 30 (1) & 5 ($<$1) & 470 (13) & 0 & 7 ($<$1) & 3540 \\
UDA & 408 (67) & 193 (31) & 3 ($<$1) & 2 ($<$1) & 7 (1) & 0 & 0 & 613 \\
OTH & 4523 (55) & 1669 (20) & 513 (6) & 124 (2) & 889 (11) & 95 (1) & 388 (5) & 8201 \\
D7P & 1150 (60) & 501 (26) & 84 (4) & 40 (2) & 90 (5) & 58 (3) & 0 & 1923 \\
PH2 & 160 (80) & 40 (20) & 0 & 0 & 0 & 0 & 0 & 200 \\
\bottomrule
\end{tabular}
}
\caption{Dataset statistics of skin lesion classification datasets used in this study. nv: Melanocytic nevi, mel: Melanoma, bcc: Basal cell carcinoma, df: Dermatofibroma, bkl: Benign keratosis, vasc: Vascular skin lesions, akiec: Actinic Keratoses (Solar Keratoses) and Intraepithelial Carcinoma (Bowen’s disease).}
\label{tabs}
\end{table}

\begin{table*}[ht]
\centering
 \resizebox{\linewidth}{!}{%
    \begin{tabular}{ccccccccccccc}
    \toprule
        \multicolumn{3}{c}{\multirow{3}{*}{\shortstack{Evaluation \\  Metric and \\ Dataset}}} & &  \multicolumn{9}{c}{Confidence Score}  \\
        & & & \multirow{2}{*}{\shortstack{Estimation \\ Algorithm}} & \multicolumn{4}{c}{Class-Agnostic} & \multicolumn{4}{c}{Class-Specific} & \multirow{2}{*}{\shortstack{MOVAL \\ Ensemble}}  \\
        & & & & MCP & Energy & Entropy & Doctor & MCP & Energy & Entropy & Doctor \\
        \cmidrule(r){1-3} \cmidrule(r){4-4} \cmidrule(lr){5-8} \cmidrule(lr){9-12} \cmidrule(lr){13-13}
        
        \multirow{15}{*}{\rotatebox{90}{\shortstack{Sensitivity}}} & \multirow{5}{*}{\rotatebox{90}{\shortstack{CIFAR10}}} & \multirow{5}{*}{\rotatebox{90}{\shortstack{Synthetic-95}}} & Vanilla  & \ms{32.3}{5.2} & \ms{31.4}{3.2} & \ms{30.6}{5.1} & \ms{35.9}{5.7} & --- & --- & --- & --- & \multirow{5}{*}{\msbb{9.5}{3.6}} \\
        & & & TS  & \ms{20.8}{3.7} & \ms{24.6}{5.5} & \ms{21.3}{3.7} & \ms{21.6}{3.5} & \ms{12.2}{5.5} & \ms{15.4}{6.9} & \ms{11.8}{5.1} & \ms{10.9}{4.4} \\
        & & & DoC  & \ms{21.9}{4.2} & \ms{22.9}{4.6} & \ms{21.7}{4.2} & \ms{22.6}{4.5} & \ms{13.8}{6.1} & \ms{13.1}{5.8} & \ms{13.3}{5.8} & \ms{14.5}{6.3} \\
        & & & ATC  & \ms{18.9}{3.0} & \ms{25.0}{3.9} & \ms{19.1}{3.1} & \ms{18.9}{3.0} & \msbb{9.5}{3.8} & \ms{10.8}{4.2} & \msb{9.4}{3.7} & \msb{9.4}{3.7} \\
        & & & TS-ATC  & \ms{21.2}{3.7} & \ms{27.7}{4.9} & \ms{22.9}{3.4} & \ms{22.6}{3.4} & \ms{10.2}{4.4} & \ms{12.0}{4.8} & \ms{10.2}{3.8} & \ms{10.0}{3.6}\\
\bottomrule
    \end{tabular}%
}
\caption{Evaluation of sensitivity and precision estimation for different classification tasks under various types of domain shifts based on MAE. The best and second-best results within the same category are in \textbf{bold} and \underline{underlined} for each configuration.}
\label{tab:cls_sensitivity_precision}
\end{table*}

\begin{table*}[ht]
\centering
 \resizebox{\linewidth}{!}{%
    \begin{tabular}{ccccccccccccc}
    \toprule
        \multicolumn{3}{c}{\multirow{3}{*}{\shortstack{Evaluation \\  Metric and \\ Dataset}}} & &  \multicolumn{9}{c}{Confidence Score}  \\
        & & & \multirow{2}{*}{\shortstack{Estimation \\ Algorithm}} & \multicolumn{4}{c}{Class-Agnostic} & \multicolumn{4}{c}{Class-Specific} & \multirow{2}{*}{\shortstack{MOVAL \\ Ensemble}}  \\
        & & & & MCP & Energy & Entropy & Doctor & MCP & Energy & Entropy & Doctor \\
        \cmidrule(r){1-3} \cmidrule(r){4-4} \cmidrule(lr){5-8} \cmidrule(lr){9-12} \cmidrule(lr){13-13}
        
        \multirow{15}{*}{\rotatebox{90}{\shortstack{F1-Score}}} & \multirow{5}{*}{\rotatebox{90}{\shortstack{CIFAR10}}} & \multirow{5}{*}{\rotatebox{90}{\shortstack{Synthetic-95}}} & Vanilla  & \ms{40.2}{7.4} & \ms{16.0}{2.9} & \ms{38.9}{7.3} & \ms{42.4}{7.7} & --- & --- & --- & --- & \multirow{5}{*}{\msb{8.5}{3.1}} \\
        & & & TS  & \ms{22.3}{6.0} & \ms{28.7}{8.1} & \ms{21.6}{6.0} & \ms{19.4}{5.9} & \ms{11.8}{5.5} & \ms{15.1}{6.8} & \ms{11.3}{5.0} & \ms{10.2}{4.1} \\
        & & & DoC  & \ms{26.4}{7.1} & \ms{24.9}{6.8} & \ms{25.9}{6.9} & \ms{27.3}{7.5} & \ms{14.0}{6.4} & \ms{13.1}{5.8} & \ms{13.5}{6.2} & \ms{14.9}{6.9} \\
        & & & ATC  & \ms{17.1}{4.4} & \ms{17.2}{6.1} & \ms{17.3}{4.5} & \ms{17.1}{4.4} & \msb{8.5}{3.1} & \ms{11.6}{4.8} & \msbb{8.7}{3.2} & \msb{8.5}{3.1} \\
        & & & TS-ATC  & \ms{16.4}{6.0} & \ms{19.4}{6.8} & \ms{16.3}{5.4} & \ms{16.3}{5.4} & \ms{9.8}{3.8} & \ms{13.2}{5.8} & \ms{10.7}{4.4} & \ms{10.0}{3.7} \\
\bottomrule
    \end{tabular}%
}
\caption{Evaluation of F1-Score and AUC estimation for different classification tasks under various types of domain shifts based on MAE. The best and second-best results within the same category are in \textbf{bold} and \underline{underlined} for each configuration.}
\label{tab:cls_f1score_auc}
\end{table*}

\begin{table*}[ht]
\centering
 \resizebox{\linewidth}{!}{%
    \begin{tabular}{ccccccccccccc}
    \toprule
        \multicolumn{3}{c}{\multirow{3}{*}{\shortstack{Dataset \\ and Training \\ Strategy}}} & &  \multicolumn{9}{c}{Confidence Score}  \\
        & & & \multirow{2}{*}{\shortstack{Estimation \\ Algorithm}} & \multicolumn{4}{c}{Class-Agnostic} & \multicolumn{4}{c}{Class-Specific} & \multirow{2}{*}{\shortstack{MOVAL \\ Ensemble}} \\
        & & & & MCP & Energy & Entropy & Doctor & MCP & Energy & Entropy & Doctor \\
        \cmidrule(r){1-3} \cmidrule(r){4-4} \cmidrule(lr){5-8} \cmidrule(lr){9-12} \cmidrule(lr){13-13}
        
        \multirow{5}{*}{\rotatebox{90}{\shortstack{CIFAR10}}} & \multirow{5}{*}{\rotatebox{90}{\shortstack{Synthetic-95}}} & \multirow{5}{*}{\rotatebox{90}{\shortstack{Vanilla \\ Training}}} & Vanilla  & \ms{57.0}{10.4} & \ms{8.6}{5.3} & \ms{54.5}{10.4} & \ms{61.1}{10.5} & --- & --- & --- & --- & \multirow{5}{*}{---} \\
        & & & TS  & \ms{28.8}{9.6} & \ms{39.4}{9.5} & \ms{27.2}{9.9} & \ms{24.2}{11.1} & \ms{27.4}{4.4} & \ms{35.9}{6.1} & \ms{24.3}{3.7} & \ms{18.9}{3.4} \\
        & & & DoC  & \ms{36.1}{10.4} & \ms{32.4}{8.5} & \ms{35.0}{10.4} & \ms{37.4}{10.5} & \ms{34.1}{5.5} & \ms{28.9}{4.5} & \ms{32.9}{5.3} & \ms{35.2}{5.6}  \\
        & & & ATC  & \ms{23.1}{11.3} & \ms{20.2}{10.1} & \ms{22.9}{11.4} & \ms{23.2}{11.3} & \ms{15.7}{4.1} & \msbb{11.2}{5.2} & \ms{14.7}{4.2} & \ms{15.2}{4.1} \\
        & & & TS-ATC  & \ms{21.5}{9.9} & \ms{18.1}{9.1} & \ms{21.4}{11.3} & \ms{21.4}{11.4} & \ms{14.3}{3.4} & \msb{10.5}{6.4} & \ms{14.0}{3.4} & \ms{13.7}{3.5} \\
        
        \bottomrule
    \end{tabular}%
}
\caption{Evaluation of accuracy estimation for different classification tasks under various types of domain shifts based on MAE. The test datasets include only samples from the 2 most minority classes. The best and second-best results within the same category are in \textbf{bold} and \underline{underlined} for each configuration.}
\label{tab:cls_acc_imbalanced}
\end{table*}

\begin{table*}[ht]
\centering
 \resizebox{\linewidth}{!}{%
    \begin{tabular}{ccccccccccccc}
    \toprule
        \multicolumn{3}{c}{\multirow{3}{*}{\shortstack{Evaluation \\  Metric and \\ Dataset}}} & &  \multicolumn{9}{c}{Confidence Score}  \\
        & & & \multirow{2}{*}{\shortstack{Estimation \\ Algorithm}} & \multicolumn{4}{c}{Class-Agnostic} & \multicolumn{4}{c}{Class-Specific} & \multirow{2}{*}{\shortstack{MOVAL \\ Ensemble}} \\
        & & & & MCP & Energy & Entropy & Doctor & MCP & Energy & Entropy & Doctor \\
        \cmidrule(r){1-3} \cmidrule(r){4-4} \cmidrule(lr){5-8} \cmidrule(lr){9-12} \cmidrule(lr){13-13}
        
        \multirow{30}{*}{\rotatebox{90}{\shortstack{Sensitivity}}} & \multirow{10}{*}{\rotatebox{90}{\shortstack{Cardiac-2D}}} & \multirow{5}{*}{\rotatebox{90}{\shortstack{Synthetic-65}}} & Vanilla  & \ms{15.1}{6.5} & \ms{62.3}{9.2} & \ms{12.4}{5.9} & \ms{14.5}{6.3} & --- & --- & --- & --- & \multirow{5}{*}{\ms{3.3}{2.5}} \\
        & & & TS  & \ms{9.8}{5.8} & \ms{18.8}{7.8} & \ms{7.4}{5.3} & \ms{8.1}{5.3} & \ms{4.4}{3.7} & \ms{6.0}{6.3} & \ms{4.2}{3.2} & \ms{4.1}{3.1} \\
        & & & DoC  & \ms{14.4}{6.5} & \ms{13.5}{7.0} & \ms{11.8}{5.9} & \ms{13.8}{6.3} & \ms{5.1}{5.1} & \ms{5.5}{5.1} & \ms{4.6}{4.5} & \ms{4.9}{4.9} \\
        & & & ATC  & \msbb{3.2}{2.4} & \ms{18.8}{7.8} & \ms{3.3}{2.5} & \msbb{3.2}{2.4} & \msb{3.1}{2.5} & \ms{4.5}{5.9} & \msb{3.1}{2.5} & \msb{3.1}{2.5} \\
        & & & TS-ATC  &\ms{3.5}{5.3} & \ms{18.8}{7.8} & \ms{3.5}{2.7} & \ms{3.5}{2.8} & \ms{3.4}{3.4} & \ms{6.4}{9.1} & \ms{3.4}{3.4} & \ms{3.4}{3.5} \\
\bottomrule
    \end{tabular}%
}
\caption{Evaluation of sensitivity estimation for different segmentation tasks under various types of domain shifts based on MAE. The best and second-best results within the same category are in \textbf{bold} and \underline{underlined} for each configuration.}
\label{tab:seg_sensitivity}
\end{table*}

\begin{table*}[ht]
\centering
 \resizebox{\linewidth}{!}{%
    \begin{tabular}{ccccccccccccc}
    \toprule
        \multicolumn{3}{c}{\multirow{3}{*}{\shortstack{Evaluation \\  Metric and \\ Dataset}}} & &  \multicolumn{9}{c}{Confidence Score}  \\
        & & & \multirow{2}{*}{\shortstack{Estimation \\ Algorithm}} & \multicolumn{4}{c}{Class-Agnostic} & \multicolumn{4}{c}{Class-Specific} & \multirow{2}{*}{\shortstack{MOVAL \\ Ensemble}} \\
        & & & & MCP & Energy & Entropy & Doctor & MCP & Energy & Entropy & Doctor \\
        \cmidrule(r){1-3} \cmidrule(r){4-4} \cmidrule(lr){5-8} \cmidrule(lr){9-12} \cmidrule(lr){13-13}

        \multirow{30}{*}{\rotatebox{90}{\shortstack{Precision}}} & \multirow{10}{*}{\rotatebox{90}{\shortstack{Cardiac-2D}}} & \multirow{5}{*}{\rotatebox{90}{\shortstack{Synthetic-65}}} & Vanilla  & \ms{9.5}{1.5} & \ms{55.0}{3.7} & \msb{8.2}{0.9} & \ms{9.3}{1.3} & --- & --- & --- & --- & \multirow{5}{*}{\msbb{8.4}{0.8}} \\
        & & & TS  & \ms{10.8}{1.5} & \ms{17.7}{7.7} & \ms{11.0}{2.0} & \ms{11.0}{1.9} & \ms{10.1}{0.6} & \ms{52.1}{3.6} & \ms{10.1}{1.9} & \ms{9.3}{1.9} \\
        & & & DoC  & \ms{10.8}{1.6} & \ms{10.3}{1.4} & \ms{11.0}{1.3} & \ms{10.9}{1.5} & \ms{9.2}{0.6} & \ms{52.1}{3.6} & \ms{9.2}{1.0} & \ms{9.0}{0.7} \\
        & & & ATC  & --- & --- & --- & --- & --- & --- & --- & ---  \\
        & & & TS-ATC  & --- & --- & --- & --- & --- & --- & --- & ---  \\
\bottomrule
    \end{tabular}%
}
\caption{Evaluation of precision estimation for different segmentation tasks under various types of domain shifts based on MAE. The best and second-best results within the same category are in \textbf{bold} and \underline{underlined} for each configuration.}
\label{tab:seg_precision}
\end{table*}

\begin{table*}[ht]
\centering
 \resizebox{\linewidth}{!}{%
    \begin{tabular}{ccccccccccccc}
    \toprule
        \multicolumn{3}{c}{\multirow{3}{*}{\shortstack{Evaluation \\  Metric and \\ Dataset}}} & &  \multicolumn{9}{c}{Confidence Score}  \\
        & & & \multirow{2}{*}{\shortstack{Estimation \\ Algorithm}} & \multicolumn{4}{c}{Class-Agnostic} & \multicolumn{4}{c}{Class-Specific} & \multirow{2}{*}{\shortstack{MOVAL \\ Ensemble}} \\
        & & & & MCP & Energy & Entropy & Doctor & MCP & Energy & Entropy & Doctor \\
        \cmidrule(r){1-3} \cmidrule(r){4-4} \cmidrule(lr){5-8} \cmidrule(lr){9-12} \cmidrule(lr){13-13}
        
        \multirow{30}{*}{\rotatebox{90}{\shortstack{Accuracy}}} & \multirow{10}{*}{\rotatebox{90}{\shortstack{Cardiac-2D}}} & \multirow{5}{*}{\rotatebox{90}{\shortstack{Synthetic-65}}} & Vanilla  & \ms{0.8}{0.3} & \ms{80.9}{0.2} & \ms{0.7}{0.3} & \ms{0.8}{0.3} & --- & --- & --- & --- & \multirow{5}{*}{\ms{0.3}{0.5}} \\
        & & & TS  & \msbb{0.2}{0.4} & \ms{1.2}{0.5} & \ms{0.3}{0.6} & \ms{0.3}{0.6} & \msbb{0.2}{0.4} & \ms{0.7}{0.5} & \ms{0.3}{0.6} & \ms{0.3}{0.6} \\
        & & & DoC  & \msbb{0.2}{0.3} & \msb{0.1}{0.2} & \ms{0.3}{0.5} & \ms{0.3}{0.5} & \msbb{0.2}{0.3} & \msbb{0.2}{0.2} & \ms{0.3}{0.5} & \ms{0.3}{0.5} \\
        & & & ATC  & \ms{0.4}{0.7} & \ms{1.0}{0.3} & \ms{0.4}{0.8} & \ms{0.4}{0.7} & \ms{0.4}{0.8} & \ms{0.6}{0.6} & \ms{0.4}{0.8} & \ms{0.4}{0.8} \\
        & & & TS-ATC  & \ms{0.4}{0.6} & \ms{1.2}{0.5} & \ms{0.4}{0.8} & \ms{0.4}{0.8} & \ms{0.5}{0.8} & \ms{0.9}{0.4} & \ms{0.5}{0.8} & \ms{0.5}{0.8} \\
\bottomrule
    \end{tabular}%
}
\caption{Evaluation of accuracy estimation for different segmentation tasks under various types of domain shifts based on MAE. The best and second-best results within the same category are in \textbf{bold} and \underline{underlined} for each configuration.}
\label{tab:seg_acc}
\end{table*}

\begin{table*}[ht]
\centering
 \resizebox{\linewidth}{!}{%
    \begin{tabular}{ccccccccccccc}
    \toprule
        \multicolumn{3}{c}{\multirow{3}{*}{\shortstack{Evaluation \\  Metric and \\ Dataset}}} & &  \multicolumn{9}{c}{Confidence Score}  \\
        & & & \multirow{2}{*}{\shortstack{Estimation \\ Algorithm}} & \multicolumn{4}{c}{Class-Agnostic} & \multicolumn{4}{c}{Class-Specific} & \multirow{2}{*}{\shortstack{MOVAL \\ Ensemble}} \\
        & & & & MCP & Energy & Entropy & Doctor & MCP & Energy & Entropy & Doctor \\
        \cmidrule(r){1-3} \cmidrule(r){4-4} \cmidrule(lr){5-8} \cmidrule(lr){9-12} \cmidrule(lr){13-13}

        \multirow{30}{*}{\rotatebox{90}{\shortstack{AUC}}} & \multirow{10}{*}{\rotatebox{90}{\shortstack{Cardiac-2D}}} & \multirow{5}{*}{\rotatebox{90}{\shortstack{Synthetic-65}}} & Vanilla  & \ms{2.9}{0.6} & \ms{45.6}{0.8} & \ms{3.0}{0.7} & \ms{2.7}{0.6} & --- & --- & --- & --- & \multirow{5}{*}{\msb{2.3}{0.4}} \\
        & & & TS  &\ms{4.3}{1.0} & \ms{3.1}{0.8} & \ms{4.3}{1.1} & \ms{4.1}{1.0} & \ms{2.5}{0.5} & \ms{33.1}{0.6} & \msbb{2.4}{0.4} & \msb{2.3}{0.4} \\
        & & & DoC  & \ms{3.4}{0.6} & \ms{5.0}{1.3} & \ms{3.5}{0.7} & \ms{3.2}{0.6} & \ms{2.5}{0.5} & \ms{19.2}{0.3} & \msbb{2.4}{0.4} & \msb{2.3}{0.4} \\
        & & & ATC  & --- & --- & --- & --- & --- & --- & --- & --- \\
        & & & TS-ATC  & --- & --- & --- & --- & --- & --- & --- & --- \\
\bottomrule
    \end{tabular}%
}
\caption{Evaluation of AUC estimation for different segmentation tasks under various types of domain shifts based on MAE. The best and second-best results within the same category are in \textbf{bold} and \underline{underlined} for each configuration.}
\label{tab:seg_auc}
\end{table*}

\begin{table*}[ht]
\centering
 \resizebox{\linewidth}{!}{%
    \begin{tabular}{ccccccccccccc}
    \toprule
        \multicolumn{3}{c}{\multirow{3}{*}{\shortstack{Dataset \\ and Training \\ Strategy}}} & &  \multicolumn{9}{c}{Confidence Score}  \\
        & & & \multirow{2}{*}{\shortstack{Estimation \\ Algorithm}} & \multicolumn{4}{c}{Class-Agnostic} & \multicolumn{4}{c}{Class-Specific} & \multirow{2}{*}{\shortstack{MOVAL \\ Ensemble}} \\
        & & & & MCP & Energy & Entropy & Doctor & MCP & Energy & Entropy & Doctor \\
        \cmidrule(r){1-3} \cmidrule(r){4-4} \cmidrule(lr){5-8} \cmidrule(lr){9-12} \cmidrule(lr){13-13}
        
        \multirow{20}{*}{\rotatebox{90}{\shortstack{CIFAR10}}} & \multirow{20}{*}{\rotatebox{90}{\shortstack{Synthetic-95}}} & \multirow{5}{*}{\rotatebox{90}{\shortstack{Balanced \\ Training \uppercase\expandafter{\romannumeral1}}}} & Vanilla  & \ms{25.2}{8.2} & \ms{21.8}{8.5} & \ms{49.8}{10.3} & \ms{13.0}{2.7} & --- & --- & --- & --- & \multirow{5}{*}{\ms{11.5}{6.4}} \\
        & & & TS  & \ms{27.9}{8.0} & \ms{32.9}{11.0} & \ms{27.3}{7.8} & \ms{27.7}{8.2} & \ms{15.7}{7.7} & \ms{23.5}{11.9} & \ms{14.3}{6.9} & \ms{14.4}{7.4} \\
        & & & DoC  & \ms{31.5}{10.2} & \ms{32.8}{11.0} & \ms{31.9}{10.3} & \ms{29.6}{9.1} & \ms{19.9}{9.9} & \ms{23.5}{11.9} & \ms{20.4}{10.2} & \ms{16.0}{7.8} \\
        & & & ATC  &  \ms{22.0}{5.6} & \ms{36.5}{10.3} & \ms{28.3}{8.8} & \ms{26.7}{8.1} & \msbb{9.4}{6.2} & \ms{29.0}{14.3} & \ms{13.3}{7.1} & \ms{12.0}{6.6} \\
        & & & TS-ATC  & \ms{22.3}{7.8} & \ms{36.5}{10.3} & \ms{22.2}{5.1} & \ms{25.5}{7.4} & \msb{8.8}{5.0} & \ms{29.1}{14.3} & \msb{8.8}{5.2} & \ms{11.6}{6.7} \\
\bottomrule
    \end{tabular}%
}
\caption{Evaluation of F1-Score estimation for CIFAR10 classification under different training strategies based on MAE. The best and second-best results within the same category are in \textbf{bold} and \underline{underlined} for each configuration.}
\label{tab:cls_f1_bl}
\end{table*}

\begin{table*}[ht]
\centering
 \resizebox{\linewidth}{!}{%
    \begin{tabular}{ccccccccccccc}
    \toprule
        \multicolumn{3}{c}{\multirow{3}{*}{\shortstack{Dataset \\ and Training \\ Strategy}}} & &  \multicolumn{9}{c}{Confidence Score}  \\
        & & & \multirow{2}{*}{\shortstack{Estimation \\ Algorithm}} & \multicolumn{4}{c}{Class-Agnostic} & \multicolumn{4}{c}{Class-Specific} & \multirow{2}{*}{\shortstack{MOVAL \\ Ensemble}} \\
        & & & & MCP & Energy & Entropy & Doctor & MCP & Energy & Entropy & Doctor \\
        \cmidrule(r){1-3} \cmidrule(r){4-4} \cmidrule(lr){5-8} \cmidrule(lr){9-12} \cmidrule(lr){13-13}
        
        \multirow{20}{*}{\rotatebox{90}{\shortstack{CIFAR10}}} & \multirow{20}{*}{\rotatebox{90}{\shortstack{Synthetic-95}}} & \multirow{5}{*}{\rotatebox{90}{\shortstack{Balanced \\ Training \uppercase\expandafter{\romannumeral1}}}} & Vanilla  & \ms{37.3}{9.3} & \ms{8.3}{7.2} & \ms{52.4}{9.4} & \ms{20.2}{8.9} & --- & --- & --- & --- & \multirow{5}{*}{\ms{12.4}{11.7}} \\
        & & & TS  & \ms{12.4}{7.7} & \ms{22.5}{10.9} & \ms{11.2}{7.7} & \ms{10.5}{9.4} & \ms{13.6}{10.3} & \ms{20.2}{13.6} & \ms{12.7}{10.3} & \ms{11.1}{10.7}\\
        & & & DoC  & \ms{18.4}{9.3} & \ms{22.4}{10.9} & \ms{19.0}{9.4} & \ms{14.0}{9.3} & \ms{18.2}{11.9} & \ms{20.2}{13.6} & \ms{18.8}{12.2} & \ms{14.3}{11.0} \\
        & & & ATC  &  \msb{4.5}{6.0} & \ms{26.4}{12.9} & \ms{8.1}{9.8} & \ms{8.0}{8.7} & \ms{6.4}{8.8} & \ms{23.1}{17.1} & \ms{8.3}{9.8} & \ms{8.2}{9.5} \\
        & & & TS-ATC  & \ms{5.8}{7.7} & \ms{26.5}{12.9} & \msbb{5.3}{5.4} & \ms{7.6}{8.0} & \ms{7.5}{8.9} & \ms{23.1}{17.1} & \ms{7.3}{9.1} & \ms{8.0}{9.1} \\
\bottomrule
    \end{tabular}%
}
\caption{Evaluation of accuracy estimation for CIFAR10 classification under different training strategies based on MAE. The proposed CS variants underperform in accuracy estimation compared to their counterparts when models are trained using advanced balanced training strategies. We observe that balanced training strategies largely affect the model's predicted probabilities, resulting in a more significant difference in the confusion matrix when exposed to domain shifts. These changes, in particular, create difficulties for CS variants, as they rely on predictions to choose class parameters. In such cases, we recommend performing class-wise calibration with group-wise parameters (c.f. Table~\ref{tab:cls_acc_bl_batch}) or alternatively tracking the F1-Score. The best and second-best results within the same category are in \textbf{bold} and \underline{underlined} for each configuration.}
\label{tab:cls_acc_bl}
\end{table*}

\begin{table*}[ht]
\centering
 \resizebox{\linewidth}{!}{%
    \begin{tabular}{ccccccccccccc}
    \toprule
        \multicolumn{3}{c}{\multirow{3}{*}{\shortstack{Dataset \\ and Training \\ Strategy}}} & &  \multicolumn{9}{c}{Confidence Score}  \\
        & & & \multirow{2}{*}{\shortstack{Estimation \\ Algorithm}} & \multicolumn{4}{c}{Class-Agnostic} & \multicolumn{4}{c}{Class-Specific} & \multirow{2}{*}{\shortstack{MOVAL \\ Ensemble}} \\
        & & & & MCP & Energy & Entropy & Doctor & MCP & Energy & Entropy & Doctor \\
        \cmidrule(r){1-3} \cmidrule(r){4-4} \cmidrule(lr){5-8} \cmidrule(lr){9-12} \cmidrule(lr){13-13}
        
        \multirow{20}{*}{\rotatebox{90}{\shortstack{CIFAR10}}} & \multirow{20}{*}{\rotatebox{90}{\shortstack{Synthetic-95}}} & \multirow{5}{*}{\rotatebox{90}{\shortstack{Balanced \\ Training \uppercase\expandafter{\romannumeral1}}}} & Vanilla  & \ms{37.3}{9.3} & \ms{8.3}{7.2} & \ms{52.4}{9.4} & \ms{20.2}{8.9} & --- & --- & --- & --- & \multirow{5}{*}{\ms{11.5}{6.9}} \\
        & & & TS  & \ms{12.4}{7.7} & \ms{22.5}{10.9} & \ms{11.2}{7.7} & \ms{10.5}{9.4} & \ms{12.2}{7.4} & \ms{20.7}{9.8} & \ms{11.1}{7.5} & \ms{10.6}{9.3} \\
        & & & DoC  & \ms{18.4}{9.3} & \ms{22.4}{10.9} & \ms{19.0}{9.4} & \ms{14.0}{9.3} & \ms{17.6}{8.7} & \ms{20.7}{9.8} & \ms{18.2}{8.9} & \ms{13.8}{9.1} \\
        & & & ATC  &  \msb{4.5}{6.0} & \ms{26.4}{12.9} & \ms{8.1}{9.8} & \ms{8.0}{8.7} & \msbb{4.7}{6.0} & \ms{24.2}{11.2} & \ms{8.3}{10.2} & \ms{8.1}{9.1} \\
        & & & TS-ATC  & \ms{5.8}{7.7} & \ms{26.5}{12.9} & \ms{5.3}{5.4} & \ms{7.6}{8.0} & \ms{5.4}{5.2} & \ms{24.2}{11.2} & \ms{5.3}{5.4} & \ms{7.8}{8.4} \\

        \cmidrule(r){3-13}

        & & \multirow{5}{*}{\rotatebox{90}{\shortstack{Balanced \\ Training \uppercase\expandafter{\romannumeral2}}}} & Vanilla  & \ms{30.5}{7.9} & \ms{31.9}{6.6} & \ms{28.3}{7.6} & \ms{34.6}{8.2} & --- & --- & --- & --- & \multirow{5}{*}{\msbb{3.7}{3.0}} \\
        & & & TS  & \ms{8.1}{6.2} & \ms{15.0}{8.4} & \ms{7.4}{5.6} & \ms{5.9}{4.6} & \ms{8.0}{5.8} & \ms{14.5}{8.0} & \ms{7.1}{5.4} & \ms{5.5}{4.6} \\
        & & & DoC  & \ms{13.0}{7.9} & \ms{10.3}{6.8} & \ms{12.3}{7.6} & \ms{14.1}{8.2} & \ms{12.8}{7.5} & \ms{10.0}{6.4} & \ms{12.0}{7.2} & \ms{13.7}{7.8} \\
        & & & ATC  & \ms{5.9}{5.4} & \ms{4.3}{2.6} & \ms{5.7}{5.2} & \ms{5.9}{5.4} & \ms{5.7}{5.5} & \ms{3.9}{2.3} & \ms{5.1}{5.1} & \ms{5.6}{5.4} \\
        & & & TS-ATC  & \ms{4.4}{5.6} & \ms{4.8}{3.3} & \ms{3.9}{2.9} & \ms{3.9}{3.0} & \ms{4.2}{3.9} & \ms{4.6}{3.0} & \msb{3.6}{3.0} & \msb{3.6}{3.0} \\

        \cmidrule(r){3-13}

        & & \multirow{5}{*}{\rotatebox{90}{\shortstack{Robust \\ Training \uppercase\expandafter{\romannumeral1}}}} & Vanilla  & \ms{30.8}{7.1} & \ms{30.4}{7.0} & \ms{28.3}{6.8} & \ms{38.8}{8.3} & --- & --- & --- & --- & \multirow{5}{*}{\ms{3.7}{3.0}} \\
        & & & TS  & \ms{9.2}{5.4} & \ms{18.6}{9.2} & \ms{8.4}{5.3} & \ms{6.9}{5.1} & \ms{8.6}{4.8} & \ms{17.7}{8.7} & \ms{7.7}{4.7} & \ms{6.3}{4.5} \\
        & & & DoC  & \ms{14.4}{7.1} & \ms{13.0}{7.1} & \ms{13.5}{6.8} & \ms{17.3}{8.3} & \ms{13.8}{6.6} & \ms{12.4}{6.7} & \ms{12.9}{6.2} & \ms{16.5}{7.7} \\
        & & & ATC  & \ms{5.4}{4.4} & \ms{4.8}{3.4} & \ms{4.6}{4.4} & \ms{5.2}{4.4} & \ms{4.0}{3.5} & \ms{4.3}{3.3} & \msbb{3.5}{3.4} & \ms{3.9}{3.5} \\
        & & & TS-ATC  & \ms{3.9}{5.3} & \ms{4.3}{4.2} & \ms{4.4}{2.6} & \ms{4.8}{2.9} & \msb{3.3}{2.8} & \ms{4.4}{4.1} & \ms{3.9}{2.4} & \ms{4.0}{2.5} \\

        \cmidrule(r){3-13}

        & & \multirow{5}{*}{\rotatebox{90}{\shortstack{Robust \\ Training \uppercase\expandafter{\romannumeral2}}}} & Vanilla  & \ms{24.9}{4.8} & \ms{36.8}{4.5} & \ms{22.2}{4.7} & \ms{29.9}{5.1} & --- & --- & --- & ---& \multirow{5}{*}{\ms{6.4}{5.1}} \\
        & & & TS  & \ms{5.6}{4.3} & \ms{12.8}{5.9} & \ms{4.9}{4.3} & \ms{5.1}{4.0} & \ms{5.1}{3.8} & \ms{12.0}{5.4} & \ms{4.6}{3.9} & \ms{5.0}{3.6} \\
        & & & DoC  & \ms{9.6}{4.8} & \ms{8.0}{4.6} & \ms{8.5}{4.7} & \ms{10.7}{5.1} & \ms{8.8}{4.4} & \ms{7.6}{4.3} & \ms{7.8}{4.3} & \ms{9.9}{4.7} \\
        & & & ATC  & \ms{4.6}{3.7} & \ms{8.8}{6.8} & \ms{4.2}{3.5} & \ms{4.5}{3.6} & \msb{3.7}{3.0} & \ms{8.6}{6.7} & \ms{4.0}{3.1} & \msbb{3.8}{3.0} \\
        & & & TS-ATC  & \ms{5.1}{4.3} & \ms{10.8}{8.2} & \ms{6.3}{4.7} & \ms{6.4}{4.8} & \ms{5.0}{3.7} & \ms{10.7}{8.2} & \ms{6.0}{4.8} & \ms{6.1}{4.9} \\
        
        \bottomrule
    \end{tabular}%
}
\caption{Related to Extended Table~\ref{tab:cls_acc_bl}. Evaluation of accuracy estimation for CIFAR10 classification under different training strategies based on MAE. To achieve stable estimation in complex scenarios, we employ a group-wise optimization strategy. Specifically, we estimate two parameters for the CS variants: one shared by the 8 classes with the highest number of predicted samples, and another shared by the remaining 2 classes. Compared to CS variants applied to individual classes, this technique reduces the number of required validation samples and mitigates the complexity of the confusion matrix (e.g., misclassifications within a group have less impact on the estimation results). In this ways, the CS variants can effectively generalize with advanced balanced training strategies. The best and second-best results within the same category are in \textbf{bold} and \underline{underlined} for each configuration.}
\label{tab:cls_acc_bl_batch}
\end{table*}

\begin{table*}[ht]
\centering
 \resizebox{\linewidth}{!}{%
\begin{tabular}{ccccccccccccc}
    \toprule
        \multicolumn{3}{c}{\multirow{3}{*}{\shortstack{Dataset \\ and Training \\ Strategy}}} & &  \multicolumn{9}{c}{Confidence Score}  \\
        & & & \multirow{2}{*}{\shortstack{Estimation \\ Algorithm}} & \multicolumn{4}{c}{Class-Agnostic} & \multicolumn{4}{c}{Class-Specific} & \multirow{2}{*}{\shortstack{MOVAL \\ Ensemble}} \\
        & & & & MCP & Energy & Entropy & Doctor & MCP & Energy & Entropy & Doctor \\
        \cmidrule(r){1-3} \cmidrule(r){4-4} \cmidrule(lr){5-8} \cmidrule(lr){9-12} \cmidrule(lr){13-13}
        
        \multirow{40}{*}{\rotatebox{90}{\shortstack{Brain Lesion-3D}}} & \multirow{20}{*}{\rotatebox{90}{\shortstack{Synthetic-42}}} & \multirow{5}{*}{\rotatebox{90}{\shortstack{Balanced \\ Training \uppercase\expandafter{\romannumeral1}}}} & Vanilla  & \ms{24.8}{9.7} & \ms{70.2}{10.0} & \ms{21.6}{7.5} & \ms{23.3}{8.5} & --- & --- & --- & --- & \multirow{5}{*}{\msbb{2.6}{4.2}} \\
        & & & TS  & \ms{22.5}{8.9} & \ms{25.9}{10.5} & \ms{19.6}{6.8} & \ms{20.2}{7.1} & \ms{5.8}{9.0} & \ms{6.6}{10.4} & \ms{2.8}{4.7} & \ms{2.9}{4.8} \\
        & & & DoC  & \ms{24.8}{9.7} & \ms{25.9}{10.5} & \ms{21.5}{7.5} & \ms{23.2}{8.5} & \ms{5.8}{9.2} & \ms{6.5}{10.2} & \ms{4.0}{6.8} & \ms{4.8}{7.9} \\
        & & & ATC  & \ms{17.1}{5.2} & \ms{25.9}{10.5} & \ms{17.1}{5.2} & \ms{17.1}{5.2} & \ms{2.7}{4.0} & \ms{2.7}{4.0} & \ms{2.7}{4.0} & \ms{2.7}{4.0} \\
        & & & TS-ATC  &\ms{17.1}{6.8} & \ms{25.9}{10.5} & \ms{17.1}{5.2} & \ms{17.1}{5.2} & \ms{2.7}{4.0} & \msb{2.5}{3.4} & \ms{2.7}{4.0} & \ms{2.7}{4.0} \\
\bottomrule
    \end{tabular}%
}
\caption{Evaluation of F1-Score estimation for brain lesion segmentation under different training strategies based on MAE. The best and second-best results within the same category are in \textbf{bold} and \underline{underlined} for each configuration.}
\label{tab:seg_training}
\end{table*}

\begin{table*}[ht]
\centering
\resizebox{\linewidth}{!}{%
    \begin{tabular}{cccccccc}
    \toprule
        \multirow{2}{*}{\shortstack{Estimating\\Metric}} & \multirow{2}{*}{\shortstack{Group-Wise\\Calibration}} & \multicolumn{6}{c}{Samples Used for Estimating}\\
        & &  3000 (100$\%$) & 1500 (50$\%$) & 600 (20$\%$) & 300 (10$\%$) & 150 (5$\%$) & 90 (3$\%$) \\
        \cmidrule(r){1-1} \cmidrule(r){2-2} \cmidrule(r){3-8} 

         Accuracy & No & \ms{2.8}{2.1} & \ms{2.8}{2.3} & \ms{2.9}{2.3} & \ms{3.4}{2.1} & \ms{3.9}{3.0} & \ms{4.6}{3.6} \\

         Sensitivity & No & \ms{9.5}{3.6} & \ms{9.7}{3.5} & \ms{11.0}{3.6} & \ms{12.2}{4.0} & \ms{16.4}{4.4} & \ms{25.2}{8.1} \\

         Precision & No & \ms{31.6}{7.0} & \ms{32.0}{7.2} & \ms{31.7}{6.8} & \ms{32.0}{7.0} & \ms{31.9}{6.8} & \ms{32.8}{7.2} \\

         F1-Score & No & \ms{8.5}{3.1} & \ms{8.7}{3.0} & \ms{10.5}{3.8} & \ms{11.4}{4.2} & \ms{15.4}{4.4} & \ms{20.4}{6.2}  \\

         AUC & No & \ms{3.0}{1.1} & \ms{3.0}{1.1} & \ms{3.3}{1.4} & \ms{3.6}{1.6} & \ms{3.8}{1.8} & \ms{4.0}{2.0} \\

         Accuracy & 5 Groups & \ms{2.9}{2.1} & \ms{2.9}{2.2} & \ms{3.0}{2.4} & \ms{3.2}{2.4} & \ms{3.1}{2.2} & \ms{3.9}{3.0} \\

         Sensitivity & 5 Groups & \ms{10.8}{4.3} & \ms{10.8}{4.4} & \ms{11.5}{4.1} & \ms{12.4}{4.4} & \ms{14.5}{4.4} & \ms{15.6}{4.6} \\

         Precision & 5 Groups & \ms{32.7}{7.1} & \ms{32.7}{7.1} & \ms{32.8}{7.2} & \ms{32.8}{7.2} & \ms{32.9}{7.2} & \ms{32.9}{7.3} \\

         F1-Score & 5 Groups & \ms{8.4}{3.4} & \ms{8.6}{3.4} & \ms{9.3}{3.6} & \ms{9.7}{3.7} & \ms{13.4}{4.4} & \ms{14.2}{4.7}  \\

         AUC & 5 Groups & \ms{3.2}{1.2} & \ms{3.2}{1.3} & \ms{3.4}{1.5} & \ms{3.7}{1.8} & \ms{3.7}{1.8} & \ms{4.0}{2.0} \\
        
        \bottomrule
    \end{tabular}%
    }
\caption{MAE of estimation results for CIFAR10 classification when MOVAL Ensemble is optimized with different amounts of validation data. For group-wise calibration, we split the 10 cases into 5 groups, with each group containing 2 classes.}
\label{tab:moval_cls_ablation}
\end{table*}

\begin{table*}[ht]
\centering
    \begin{tabular}{ccccccc}
    \toprule
        \multirow{2}{*}{Dataset} & \multicolumn{6}{c}{Samples Used for Estimating} \\
        & 19 (100$\%$) & 12 (63$\%$) & 6 (32$\%$) & 3 (16$\%$) & 2 (11$\%$) & 1 (5$\%$) \\
        \cmidrule(r){1-1} \cmidrule(r){2-7} 
        
        Synthetic-65 & \ms{4.3}{5.0} & \ms{4.6}{4.4} & \ms{4.7}{4.6} & \ms{5.4}{4.2} & \ms{6.2}{3.3} & \ms{8.3}{2.4} \\

        Natural-4 & \ms{2.6}{0.4} & \ms{2.6}{0.3} & \ms{3.3}{0.3} & \ms{3.6}{0.3} & \ms{4.2}{0.2} & \ms{6.5}{0.4} \\

        \midrule
        
        Average & 3.5 & 3.6 & 4.0 & 4.5 & 5.2 & 7.4 \\
        
        \bottomrule
    \end{tabular}%
\caption{MAE of F1-Score estimation results for cardiac segmentation when MOVAL Ensemble is optimized with different amounts of validation data.}
\label{tab:moval_seg2d_ablation}
\end{table*}

\begin{table*}[ht]
\centering
\resizebox{\linewidth}{!}{%
    \begin{tabular}{ccccccc}
    \toprule
        \multirow{2}{*}{Dataset} & \multirow{2}{*}{\shortstack{Training\\Strategy}} & \multicolumn{5}{c}{Samples Used for Estimating}\\
        & & 9 (100$\%$) & 6 (66$\%$) & 3 (33$\%$) & 2 (22$\%$) & 1 (11$\%$) \\
        \cmidrule(r){1-1} \cmidrule(r){2-2} \cmidrule(r){3-7} 
        
        \multirow{5}{*}{Synthetic-42} & Vanilla & \ms{0.5}{0.6} & \ms{2.9}{0.3} & \ms{4.0}{0.4} & \ms{5.8}{0.1} & \ms{13.5}{0.3} \\
        & Balanced Training \uppercase\expandafter{\romannumeral1} & \ms{0.5}{0.5} & \ms{3.4}{0.1} & \ms{3.8}{0.1} & \ms{6.5}{0.2} & \ms{14.3}{0.4}\\
        & Balanced Training \uppercase\expandafter{\romannumeral2} & \ms{0.2}{0.3} & \ms{1.5}{0.1} & \ms{3.5}{0.4} & \ms{5.4}{0.2} & \ms{8.4}{0.2} \\
        & Robust Training \uppercase\expandafter{\romannumeral1} & \ms{0.9}{0.7} & \ms{4.4}{0.3} & \ms{4.3}{0.3} & \ms{8.8}{0.4} & \ms{21.0}{0.8} \\
        & Robust Training \uppercase\expandafter{\romannumeral2} & \ms{0.6}{0.6} & \ms{3.8}{0.4} & \ms{3.6}{0.2} & \ms{6.8}{0.1} & \ms{15.0}{0.2} \\

        \multirow{5}{*}{Natural-12} & Vanilla & \ms{25.5}{12.3} & \ms{25.1}{12.5} & \ms{29.2}{12.0} & \ms{29.3}{12.4} & \ms{26.3}{12.1} \\
\bottomrule
    \end{tabular}%
    }
\caption{MAE of F1-Score estimation results for brain lesion segmentation under different training strategies when MOVAL Ensemble is optimized with different amounts of validation data.}
\label{tab:moval_seg3d_ablation}
\end{table*}

\begin{table*}[ht]
\centering
    \begin{tabular}{ccccccc}
    \toprule
        \multirow{2}{*}{\shortstack{Estimating \\ Metric}} & \multicolumn{3}{c}{Classification}\\
        & Choice 1 & Choice 2 & Choice 3 \\
        \cmidrule(r){1-1} \cmidrule(r){2-4} 
        
        Accuracy & MCP $|$ CS TS-ATC & ENT $|$ CS TS-ATC & ENE $|$ CS ATC \\
        Sensitivity  & DOC $|$ CS TS-ATC & ENT $|$ CS ATC & MCP $|$ CS ATC   \\
        Precision  & DOC $|$ CS TS & MCP $|$ CS DoC & DOC $|$ CS DoC \\
        F1-Score  & DOC $|$ CS ATC & ENT $|$ CS ATC & MCP $|$ CS ATC   \\
        AUC  &  ENT $|$ CS TS & DOC $|$ CS TS & DOC $|$ CS DoC  \\

        \midrule
        \midrule

        \multirow{2}{*}{\shortstack{Estimating \\ Metric}} & \multicolumn{3}{c}{Segmentation-2D}\\
        & Choice 1 & Choice 2 & Choice 3 \\
        \cmidrule(r){1-1} \cmidrule(r){2-4} 
        
        F1-Score  & ENT $|$ CS DoC & DOC $|$ CS DoC & MCP $|$ CS ATC \\
        Sensitivity  & MCP $|$ ATC & MCP $|$ TS-ATC & DOC $|$ ATC  \\
        Precision  & ENT $|$ CS TS & MCP $|$ CS DoC & ENT $|$ CS DoC   \\
        Accuracy & MCP $|$ ATC & ENT $|$ CS DoC & MCP $|$ CS TS \\
        AUC  &  ENE $|$ TS & MCP $|$ CS TS & ENE $|$ CS DoC  \\

        \midrule
        \midrule

        \multirow{2}{*}{\shortstack{Estimating \\ Metric}} & \multicolumn{3}{c}{Segmentation-3D}\\
        & Choice 1 & Choice 2 & Choice 3 \\
        \cmidrule(r){1-1} \cmidrule(r){2-4} 
        
        F1-Score  & DOC $|$ CS TS & ENT $|$ CS TS & ENE $|$ CS TS-ATC  \\
        Sensitivity  & ENE $|$ CS DoC & MCP $|$ CS TS & ENE $|$ CS TS-ATC  \\
        Precision  & MCP $|$ CS TS & MCP $|$ CS DoC & DOC $|$ TS   \\
        Accuracy & MCP $|$ TS & MCP $|$ CS TS & DOC $|$ CS DoC \\
        AUC  &  MCP $|$ CS TS & MCP $|$ TS & MCP $|$ CS DoC \\
        
        \bottomrule
    \end{tabular}%
\caption{The confidence scores chosen for creating MOVAL Ensemble.}
\label{tab:moval_ensemble}
\end{table*}

\begin{table*}
\centering
\newsavebox{\tableboxauglist}
\begin{lrbox}{\tableboxauglist}

\begin{tabular}{m{28mm}<{\centering}m{22mm}<{\centering}m{46mm}<{\centering}m{80mm}<{\centering}m{35mm}<{\centering}}
\hline
Category & Operation ID & Operation Name & Description & Range of magnitudes \\
\hline
\multirow{18}{*}{\tabincell{c}{Spatial \\ transformations}} 
& 1,2,3,4,5 & Scaling down & Scale down the image by factor 1 + $M$ & [0.05, 0.15, 0.25, 0.35, 0.45] \\
\cmidrule{2-5}
& 6,7,8,9,10 & Scaling up & Scale up the image by factor 1 + $M$ & [0.05, 0.15, 0.25, 0.35, 0.45] \\
\cmidrule{2-5}
& 11,12,13,14,15 & RotateFrontal ACW & Rotate the image along frontal axis anticlockwise & [5$\degree$, 15$\degree$, 25$\degree$, 90$\degree$, 180$\degree$] \\
\cmidrule{2-5}
& 16,17,18,19 & RotateFrontal CCW & Rotate the image along frontal axis clockwise & [5$\degree$, 15$\degree$, 25$\degree$, 90$\degree$] \\
\cmidrule{2-5}
& 20,21,22,23,24 & RotateSagittal ACW & Rotate the image along sagittal axis anticlockwise & [5$\degree$, 15$\degree$, 25$\degree$, 90$\degree$, 180$\degree$] \\
\cmidrule{2-5}
& 25,26,27,28 & RotateSagittal CCW & Rotate the image along sagittal axis clockwise & [5$\degree$, 15$\degree$, 25$\degree$, 90$\degree$] \\
\cmidrule{2-5}
& 29,30,31,32,33 & RotateLongitudinal ACW & Rotate the image along longitudinal axis anticlockwise & [5$\degree$, 15$\degree$, 25$\degree$, 90$\degree$, 180$\degree$] \\
\hline
\end{tabular}
\end{lrbox}
\scalebox{0.6}{\usebox{\tableboxauglist}}
\caption{List of all transformations that we use to generate synthetic domain shifts for segmentation. There are 24 types of operations with a total of 83 choices. For 2D segmentation, we exclude spatial transformations in the frontal and sagittal axes. For 3D segmentation with advanced training strategies, we exclude all spatial transformations as some algorithms impose strong spatial priors.}\label{tabteaug}
\end{table*}

\end{document}